# Improving Model Performance by Adapting the KGE Metric to Account for System Non-Stationarity

M Jawad[1], HV Gupta[1], YH Wang[1,2], MA Farmani[1], A Behrangi[1], and GY Niu[1]

[1]Department of Hydrology and Atmospheric Sciences, The University of Arizona, Tucson, AZ, USA, 85721.

[2] Earth and Environmental Science Area, Lawrence Berkeley National Lab, Berkeley, CA

## Key Points:

- Traditional metrics fail to account for temporal shifts in system dynamics, which can lead to misleading assessments of model performance.
- The $JKGE_{ss}$ metric detects and accounts for dynamical non-stationarity, thereby improving information extraction and model performance.
- Performance gains are robust to model adequacy, model type (physical-conceptual or data-driven), and hydroclimatic condition.

## Abstract

Geoscientific systems tend to be characterized by pronounced temporal non-stationarity, arising from seasonal and climatic variability in hydrometeorological drivers, and from natural and anthropomorphic changes to land use and cover. As has been pointed out, such variability renders "*the assumption of statistical stationarity obsolete in water management*", and requires us to "*account for, rather than ignore, non-stationary trends*" in the data. However, metrics used for model development are typically based in the implicit and unjustifiable assumption that the data generating process is time-stationary. Here, we introduce the *Jawad-Kling-Gupta Efficiency* ($JKGE_{ss}$) metric, adapted from $KGE_{ss}$, that detects and accounts for dynamical non-stationarity in the statistical properties of the data and thereby improves information extraction and model performance. Unlike $NSE$ and $KGE_{ss}$ that use the long-term mean as a benchmark against which to evaluate model efficiency, $JKGE_{ss}$ emphasizes reproduction of temporal variations in system storage. We tested robustness of the new metric by training physical-conceptual and data-based catchment-scale models of varying complexity, across a wide range of hydroclimatic conditions from *recent-precipitation-dominated* to *snow-dominated* to strongly *arid*. In all cases, the result was improved reproduction of system temporal dynamics at all time scales, across wet to dry years, and over the full range of flow levels (especially recession periods). Since traditional metrics fail to adequately account for temporal shifts in system dynamics, potentially resulting in misleading assessments of model performance under changing conditions, we recommend the adoption of the $JKGE_{ss}$ for geoscientific model development.

## Plain Language Summary

Geoscientific systems tend to be characterized by pronounced temporal non-stationarity, arising from seasonal and climatic variability in the hydrometeorological drivers, and from natural and anthropomorphic changes to land and cover. Such variability renders "*the assumption of statistical stationarity obsolete in water management*", and requires us to "*account for, rather than ignore, non-stationary trends*" in the data. The $JKGE_{ss}$ metric introduced here detects and accounts for dynamical non-stationarity in the data, and thereby improves information extraction and model performance. Unlike metrics that evaluate model efficiency against the long-term observed mean, $JKGE_{ss}$ emphasizes the reproduction of temporal variations in system storage. We tested robustness of the new metric by training physical-conceptual and data-based catchment-scale models of varying complexity, across a wide range of hydroclimatic conditions from *recent-precipitation-dominated* to *snow-dominated* to strongly *arid*. In all cases, the result was improved reproduction of system temporal dynamics at all time scales, across wet to dry years, and over the full range of flow levels (especially recession periods). Since traditional metrics fail to adequately account for temporal shifts in system dynamics, potentially resulting in misleading assessments of model performance under changing conditions, we recommend the adoption of the $JKGE_{ss}$ for geoscientific model development.

## 1. Introduction

### 1.1. Background and Motivation

[1] When constructing models of dynamical systems, a fundamental aspect is the proper choice of an evaluation framework that can be used:

a) During *model development* to extract relevant information about model architecture and parameter values (which together determine model behavior) from the data, and

b) During *model performance evaluation* as useful assessments of model adequacy.

During model *development*, this is typically achieved by designing/selecting one or more scalar-aggregate-metrics of overall model performance that can provide feedback in the form of (i) "*values*" that indicate the current degree of model adequacy, and (ii) "*gradients*" that indicate in which "*directions*" to make adjustments to improve model performance. These same metrics can then be used during *performance evaluation*, in conjunction with various other complementary metrics and/or diagnostic plots, to assess both the quality of the model and its suitability for the task at hand.

[2] It is current common practice, for both physics-based and machine-learning-based model development, to use a *single* scalar-aggregate-metric to guide the parameter estimation (training) process (*Gupta et al., 2009; Kratzert et al., 2019; Lin et al., 2024; Jawad et al., 2024*), as this approach facilitates use of powerful automated optimization algorithms. Certainly, vector-metric (i.e., multiple scalar-metric) approaches are also possible *(Gupta et al., 1998; Boyle et al., 2001*) when used in conjunction with multi-criteria optimization (*Yapo et al., 1998; Gupta et al., 1998; Deb et al., 2002; Efstratiadis & Koutsoyiannis, 2010*), and these can be very effective – especially when the metrics are designed to be mutually complementary and to measure performance in terms of diagnostic signature properties of the system *(Wagener et al., 2007; Yilmaz et al., 2008; Gupta et al., 2008*). Nonetheless, due in part to vector-metric approaches typically providing multiple "*Pareto-optimal*" solutions (in which no single resulting model solution can be judged as being "*better*" than any other without the user imposing additional subjective preferences), the single-aggregate-metric approach remains the dominant paradigm.

[3] Therefore, when choosing a single aggregate metric to guide model development, the design of that metric will determine overall effectiveness of the training procedure (*Sorooshian, 1978; Sorooshian & Dracup, 1980*). Aggregating a time series of model residuals into a single summary statistic inevitably entails loss of information *(Gupta et al., 2008*), design of a robust metric requires proper consideration of sampling variability/uncertainty (*Zheng et al., 2018; Clark et al., 2021*), and no single metric can be universally appropriate for all hydrological applications (*Gupta et al., 2009; Santos et al., 2018*).

### 1.2. Limitations of NSE and the Rise of KGE

[4] Nonetheless, in hydrology, it has long been common to use the *Mean Squared Error* ($MSE$; Eq 1), or its variance-normalized version the *Nash–Sutcliffe Efficiency (NSE;* Eq 2; *Nash & Sutcliffe, 1970*), to summarize/measure the overall degree of agreement between simulated and observed system behavior, a practice that persists *(Kratzert et al., 2018; Kratzert et al., 2019; Nearing et al., 2021)* despite critiques regarding its suitability *(Legates & McCabe Jr, 1999; Gupta et al., 2009; Clark et al., 2021*), particularly in the context of catchment-scale streamflow prediction. In fact, the landmark 1970 paper by *Nash and Sutcliffe* remains one of the most cited in hydrological science (with estimates exceeding 28,000 citations on Google Scholar and over 18,000 in Scopus), and continues to be a standard metric for evaluating model performance in hydrology and the environmental sciences (*Melsen et al., 2025*).

[5] A little over a decade and a half ago, *Gupta et al. (2009)* presented a theoretical analysis of the $MSE$ (and hence $NSE$) metric, showing how it can be decomposed into three components representing mean (water balance) bias, variability bias, and temporal cross-correlation. Based on this, they showed that

optimizing on $MSE$ ($NSE$) during model development would, both theoretically and in practice, result in a tendency to underestimate the variability of the target time series (e.g., streamflow) particularly when the model is a simplified representation of reality (i.e., when model errors are significant) – see *(Gupta et al., 2012)* for an extended discussion of model inadequacy. This explained clearly why hydrological models trained using these metrics tend to underestimate flood peak magnitudes and provide positively biased estimates of streamflow recessions, periods for which model predictive accuracy is especially important. To address this problem, they proposed the *Kling–Gupta Efficiency* (KGE; Eq 3; *Gupta et al., 2009; Gupta & Kling, 2011*) that combines the decomposition components of $MSE$ ($NSE$) (i.e., water balance error, variability error, and temporal cross-correlation) into a single aggregate metric in a somewhat different manner. The $KGE$ has since been widely adopted, is frequently cited as a crucial alternative to $NSE$ (accumulating over 5,000 citations since it was proposed), and has become a standard tool for calibration and evaluation of hydrological (and other domain science) models.

[6] A perhaps less well-recognized aspect of *Gupta et al. (2009)* is that it helps to bridge the conceptual gap between scalar- and vector-metric approaches by demonstrating how a metric can be designed with proper consideration of diagnostic signature properties of the system (properties the user wants the model to reproduce). In fact, they point out that the primary purpose of their study was "*not to design an improved measure of model performance*" but to show that "*an analysis of [diagnostic] components that constitute the overall model performance can significantly enhance our understanding of model behavior and provide insights for diagnosing differences between models … within a [given] context*".

[7] Accordingly, the $\beta_{so}$ (*Beta*; Eq 4) , $\alpha_{so}$ (*Alpha*; Eq 5) and $\rho_{so}$ (*Rho*; Eq 6) statistical components that make up $KGE$ enable the user to track model performance in an ***interpretable*** manner, and in more detail - $\beta_{so}$ measures performance with respect to matching the (stationary) *long-term mean of the observed signal*, $\alpha_{so}$ measures performance with respect to matching the *observed signal variance*, and $\rho_{so}$ measures performance in terms of matching the *dynamical correlative pattern (timing and shape) of the observed signal*. These statistics (signal mean, variance, and cross-correlation) serve as ***diagnostic signature properties*** (*Gupta et al., 2008*) that can be used to reveal sources of model inadequacy.

### 1.3. Benchmarking

[8] Nonetheless, one important property shared by both $NSE$ and $KGE$, and in particular the skill-score adjusted version of the latter ($KGE_{ss}$; *(Knoben et al., 2019*), is their use of the observed *long-term mean* (LTM) as a "*benchmark*" against which performance is evaluated. The rationale for this choice of benchmark is that:

a) The upper-bound value of 1.0 indicates perfect performance, due to perfect matching of the observed signal (and therefore its *variance* and *dynamical pattern* around that LTM).

b) The baseline value of 0.0 indicates that performance is (from an aggregate metric point of view) no better than discarding the model and instead using the temporally-constant observed LTM as the "*model*" (i.e., the simulation/prediction at every time step is simply set to the LTM).

c) Negative values indicate undesirably poor performance, due largely to severe mass balance errors in the data *(Gupta et al., 2009; Gupta & Kling, 2011)*.

[9] It is important to note that explicit benchmarking practice has been advocated for at least the past quarter century *(Garrick et al. 1978; Schaefli et al. 2005; Schaefli & Gupta, 2007; Clark et al., 2015; Clark et al., 2021; Knoben, 2024*), and the choice of observed LTM as a benchmark has been subject to criticism *(Schaefli & Gupta, 2007; Clark et al., 2015; Clark et al., 2021; Knoben, 2024*). For example, for catchments showing strong seasonality, a proposed benchmark has been the "*interannual mean value for every calendar day*" (effectively the climatology; *Garrick et al. (1978), Schaefli et al. 2005*). Alternatively, *Schaefli and Gupta (2007)* suggest an "*adjusted smoothed precipitation benchmark*" (APB) that measures whether

the hydrologic model has more explanatory power than already contained in the frequency content of the dominant driving process (i.e. rainfall). More recently, *Knoben (2024)* proposed evaluating hydrologic models against an ensemble of simple, process-relevant benchmark models rather than relying on a single baseline. He showed that structured benchmarking provides clearer expectations for model performance and enables a more objective assessment of model skill and added value.

[10] However, despite development of structured frameworks and supporting theoretical foundations (*Nearing & Gupta, 2015; Nearing et al., 2018*), formal application of benchmarks other than the LTM remains relatively uncommon in hydrological modeling.

### 1.4. Accounting for Temporal Non-Stationarity

[11] Importantly, what has gone largely unrecognized is that choice of the LTM as benchmark imposes a major limitation on the *ability of the metric to extract relevant information from the data*. This is because doing so (implicitly and incorrectly) assumes that the stochastic data generating process (DGP) giving rise to the observed system behavior (e.g., streamflow) is "*time-stationary*".

[12] In actual fact, geoscientific (e.g., hydrological) systems tend to be characterized by pronounced temporal non-stationarity – arising from seasonal and climatic variability in the hydrometeorological drivers, and also from natural and anthropomorphic changes to land use and land cover type. As pointed out by *Milly et al. (2008)* among others, climate and land use changes "*make historical hydrological patterns unreliable for future planning, rendering the assumption of statistical stationarity obsolete in water management*", requiring us to "*account for, rather than ignore, non-stationary trends*" in the data.

[13] In this regard, a simple analysis of hydrometeorological and streamflow data for catchments across the continental US (**Figure 1**) shows that seasonal, interannual and longer term temporal variability is sufficiently pronounced that the DGP giving rise to observed behaviors cannot reasonably be assumed to be stationary. Further, there is little to suggest that other statistical properties of the catchment-scale streamflow DGP (besides the mean), evaluated over whatever time scale, are actually stationary in time.

[14] Given this, it seems reasonable to question (from an *information extraction* point of view) the desirability of using of the time-stationary observed LTM (and indeed the use of any other observed time-stationary long-term property of the system, such as climatology) as the benchmark when constructing an "*efficiency*" metric. Relying on a time-stationary benchmark can, in principle, cause the metric to fail to adequately account for temporal shifts in system dynamics, resulting in misleading assessments of model performance under changing conditions.

### 1.5. Goals and Scope of this Study

[15] The goal of this study is to investigate whether use of a "*context-aware*" benchmark that *explicitly accounts for temporal non-stationarity* during metric design and implementation (for model development and performance evaluation) can improve information extraction from the data. Doing so should result in a more robust model that reproduces all relevant dynamical aspects of behavior in a balanced and appropriate manner compared to what can be achieved under the assumption that the DGP is stationary.

[16] We test this hypothesis across a diversity of hydroclimatic zones and for different model types, ranging from physical-conceptual models of various levels of complexity/adequacy to pure machine learning-based models.

[17] Section 2 introduces the mathematical explanation of the newly proposed metric ($JKGE_{ss}$; Eq 8) designed to explicitly account for temporal non-stationarity by replacing the traditional constant benchmark with a context-aware time-varying one derived from observed data. Importantly, the section

also details the different types of non-stationary benchmarks considered and how they are constructed for use in the analysis.

[18] Section 3 outlines the methodology adopted in this study, including selected catchments, datasets used, and data preprocessing steps such as training–evaluation splitting. It further describes the training strategies, and the range of model architectures used to evaluate the $JKGE_{ss}$ metric.

[19] Section 4 presents model results obtained using both time-stationary ($KGE_{ss}$) and time-non-stationary ($JKGE_{ss}$) metrics, demonstrating how $JKGE_{ss}$ improves information extraction during training, and assessing cross-metric performance between models trained with $JKGE_{ss}$ and $KGE_{ss}$.

[20] Section 5 further examines augmented and ablated versions of $JKGE_{ss}$, evaluates its robustness to model adequacy (including purely data-based machine-learning models), and assesses the potential value of log-transforming streamflow before applying the metric.

[21] Section 6 presents our conclusions and outlines directions for future work.

## 2. Designing a Metric to Explicitly Account for Temporal Non-Stationarity

### 2.1. Background on $MSE$, $NSE$, $KGE$ and $KGE_{ss}$

[22] Let $\bar{S} = [s_1, s_2, \dots s_N]^T$ be a vector containing the model-simulated time-series that we seek to make as similar as possible to the corresponding system-observed time-series $\bar{O} = [o_1, o_2, \dots o_N]^T$ via the process of model development. Further, let $\bar{B} = [b_1, b_2, \dots b_N]^T$ be a vector containing what we will refer to as the "*benchmark*" time series.

[23] We can compute *estimates* of the long-term means ($\mu_s$ and $\mu_o$ respectively) of these signals as $\mu_s = \frac{1}{N}\sum_{t=1}^{N} s_t$ and $\mu_o = \frac{1}{N}\sum_{t=1}^{N} o_t$ and, consequently, *estimates* of the long-term standard deviations ($\sigma_s$ and $\sigma_o$ respectively) as $\sigma_s = \sqrt{\frac{1}{N}\sum_{t=1}^{N}(s_t - \mu_s)^2}$ and $\sigma_o = \sqrt{\frac{1}{N}\sum_{t=1}^{N}(o_t - \mu_o)^2}$. Finally, we can compute an *estimate* of the temporal cross-correlation between the simulated and observed signals as $\rho_{so} = \frac{1}{N}\sum_{t=1}^{N}\left(\frac{s_t - \mu_s}{\sigma_s}\right)\left(\frac{o_t - \mu_o}{\sigma_o}\right)$.

[24] Then the $MSE$ metric, and its variance-normalized version $NSE$ are defined as:

$$MSE = \frac{1}{N}\sum_{t=1}^{N}(s_t - o_t)^2 \qquad \text{(Eq. 1)}$$

$$NSE = 1 - \frac{MSE}{\sigma_o^2} \qquad \text{(Eq. 2)}$$

where setting $s_t = \mu_o$ at every time step results in $MSE = \frac{1}{N}\sum_{t=1}^{N}(s_t - \mu_o)^2 \approx \sigma_o^2$ so that $NSE = 1 - \frac{\sigma_o^2}{\sigma_o^2} = 0$ under those conditions.

[25] Meanwhile, the $KGE$ metric (*Gupta et al. 2009; Gupta and Kling 2011*) is defined as:

$$KGE = 1 - \sqrt{\left(1 - \frac{\mu_s}{\mu_o}\right)^2 + \left(1 - \frac{\sigma_s}{\sigma_o}\right)^2 + (1 - \rho_{so})^2} \qquad \text{(Eq. 3)}$$

where:

$$\beta_{so} = \frac{\mu_s}{\mu_o} \qquad \text{(Eq. 4)}$$

$$\alpha_{so} = \frac{\sigma_s}{\sigma_o} \qquad \text{(Eq. 5)}$$

$$\rho_{so} = \frac{1}{N}\sum_{t=1}^{N}\left(\frac{s_t - \mu_s}{\sigma_s}\right)\left(\frac{o_t - \mu_o}{\sigma_o}\right) \qquad \text{(Eq. 6)}$$

and defining $M = (1 - \beta_{so})^2$, $V = (1 - \alpha_{so})^2$, and $C = (1 - \rho_{so})^2$ we can write:

$$KGE = 1 - \sqrt{M + V + C} \qquad \text{(Eq. 7)}$$

[26] To develop the skill-score adjusted version $KGE_{ss}$, wherein a value of $0$ corresponds to performance being no better than using the observed LTM as the "*benchmark*" against which model is evaluated, we set $s_t = \mu_o$ at every time step resulting in $\mu_s = \mu_o$, $\sigma_s = 0$ and $\rho_{so} = 0$, so that $M = 0$ (because $\beta_{so} = 1$), $V = 1$, and $C = 1$. Substituting in Eq. 6 we obtain $KGE = 1 - \sqrt{2}$ which is *not* equal to $0$. Accordingly, by defining $KGE_{ss} = 1 - \sqrt{(M + V + C)/2}$, we obtain the skill-score adjusted version in which $KGE_{ss} = 0$ when $s_t = \mu_o$ at every time step (*Knoben et al., 2019*).

### 2.2. Consequences of Using the Observed Long-Term-Mean as Benchmark

[27] As discussed above, a defining characteristic of $NSE$ and $KGE_{ss}$ is the crucial role of the observed *long-term mean* (LTM) $\mu_o$ as the "*benchmark*" against which performance is evaluated. In other words, the benchmark time series is chosen to be $\bar{B} = [\mu_o, \mu_o, \dots \mu_o]^T$.

[28] From Eq. 3 we see that, to maximize $KGE_{ss}$, the optimization procedure must drive $\beta = \mu_s/\mu_o \to 1.0$, at which point $\mu_s = \mu_o$ regardless of whether any other aspect of signal dynamics (such as variability or temporal pattern) has been reproduced. Similarly, the procedure must also drive $\alpha_{so} = \sigma_s/\sigma_o \to 1.0$, at which point $\sigma_s = \sigma_o$ regardless of whether any other aspect of signal dynamics (such as mean or temporal pattern) has been reproduced. And finally, it must also drive $\rho_{so} \to 1$, at which point the temporal pattern has been matched regardless of whether any other aspect of signal dynamics (such as mean or variability) has been reproduced; this final condition will generally be impossible to perfectly achieve due to random measurement errors in the observed time-series $\bar{O}$ and due to inadequacy of the model itself.

[29] Note, also, that the values computed for $\sigma_s$ and $\sigma_o$ depend on values previously computed for $\mu_s$ and $\mu_o$, while the value for $\rho_{so} = \frac{1}{N}\sum_{t=1}^{N} z_t^s \cdot z_t^o$ , where the standardized values $z_t^s = \left(\frac{s_t - \mu_s}{\sigma_s}\right)$ and $z_t^o = \left(\frac{o_t - \mu_o}{\sigma_o}\right)$ depend in turn on the previously computed values for $\mu_s$, $\mu_o$, $\sigma_s$ and $\sigma_o$. So, $\alpha_{so}$, $\beta_{so}$, and $\rho_{so}$ (and therefore $V$, $M$ and $C$) in $KGE_{ss}$ *depend ultimately* on the values computed for $\mu_s$ and $\mu_o$.

[30] From this analysis, it becomes apparent that the $\bar{B}$ time-series chosen as "*benchmark"* when defining the efficiency metric can dramatically influence the *nature of the information that is extracted from the data* during model development and training, and hence the final result. For example, it has been well-established that use of $NSE$ (or $MSE$) for model development typically results in models that reproduce higher-magnitude flows better than intermediate- and lower-magnitude flows (in main part due to the strongly skewed nature of the distribution of observed streamflow). And, although use of $KGE$ (or $KGE_{ss}$) tends to improve reproduction of overall flow variability, this bias towards better reproduction of higher-magnitude flows is not mitigated – a tendency that will be well-illustrated in Section 4.9 (along with the fact that *log-transforming* the flows before application of the metric results in the opposite problem – better reproduction of low-magnitude flows at the expense of high-magnitude flows).

[31] Further, as anticipated by *(Garrick et al., 1978)* and later by *(Schaefli & Gupta, 2007)*, the benchmark series $\bar{B}$ need not consist of a time-constant value (such as $\mu_o$), but can instead be any temporally (and more generally spatio-temporally) varying value that represents a meaningful baseline against which improvement is to be evaluated (i.e., the simulations $\bar{S}$ must be "*closer*" to the observations $\bar{O}$ than is the benchmark $\bar{B}$). The challenge, however, will be to select $\bar{B}$ in such a manner that it:

a) Represents a meaningful (interpretable) trajectory against which performance should be measured (and thereby facilitate a useful assessment of performance), and

b) Can facilitate *improved extraction of information from the data*, so that the model can reproduce all relevant dynamical aspects of observed behavior (across its full range) in a balanced and appropriate manner.

[32] In the next sub-section we discuss how the $KGE_{ss}$ metric can be adapted to explicitly account for temporal non-stationarity.

### 2.3. Accounting for Temporal Non-Stationarity

[33] Now, consider that the benchmark is defined as $\bar{B}^o = [b_1^o, b_2^o, \dots b_N^o]^T$ where $b_t^o = f_b(t|\bar{O})$ represents a time-varying value obtained by applying some pre-specified functional operation, represented symbolically by $f_b(.)$, to the observed signal $\bar{O}$. For example, $\bar{B}^o$ could represent some linear, or cyclical, or other trend in the signal (see **Figure 2**). As such $\bar{B}^o$ can be chosen to characterize any short-to-medium-to-long-term time-non-stationary properties of the observed signal. This could include non-stationarities at the sub-annual time scale (such as seasonality and/or climatology) but also longer-term temporal non-stationarities (see **Figure 1**) – for instance, mean annual rainfall (and therefore annual streamflow) volumes tend to vary significantly from year-to-year.

[34] Accordingly, a time-non-stationary benchmark can be readily computed from the observed data via some operation $f_b(t|\bar{O})$, and used as a provisional "*no-model*" prediction (or more correctly a "*minimal-model*" prediction) that can be made using only knowledge of temporal patterns (trend, cycles, etc.) extracted from the observed data. Once the operation $f_b(t|\bar{O})$ has been chosen, the corresponding model-simulated analog of this observed benchmark can be computed as $b_t^s = f_b(t|\bar{S})$ by applying the same functional operation $f_b(.)$ to the simulated signal $\bar{S}$.

[35] We can now adapt the $KGE_{ss}$ metric to account for the temporal non-stationarity observed in the system response by defining the *Jawad-Kling-Gupta Efficiency* metric:

$$JKGE_{ss} = 1 - \sqrt{(M^* + V^* + C^*)/2} \quad \text{(Eq. 8)}$$

where $M^* = \frac{1}{N}\sum_{t=1}^{N}\left(1 - \frac{b_t^s}{b_t^o}\right)^2$, $V^* = (1 - \alpha_{so}^*)^2$ and $C^* = (1 - \rho_{so}^*)^2$ with:

$$\alpha_{so}^* = \frac{\psi_s}{\psi_o} \quad \text{(Eq. 9)}$$

$$\rho_{so}^* = \frac{1}{N}\sum_{t=1}^{N}\left(\frac{s_t - b_t^s}{\psi_s}\right) \cdot \left(\frac{o_t - b_t^o}{\psi_o}\right) \quad \text{(Eq. 10)}$$

where $\psi_s = \sqrt{\frac{1}{N}\sum_{t=1}^{N}(s_t - b_t^s)^2}$ and $\psi_o = \sqrt{\frac{1}{N}\sum_{t=1}^{N}(o_t - b_t^o)^2}$.

[36] Note that $M^*$ is analogous to $M$ in the classical $KGE_{ss}$, such that $M^* = 0.0$ when $b_t^s = b_t^o$ for all $t = 1, \dots, N$, in which case the model reproduces the dynamics of the benchmark signal computed from the observed data (this means that $\bar{B}^s = \bar{B}^o$, but does <u>*not*</u> mean that $\bar{S} = \bar{O}$). Similarly, $V^*$ and $C^*$ are analogous to $V$ and $C$, in the sense that $\psi_o$ represents the standard deviation of the benchmark-centered "*observed anomaly*" time-series $\bar{A}^o = \bar{O} - \bar{B}^o$ (obtained by subtracting the time-non-stationary benchmark) and $\psi_s$ represent the analogous standard deviation of the "*simulated anomaly*" time-series $\bar{A}^s = \bar{S} - \bar{B}^s$. Accordingly, $V^*$ and $C^*$ measure how well the model simulations reproduce the variability and temporal dynamics of the benchmark-centered "*observed anomaly*" time-series.

[37] As a trivial but illustrative example of temporal non-stationarity, consider that the observed data $\bar{O}$ appears to be describable as $o_t = b_t^o + \varepsilon_t^o$, where $b_t^o$ follows a linear time-non-stationary trend $b_t^o = f_b(t|\bar{O}) = m_o \cdot t + c_o$ that can extracted from the data. We can correspondingly define $b_t^s = f_b(t|\bar{S}) = m_s \cdot t + c_s$. Since the simulated and observed anomalies are now $a_t^s = (s_t - b_t^s)$ and $a_t^o = (o_t - b_t^o)$, we

can compute the standard deviations $\psi_s = \sqrt{\frac{1}{N}\sum_{t=1}^{N}(a_t^s)^2}$ and $\psi_o = \sqrt{\frac{1}{N}\sum_{t=1}^{N}(a_t^o)^2}$ of the *de-trended* simulated and observed signals respectively and, further, if detrended anomaly time-series $a_t^s$ and $a_t^o$ are temporally autocorrelated we can compute $\rho_{so}^* = \frac{1}{N}\sum_{t=1}^{N}\left(\frac{s_t-b_t^s}{\psi_s}\right)\cdot\left(\frac{o_t-b_t^o}{\psi_o}\right)$.

[38] To train the model, we would then need to drive the descriptors $m_s \to m_o$ and $c_s \to c_o$, (causing $b_t^s \to b_t^o$ so that $M^* \to 0$), and further drive $\psi_s \to \psi_o$ (causing $V^* \to 0$) and $\rho_{so}^* \to 1.0$ (causing $C^* \to 0$) so that $JKGE_{ss} \to 1$, which means that $\bar{S} \to \bar{O}$. If, on the other hand, the model can only generate simulations $\bar{S}$ that follow the linear trend line represented by $\bar{B}^o$, and are unable to reproduce the variability and temporal pattern of the observations around that trend line, then we will at best have $JKGE_{ss} = 0$, due to the fact that $\psi_s = 0.0$ and $\rho_{so}^* = 0.0$.

[39] It is conceptually worth noting that by replacing the time-constant benchmark of the classical $KGE$ ($KGE_{ss}$) by a time-varying benchmark in the $JKGE_{ss}$ formulation, we are shifting some of the representation of the *variability* and *pattern* of the "*dynamics*" out of $V$ and $C$, and into $M$. This will affect how the metric extracts information from the data, by changing its "*focus/attention*" (the importance it gives to how the model fits various aspects of the observed signal).

### 2.4. Defining the Temporally Non-Stationarity Benchmark Time-Series

[40] With this theoretical development, all that remains is to choose how to define the operation $f_b(.)$ used to extract the benchmark $\bar{B}^o$ from the observed signal $\bar{O}$ and, analogously, the corresponding $\bar{B}^s$ from the simulated signal $\bar{S}$. We next explore how to do this and show how an appropriate choice can result in dramatically improved performance when the $JKGE_{ss}$ is used for model development.

#### 2.4.1. Time-Nonstationary Mean:

[41] To acknowledge the existence of temporal variability in the DGP giving rise to the observed system behavior, we can use a time-varying mean computed (estimated) from the data. For this study, we investigated two ways in which this can be done:

1) The first, perhaps most obvious, way is to use a time-centered running-moving-average taken over a data window at some prescribed temporal scale; this approach results in a moving-mean benchmark that varies continuously over time.

2) The second way is to instead use a time-varying but sectionally-constant average value that is computed on progressive non-overlapping sections of the data at some prescribed temporal scale.

[42] The basic idea is that progressively shortening the window (section) length enables the benchmark, and consequently the model trained against it, to be better informative about the broad spectrum of flow conditions that exist at the target location (e.g., streamflow gaging point), ranging from long-term trends to shorter-term fluctuations.

##### *2.4.1.1. Time-Nonstationary Moving-Average Mean:*

[43] For this approach, we define the benchmark value at each time step as being the smoothly-varying arithmetic mean of the observed time series values within a moving window of fixed length $N_w$, such that $b_t^o = \frac{1}{N_w}\sum_{i=t-k}^{t+k} O_t^i$, where $k = \frac{N_w-1}{2}$. By centering the window on the current time step and choosing $N_w$ to be an odd-valued integer, the current day is at the center of the window and has an equal number of days preceding and following it (e.g., for $N_w = 7$ the window spans three days before and after the current day). Similarly, for computing $JKGE_{ss}$, we define $b_t^s = \frac{1}{N_w}\sum_{i=t-k}^{t+k} S_t^i$ .

[44] We tested moving-average window sizes of 365, 181, 91, 61, 31, and 7 days, representing progressively finer temporal resolutions. **Figure 3a** shows (in the log-transformed space) an illustration where the green dashed line represents the observed streamflow signal, the horizontal black dashed line represents the long-term mean, and the differently-colored solid lines show time-nonstationary $b_t^o$ benchmark time series computed using different window lengths. As should be expected, larger window lengths produce smoother benchmarks that emphasize longer-term trends, while shorter window lengths allow the benchmark to more closely track intra-annual, seasonal and sub-seasonal variability.

[45] Hereafter, we use the notation $JKGE_{ss}^{MA(N_w)}$ to indicate use of this approach, where the superscript MA indicates "*moving average*" and the value $N_w$ in brackets specifies the (odd-integer-valued) length of the moving-average window. Note that when computing $b_t^o$ for the first and last $\frac{N_w - 1}{2}$ time steps, a full window of length $N_w$ is unavailable and so those timesteps cannot be assigned a benchmark value. For instance, with $N_w = 7$, the first three and last three points are excluded from the calculation. Larger window lengths result in proportionally more points being lost at the boundaries and were incorporated accordingly during training. Other strategies to estimate moving average values close to the end points of a time series do exist (*Gu & Zhou, 2010; Höll et al., 2019*), but were not investigated here.

#### ***2.4.1.2. Time-Nonstationary Section-Wise Mean:***

[46] For this approach, we instead define the benchmark $b_t^o$ at each time step as being a "*sectional mean*" of some fixed length $N_s$, where the sections consist of consecutive non-overlapping portions of the data. Accordingly, the same (fixed) benchmark value is assigned to all days within a section, and the value changes only from section to section, so that the time-sequence $b_t^o$ used to compute $JKGE_{ss}$ varies in a "*step-wise*" manner (see **Figure 3b**); i.e., $b_t^o = \frac{1}{N_s} \sum_{i \in G(t)} O_i$ where $G(t)$ indicates the section to which time step $t$ belongs, and $O_i$ indicates the observed timeseries within that section. Similarly, for computation of $JKGE_{ss}$, we define $b_t^s = \frac{1}{N_s} \sum_{i \in G(t)} S_i$.

[47] As above, we tested section lengths ($N_s$) of 365, 180, 90, 60, 30 and 7 days to represent progressively finer temporal resolutions, and **Figure 3b** shows results for the 5 water-years of streamflow data that were used to construct **Figure 3a**. The result is a coarse approximation to the more smoothly varying results obtained using the moving-average approach. However, the main behavioral characteristics of temporal (intra-annual, seasonal, and sub-seasonal) variability of the overall time series are reasonably well captured, especially as the section lengths become shorter. Hereafter, we use the notation $JKGE_{ss}^{SA(N_s)}$ to indicate use of this approach, where the superscript SA indicates "*section-wise average*" and the value $N_s$ in brackets specifies the length of the section.

### 2.4.2. Impact of Accounting for Non-Stationarity on the Streamflow Anomaly Distributions

[48] It is interesting to note that as we progressively shorten the segment (section or window) of streamflow data over which the benchmark time series $b_t^o$ is computed (from using the entire data, to 365 days, and so on down to 7 days), the marginal distribution $p(Z^{AL})$ of standardized anomalies of log-transformed observed streamflow values ($Z_t^{AL} = \frac{ln(O_t) - ln(b_t^o)}{\Psi_o}$ where $\Psi_o \approx \sqrt{\frac{1}{N} \sum_{t=1}^{N} \left( ln(O_t) - ln(b_t^o) \right)^2}$) becomes increasingly well-behaved (**Figure 4**).

[49] **Figure 4** illustrates the *Section-Wise Mean* approach (see also Supplementary **Figure S2**), while virtually identical results for the *Moving-Average Mean* approach are shown as Supplementary **Figure S1**. The leftmost column of **Figure 4** shows Kernel Density Estimates (KDEs) of the distributions of standardized anomalies for each of the 5 representative catchments, one from each hydroclimatic category (see Section 3.1 for details on the catchments list and classification) when using the non-time-varying long-term (LT)

mean as the benchmark time series $b_t^o$. As can be seen, and consistent with the fact that daily streamflow data are (typically) quite heavily skewed (*Zheng et al., 2018*), the distributions (indicated by solid red lines) exhibit noticeable asymmetry, and also some degree of bi-modality; the latter is particularly apparent for the *recent rainfall-dominated systems* in the Western US.

[50] This distributional behavior of the anomalies is also true when using the longer segment lengths of $N_s$=365 ($N_w$=365) and $N_s$=180 ($N_w$=181 days) — see second and third columns from left. However, as the segment lengths become shorter ($N_s$= 90 or $N_w$= 91 days and less) and more representative of *seasonal* variations, the anomaly distributions become much more symmetrical about zero. Interestingly, when using section lengths of 7 days (rightmost column), the anomaly distributions almost perfectly approach a double exponential density function.

[51] Overall, at longer section lengths (from LT to 180 days), the *snow-dominated catchment* (row 2) exhibits the most stable and well-structured anomaly distributions, consistent with streamflow dynamics strongly governed by seasonal storage and delayed release processes. In contrast, the *arid catchment* (row 5) shows the strongest departures from symmetry at long segment lengths, and its anomaly distribution is noticeably better behaved only when the segments start to represent sub-seasonal scales of temporal variability (30–7 days), reflecting the dominance of intermittent and event-driven runoff.

[52] Among the *rainfall-driven systems*, the *recent rainfall-dominated catchments* (both western → row 1, and eastern → row 4) require shorter segment lengths to achieve symmetrical anomaly distributions, consistent with strong prevalence of high-frequency rainfall variability and short response times. Finally, the *historical rainfall-dominated catchment* (row 3) shows intermediate behavior, with improvements emerging at seasonal time-scales, and getting better as we reach $N_s$= 30 or $N_w$= 31 days.

[53] We will show later (Section 4.1) that the results in **Figure 4** are consistent with the fact that shorter segment lengths result in better information-extraction from data when using $JKGE_{ss}$ for model training. This is particularly important for *rainfall-dominated* and *arid* regimes that are characterized by higher-frequency variability in their daily streamflow, whereas the information about *snow-dominated* catchments can be adequately characterized using longer, seasonally varying segments.

[54] Overall, this trend indicates that progressively shortening the segment length can enhance the ability of benchmark $\bar{B}^o$ to properly represent the full range of hydrologic variability in the data, resulting in anomaly distributions that are closer to symmetrical, and thereby more suitable for robust statistical characterization. Of course, one must be careful to not make the segments too short – in the limit $N = 1$ the benchmark $\bar{B}^o$ becomes identical to the observed time series $\bar{O}$ which may be subject to considerable observational error/noise. In our experiments, excellent results were obtained using segment lengths no smaller than 30 days, from which robust statistically representative estimates of $b_t^o$ can computed.

## 3. Experimental Setup and Methods:

### 3.1. Selection of Catchments:

[51] We followed the classification scheme proposed by *Jiang et al. (2022)* to select a representative set of 15 catchments from the CAMELS dataset, that capture a wide range of hydro-meteorological regimes and runoff-generation mechanisms (see **Table 1** and **Figure 5**). Twelve of the catchments represent four dominant streamflow-generation mechanisms: (i) *recent rainfall-dominated* systems in the Western US, where streamflow responds primarily to precipitation occurring on the same or the immediately preceding day; (ii) *recent rainfall-dominated systems* in the Eastern US (iii) *historical rainfall-dominated* systems, that are characterized by the cumulative influence of precipitation over multiple days to weeks; and (iv) *snowmelt-dominated* systems, in which temperature exerts the primary control on streamflow variability.

The remaining three catchments represent *arid* conditions (aridity index AI > 1), to allow for assessment under water-limited conditions.

### 3.2. Forcing and Target Data:

[52] To force the models, we used the near-surface atmospheric dataset developed by the *National Oceanic and Atmospheric Administration* (NOAA), the *Analysis of Record for Calibration* (AORC), which provides hourly data at 800 m resolution (*Fall et al., 2023*). We aggregated hourly surface air temperature, precipitation, downward shortwave and longwave radiation fluxes, air pressure, wind speed, and specific humidity from AORC v1.1 (Oct 1, 2003–Sept 30, 2023) to the daily timescale used by the models.

[53] For *Leaf Area Index*, we used MODIS LAI (MOD15A2H Level 4), a global dataset with 500 m and 8-day resolution (*Myneni et al., 2015*); MOD15A2H has been found to provide reasonable LAI estimates across all biome types (*Yan et al. (2016)*). Emissivity at 1 km and 8-day resolution was obtained from MODIS MOD21A2 Version 6.1 (*Hulley, 2017*). Hourly albedo values at 9 km resolution were taken from the ERA5-Land reanalysis dataset (*Muñoz-Sabater et al., 2021*) produced by the *European Centre for Medium-Range Weather Forecasts* (ECMWF). All variables were spatially averaged over the catchment.

[54] Daily streamflow data (target variable) in cubic feet per second (cusecs, $ft^3/s$), representing average daily discharge at each gauging station, was acquired from the United States Geological Survey web portal (*USGS, 2025*). Discharge was converted to millimeters per day (mm/day) by converting flow to daily volume and dividing by catchment area, producing area-normalized streamflow.

### 3.3. Model Architectures:

#### 3.3.1. Physical-Conceptual Model Architectures:

[55] To evaluate the impact (on model performance) of accounting for benchmark non-stationarity we tested three AI-augmented, conceptually-interpretable, hydrological model architectures of varying structural complexity (**Figure 6**). All are based in the behaviorally-flexible *Mass Conserving Perceptron* (MCP) introduced by *(Wang & Gupta, 2024a, 2024b, 2025)*. The rationale underlying development of the model architectures tested here (referred to as $MA_2$, $MA_3$, and $MA_5$) is discussed by *Jawad et al. (2026)*. Specifically relevant to this paper, the two-state $MA_2$ (45 trainable parameters) and three-state $MA_3$ (79 trainable parameters) are simplified versions of the five-state $MA_5$ architecture that represents a more comprehensive set of process dynamics (111 trainable parameters).

a) **$MA_2$**: This two-state architecture can be interpreted as representing temporal dynamics of snow water equivalent ($X_t^{S_{snow}}$) and soil storage ($X_t^S$). Three flow pathways collectively contribute to total system outflow ($O_t$), via infiltration excess ($P_t^{S2_{Excess}}$), soil saturation excess ($O1_t^S$), and state dependent outflow ($O2_t^S$). Conceptually, this architecture captures the most basic hydrological processes of snow accumulation/melt and soil water dynamics, while allowing runoff generation through both infiltration-driven and saturation-driven mechanisms.

b) **$MA_3$**: This three-state architecture extends $MA_2$ by adding baseflow storage ($X_t^B$) and a seepage flux that connects the soil state ($X_t^S$) to the baseflow state ($X_t^B$) through a one-directional flow path. This addition enables clearer separation between quicker surface and near-surface flow runoff processes and slower subsurface drainage. Inclusion of baseflow storage allows $MA_3$ to better represent delayed flow contributions and to improve hydrograph recession behavior, while maintaining a relatively parsimonious representation.

c) **$MA_5$**: This five-state architecture extends $MA_3$ by incorporating canopy-related processes. It introduces canopy storage states for both liquid ($X_t^{C_{rain}}$) and frozen snow ($X_t^{C_{snow}}$), representing different phases of snow water equivalent that can be retained in the canopy, enabling the model to represent dynamics of interception, melt, and delayed release of snow from the canopy, thereby improving its' ability to model behaviors of snow-dominated catchments. Compared to $MA_3$, it provides greater flexibility in simulating timing and magnitude of runoff under complex precipitation–canopy interactions.

[56] The inclusion of models with different levels of structural and functional complexity allowed investigation of the robustness of our results to model structural (in)adequacy *(Gupta et al., 2012)*, since a more complex model will tend to have higher capacity to "*fit*" the data. Note that behavioral flexibility of MCP-based models stems from the functional forms of the internal process equations (known as gating functions) being learnable directly from the training data, while being constrained to be consistent with thermodynamic principles governing fluxes; see *Wang and Gupta (2024a)*.

### 3.3.2. Purely ML Based Models:

[57] To assess how accounting for benchmark non-stationarity affects performance of machine-learning (ML) models, we employed single-layer Long-Short-Term-Memory (LSTM; *(Hochreiter & Schmidhuber, 1997)*) networks with 5, 10, and 15 hidden nodes. LSTMs are a gated form of recurrent neural network shown to be highly effective for data-driven modeling of hydrological systems (*Kratzert et al., 2018*). Their strength lies in regulation of the cell-state structure by three context-dependent gating mechanisms that can learn long-term dependencies (*Hochreiter & Schmidhuber, 1997*).

### 3.4. Data Splitting Method used for Model Development:

[58] Robust model development requires using only part of the available data $D$ for model training, while reserving an independent portion for model evaluation. Adapting the methodology of *(Zheng et al., 2022)*, we split the available 20-year record into training $D_{Train}$ and evaluation $D_{Eval}$ subsets using year-based stratified allocation. Specifically, water-years were ranked and sorted by accumulated annual streamflow. The two most extreme (wettest and driest) were assigned to the training set and the next extreme pair to the evaluation set. After removing allocated years, this process was iterated, until all years were distributed between $D_{Train}$ and $D_{Eval}$ such that 60% of the year-pairs (12 WY) were allocated to $D_{\text{Train}}$ and the remaining 40% to $D_{\text{Eval}}$. This year-based stratification ensures that both splits contain a balanced representative range of the underlying distribution of hydroclimatic conditions, thereby minimizing the possibility of representational bias.

### 3.5. Hyperparameter and Training Procedure:

[59] Model development (training) for both physical-conceptual and ML-based models broadly followed *Wang and Gupta (2024a)*. To reduce sensitivity to unknown initial conditions, we applied a three-year spin-up by repeating the first water year (WY 2004) three times at the beginning of the simulation period; this helps limit adverse effects of incorrect state-initialization (*De la Fuente et al., 2023*).

[60] Training was implemented in PyTorch, and model parameters were optimized using ADAM (*Kingma, 2014*), with loss and gradients computed over the training subset. For each catchment, and for each method (and segment length) for computing the benchmark $\bar{B}^o$, we trained the model using 10 different random seeds (random weight initializations) for 1500 epochs with full-batch updates. The learning rate was set to $1.0 \times 10^{-1}$, and all trainable parameters were initialized to small random values drawn from a zero-mean standard normal distribution (i.e. $\theta = N(0, 1)$) to promote stable optimization at the start of training. The final model was selected as the seed that achieved best average loss across both $D_{Train}$ and $D_{Eval}$.

[61] Each model architecture was trained independently to learn the input-state-output dynamics of each catchment, enabling catchment-specific assessment of hydrological behavior under different benchmark conditions. Training was performed using multiple benchmark options within the $JKGE_{ss}$ framework, as

described earlier. This experimental setup facilitated systematic comparison of model performance and sensitivity across architectures with varying complexity and non-stationary benchmark scenarios.

## 4. Experimental Results

[62] We ran experiments using both the *Moving-Average Mean* and the *Section-Wise Mean* approaches to representing temporal non-stationarity of the catchment streamflow generating process. In section (4.1) we discuss results for the *Section-Wise Mean* approach. Results for the *Moving-Average Mean* approach are similar (while being computationally more expensive) and mentioned briefly in section (4.2).

[63] Sections (4.1 to 4.6) report results obtained using the five-state-variable $MA_5$, the most complex of the physical-conceptual models used for this study, the other two being ablated versions in which key hydrological process have not been included.

[64] Section 4.7 then explores whether model inadequacy can have a significant impact on our results and findings, by conducting tests using the simpler $MA_3$ (three-state-variable) and $MA_2$ (two-state-variable) model architectures.

[65] Section 4.8 explores whether the training benefits of accounting for temporal non-stationarity also extend to pure LSTM-based machine-learning models of catchment-scale hydrological functioning.

[66] Finally, Section 4.9 examines whether performing log-transformation to the data would have a significant impact on our findings.

### 4.1. Impacts of Using a Time-Nonstationary Section-Wise Mean

[67] To evaluate the impact of using a non-stationary benchmark defined using successive ***section-wise means***, we trained the model to optimize $JKGE_{ss}^{SA(N_s)}$ and examined the effect of different choices for section length ($N_s$). Model performance was assessed using a range of diagnostic plots to examine how choice of $N_s$ influences resulting model performance.

#### *4.1.1. Impact on the Flow Duration Curve*

[68] The graphical *Flow Duration Curve* (FDC) is widely used in hydrology to represent the overall distribution of streamflow magnitudes by showing the percentage of time that a particular flow rate is equaled or exceeded at a given site. It is constructed by ranking observed streamflow values in descending order on a log-scale and then plotting them against exceedance probabilities. Accordingly, it characterizes the variability and regime of streamflow for a basin (*Ridolfi et al., 2020*), and is useful for characterizing the full range of low, high and baseflow conditions. Unlike a hydrograph, the FDC does not preserve the chronological order of flows but instead focuses on the frequency distribution of discharge magnitudes.

[69] **Figure 7** shows how use of the non-stationary benchmark helps to improve FDC matching performance. Each row corresponds to a typical catchment selected from one of the five different hydro-climatic regimes (see Section 3.4.3) – results for all 15 catchments are included in the Supplementary Materials (**Figure S3**). Each column corresponds to use of $JKGE_{ss}^{SA(N_s)}$ with a different window size ($N_s = 365, 180, 90, 30, 7,$ and $1$ day). The observed (target) FDC's are shown using black-dashed-lines, while the simulated FDC's obtained by training using $JKGE_{ss}^{SA(N_s)}$ are shown using blue lines. For comparison, the simulated FDCs obtained by training with the standard $KGE_{ss}$ (that uses the LTM as benchmark) are shown using red lines. All of the FDC's shown are constructed using the entire 20 years of data, including portions used for both training and evaluation, to be as fully representative of the catchment as possible.

[70] Note, first, that the observed FDC's (black-dashed lines) for each region (rows in **Figure 7**) exhibit different distinguishing characteristic shapes (see **Figure S3** for all 15 catchments). The *rainfall-dominated* catchments in the western US (red) display steeper FDC slopes, particularly in the high-exceedance range, reflecting flashier hydrological responses and more variable streamflow associated with episodic rainfall

events. In contrast, *snowmelt-dominated* basins (black) generally show smoother and less steep FDC curves in the mid-range exceedance probabilities, indicating more sustained flows driven by gradual snowmelt release. The *historical rainfall-dominated* basins (green) exhibit intermediate behavior, with moderately sloped curves that indicate relatively balanced flow variability. Catchments in the eastern US influenced by *recent rainfall* patterns (blue) tend to show comparatively stable mid-range flows with smoother transitions across exceedance probabilities, suggesting more persistent baseflow contributions. Finally, *arid* catchments (yellow) exhibit steep FDCs with rapid decline from high flows to very low flows, followed by an extended low-flow tail at high exceedance probabilities. This pattern reflects infrequent but intense runoff events and prolonged periods of very low discharge, indicating limited catchment storage and weak baseflow contributions typical of intermittent streams.

[71] Next we see that, regardless of hydroclimatic regime, use of the standard $KGE_{ss}$ for training (red line) leads to systematic biases (both over- and under-estimation) in reproduction of various portions of the FDC, with a clear tendency to under-estimate at low flows (compare red to black-dashed lines); note that as a reference, the red curve remains the same for all subplots moving from left to right along a row.

[72] Finally, regardless of hydroclimatic regime, as the section length ($N_s$) is shortened, use of $JKGE_{ss}^{SA(N_s)}$ (blue line) leads to clear and systematic improvement in reproduction of the observed flow distributions (compare blue to black-dashed lines).

- For larger $N_s$ (365 & 180 days), the model trained using $JKGE_{ss}$ already shows comparable or slightly improved performance relative to the traditional $KGE_{ss}$ trained model for several catchments, particularly in *snow-dominated* regions (row 2; see also **Figure S3** in the supplementary materials). In these cases, improvement emerges quickly, indicating the simply accounting for seasonal non-stationarity can enhance the representation of flow variability.
- As $N_s$ is reduced further to 90 days, the simulated FDCs increasingly align with the observed ones across a wider range of exceedance probabilities. This improvement is evident across both high- and low-flow regimes, indicating that shorter $N_s$ enable better extraction of information from the data, resulting in the model being better able to adapt to seasonal and intra-annual variability. For some of the hydroclimatic regimes, this improvement is gradual, with incremental gains observed as $N_s$ is made shorter, highlighting differences in hydrological response and climate sensitivity across catchment types.
- By the time $N_s$ reaches 30 days, nearly all the catchments exhibit strong agreement between simulated and observed FDC's. At this time scale, use of the non-stationary benchmark effectively captures sub-seasonal dynamics, leading to improved reproduction of the full flow distribution regardless of catchment category. In contrast, when trained using $KGE_{ss}$, the model consistently under- or over-estimates portions of the FDC, particularly at low- and intermediate-flows.

[73] As the section length decreases below $N_s = 30$ to 7 or 1, models trained using $JKGE_{ss}$ continue to perform well for several catchments, particularly in regions characterized by relatively stable runoff generation (rows [1, 2]). However, performance deteriorates for other catchments, most notably in the "*arid*" hydroclimate (row 5), and in the "*recent rainfall dominated*" systems in the eastern US (row 4). In these regions, due to event-based runoff processes, the use of very short section length leads to slight instability in computation of the benchmark thereby reducing the effectiveness of the $JKGE_{ss}$ formulation.

[74] Moreover, it is worth noting that as the section length becomes substantially shorter, the focus shifts increasingly toward matching low-flows, with high-flows receiving comparatively less emphasis. This behavior is highlighted in **Figure 8** where we have zoomed in to show only the 0–5% exceedance probability range (highest flows). The blue line corresponds to $N_s = 30$ while the green line corresponds to $N_s = 1$, from which we see that high-flows are consistently underestimated across all five catchment

types when $N_s$ becomes too small. In contrast, for $N_s = 30$, the high-flow estimates are comparable to those obtained by training with $KGE_{ss}$ (using the stationary LTM as benchmark).

#### *4.1.2. Impact on Reproduction of Different Flow Ranges*

[75] To not rely solely on aggregate performance metrics, we further diagnose how accounting for temporal non-stationarity influences model performance across different parts of the flow regime (see also **Figure S4** in the supplementary material). **Figure 9** shows log-flow anomalies across five distinct flow groups for the catchments selected to represent the five hydroclimatic regimes, and compares models trained using $KGE_{ss}$ with those trained using $JKGE_{ss}^{SA(N_s)}$ defined over different values for $N_s$ .

[76] For each catchment, observed streamflow was divided into five flow groups (FG) based on quantiles of the observed flow distribution (*$FG_1$:0-20%, $FG_2$:20-40%, $FG_3$:40-60%, $FG_4$:60-80%, and $FG_5$:80-100%),* ensuring balanced representation across low-, intermediate-, and high-flow conditions. Anomalies were computed as difference between simulated and observed log-transformed flows (i.e., $log(sim) - log(obs)$), providing a scale-invariant measure of model bias and variability across the flow regime. For each flow group, anomalies were aggregated separately for the $KGE_{ss}$-trained model (red) and the $JKGE_{ss}^{SA(N_s)}$-trained models (blue).

[77] Boxplots summarize the distributions of anomalies within each flow group and training scenario, while the minimum (dashed lines) and maximum (solid lines) values are explicitly highlighted to emphasize changes in spread and extremal behavior. Shown at the top of each boxplot panel (each panel represents a flow group) is the observed log-flow range for the corresponding flow group (left-to-right subpanels correspond to low-to-high flow groups). The red horizontal lines for models trained with $KGE_{ss}$ are plotted across all section-length groups to facilitate direct visual comparison, while the blue lines show how anomaly ranges change systematically for models trained with $JKGE_{ss}^{SA(N_s)}$ as $N_s$ becomes smaller.

[78] For *low-flow groups*, we note the following:

- Across nearly all hydroclimatic regimes, (rows 1 through 5), training with $KGE_{ss}$ results in the lowest flow groups consistently exhibiting the largest negative anomalies, indicating systematic underestimation of low-flows. This behavior is persistent across catchment types and highlights the inability of a time-stationary benchmark to adequately represent low-flow dynamics, particularly in regimes with strong seasonality or intermittency. In contrast, as $N_s$ is reduced, training with $JKGE_{ss}^{SA(N_s)}$ substantially improves reproduction of the low-flow regime.
- For *snow dominated catchments*, noticeable reductions in anomaly magnitudes are already evident as $N_s$ becomes smaller 365 or 180 days (consistent with **Figure 7**).
- For *other regions*, particularly those with more complex hydrological responses, the improvements occur more gradually, and become pronounced only as section length approaches 30 days. For $N_s = 30$ days, the low-flow anomalies are markedly reduced and centered closer to zero across most catchments and flow groups.

[79] For *intermediate-flow groups*, we note a similar but weaker pattern:

- Training with $KGE_{ss}$ tends to produce biased anomalies, but with smaller magnitude and spread than for low flows. Meanwhile, training with $JKGE_{ss}^{SA(N_s)}$ primarily reduces variability rather than correcting the bias suggesting that, in the mid-flow-range, accounting for temporal non-stationarity acts mainly to improve model stability rather than fundamentally altering its representation.

[80] For the *high-flow groups*, we see that:

- Differences between $KGE_{ss}$ and $JKGE_{ss}^{SA(N_s)}$ are generally less pronounced. In several catchments, high-flow anomalies are already well constrained under $KGE_{ss}$, and reductions in section length yield only marginal additional improvement. However, $JKGE_{ss}^{SA(N_s)}$ still contributes to modest narrowing of anomaly ranges, indicating improved consistency in extreme flow simulation.

[81] For a few catchments (e.g., 11473900 and 2472000), training using $JKGE_{ss}^{SA(N_s)}$ with $N_s = 30$ results in slight performance degradation (based on visual analysis of the FDC's) in the highest-flow group; however, this degradation is small compared to the substantial improvements (pronounced reductions in bias and variability) obtained for the low- to medium-flow groups.

[82] Overall, **Figure 9** demonstrates that accounting for non-stationarity primarily improves low-to-intermediate-flow performance, but that care must be taken to ensure the section lengths are not made too short. Based on tests conducted on 15 catchments representing 5 hydroclimatic regimes across the CONUS, our judgement is that the optimal section length for daily catchment-scale hydrological modeling is $N_s \approx 30$ days.

#### *4.1.3. Impact on Reproduction of Streamflow Hydrographs*

[83] Based on the above findings, we further evaluated model performance by visual examination of the streamflow hydrographs. **Figure 10** compares observed and simulated time-series for representative wet and dry water years for each of the 5 study catchments (for all 15 catchments see **Figure S5** and **Figure S6** in the SM). Hydrograph comparisons are shown using both normal scale (to better show peak flows) and logarithmic scale (to better show low-to-intermediate flows and recession behavior). As before, black is used for observations, red for simulations obtained after training with $KGE_{ss}$ and blue for simulations obtained after training with $JKGE_{ss}^{SA(N_s)}$ with $N_s$ = 30 days.

[84] From **Figure 10a** (wet years), and **Figure 10b** (dry years) we note the following:

- Across all hydroclimatic regimes, training with $KGE_{ss}$ frequently results in under- or over-estimation of peak flows, while also showing larger deviations during non-peak conditions. In contrast, training with $JKGE_{ss}^{SA(30)}$ consistently results in more accurate reproduction of the timing and magnitude of high-flow events, while also improving performance during streamflow recessions.
- During dry years $JKGE_{ss}^{SA(30)}$ improvements are even more pronounced, with improved reproduction of timing and magnitude of low-flow sections, especially for certain catchments (e.g. 11473900, 2472000), and particularly for the *arid* catchment (13161500). In contrast, training with $KGE_{ss}$ results in larger deviations and frequent under- or over-estimation of flows.

[85] These results demonstrate that for wet years (and particularly under hydrologically extreme conditions), training with $JKGE_{ss}^{SA(30)}$ leads to more robust performance across the full range of flow conditions. When viewed using the logarithmic scale, improvements are particularly evident during sections dominated by persistent low-flows (at the beginnings of water years, and during recession phases). The gains are even more apparent during dry years. Meanwhile model skill at reproducing high-flow events is comparable for both metrics.

[86] Overall, this supports our hypothesis that accounting for temporal non-stationarity enables better identification of flow-regime-specific patterns.

### 4.2. Impacts of Using a Time-Nonstationary *Moving Average* Mean

[87] Testing showed that training using the *Moving-Average mean* $JKGE_{ss}^{MA(N_w)}$ approach did not yield additional benefits over use of the *Section-Wise mean* $JKGE_{ss}^{SA(N_s)}$ approach – overall model performance was very comparable (see **Figure S7** and discussion in the supplementary materials **Section S1**). As with

the *Section-Wise* approach, decreasing window size leads to systematic performance improvements, best performance is achieved with window size of around 31 days, and performance tends to degrade for shorter windows (7 days), particularly at high- and low-flow extremes. Importantly:

- For window sizes of approximately one month ($N_w = 31$ days and $N_s = 30$ days), the FDC's obtained using the two methods are virtually indistinguishable.

[88] So, while the *Moving-Average mean* approach might theoretically seem to be somewhat more elegant (providing a smoothly varying benchmark), it provides *essentially identical results with much higher computational cost* since the average must be computed at each and every time step (in the *section-wise* approach the section is computed only once for all time steps within a section, making it more computationally more efficient).

[89] Overall, these findings reinforce those of the *Section-Wise* approach, that computing the time varying benchmark using a monthly time scale for aggregation provides an optimal balance – it effectively smooths out shorter-term hydrologic variability while preserving relevant and meaningful information about overall flow dynamics.

### 4.3. Impacts of Also Accounting for Temporal Non-Stationarity in the Standard Deviation

[90] In general, for any conditional DGP, the (i) *mean*, (ii) *overall width* (here represented crudely by the standard deviation), (iii) *shape*, and (iv) *temporal autocorrelation structure* can all vary with time. However, the versions of $JKGE_{ss}$ tested so far were constructed to only account for non-stationarity in the *mean* of the conditional streamflow DGP and, after subtracting the time-nonstationary mean, the *standard deviation* of the resulting streamflow anomaly time series was *<u>assumed</u>* to be time-stationary (constant).

[91] Examination of the observed streamflow time-series reveals that this assumption is not supported by the data (see example for $N_w = 31$ days and $N_s = 30$ days shown in **Figures 11** for catchment 11473900). Whichever approach is used (*Section-Wise* shown in blue, or time-centered *Moving-Average* shown in green), the estimated *standard deviations* of the streamflow anomaly values (**Figures 11b**) obtained after subtracting the estimated streamflow *means* (**Figures 11a**) are non-stationary. Further, **Figure 11c** shows that the so-obtained standard deviations vary in direct proportion to the means, becoming increasingly aligned with the 1:1 reference line with progressive reductions in section length.

[92] So, assuming the standard deviation to be stationary is not theoretically correct and, in principle, one could treat all terms $M^*$, $V^*$ and $C^*$ of the $JKGE_{ss}$ metric as being temporally non-stationary. We therefore tested whether *<u>additionally</u>* accounting for time-varying streamflow anomaly *standard deviation* (along with time-varying mean) would result in improved performance. In other words, for any given section/window size, the resulting standard-deviation-augmented $JKGE_{ss}^{\mu\&\sigma}$ metric (see supplementary materials **Section S2**) is designed to require the trained model to reproduce the observed time-variation of *<u>both</u>* the mean and anomaly standard deviation.

[93] We found that:

- Doing so did not noticeably improve the results.
- Further, for arid catchments where extended sections of very low or zero flows can occur, the $JKGE_{ss}^{\mu\&\sigma}$ approach proved to be problematic – for shorter section lengths, the anomaly standard deviation can approach zero, leading to numerical instability during model training.

[94] These results should be treated as preliminary, as regularizing principles could be brought to bear to prevent the standard deviation estimates from becoming too small and to ensure numerical stability – we leave such investigation for future work. Meanwhile, our results suggest that, while it can be theoretically appealing to account for temporal non-stationarity in properties other than the mean of the streamflow generating process, the practical benefits may be somewhat limited.

### 4.4. How Accounting for Temporal Non-Stationarity During Model Training Affects the Computed Value of $KGE_{ss}$

[95] Our findings indicate that using $JKGE_{ss}$ for training (with appropriate choice of section size) mainly improves reproduction of low- and intermediate-level flows. We therefore conducted cross-metric training and evaluation experiments to examine the effect that training with $JKGE_{ss}$ (non-stationary benchmark) has on the reported value of $KGE_{ss}$ (stationary benchmark).

[96] First, the model was trained using $JKGE_{ss}$ and its performance was evaluated using $KGE_{ss}$ and its components, $M$, $V$, and $C$. Subsequently, the model was trained using $KGE_{ss}$ and evaluated using $JKGE_{ss}$ and its components $M^*$, $V^*$, and $C^*$. To account for sampling variability, bootstrap resampling was applied (e.g., see *Clark et al. (2021)*), wherein the streamflow data was resampled with replacement (using yearly blocks to preserve intra-annual temporal structure) and 1000 bootstrap samples were generated to obtain the distribution of each performance metric. The median value of each metric was then reported. Note that all metric values presented throughout the paper are produced using this approach.

[97] **Figure 12** (top four panels) compares $KGE_{ss}$ and its components obtained by training using $KGE_{ss}$ (x-axis; assuming stationary mean) versus when trained using $JKGE_{ss}$ (y-axis; assuming non-stationary mean). More specifically, instead of plotting $M$ and $V$ (whose optimal values are 0), we plot values for $M' = 1 - M$ and $V' = 1 - V$, so that the optimal values in all four subplots ($KGE_{ss}$, $M'$, $V'$ and $\rho$) are 1.0. Therefore, results that fall along the 1:1 line indicate essentially equivalent $KGE_{ss}$ performance under both training objectives. Different colors correspond to catchments from different hydroclimatic regimes.

[98] The results indicate clearly that:

- Training with $JKGE_{ss}$ *does not significantly impact* the values obtained for $KGE_{ss}$ and $\rho$. Both values are generally very high ($\approx 0.8$ or greater) for all 15 study catchments, regardless of hydroclimatic regime, and are similar regardless of which metric was used for model training.
- Component $M'$ (representing water balance) is almost perfect (close to 1.0) in both cases.
- Only *very minor* degradation is seen in component $V'$ (representing almost perfect performance in reproducing streamflow variability) when $JKGE_{ss}$ is used for training – notice that the plot axes range over only 0.95 to 1.0, and the maximum deviation is only ~0.01 (approximately 1%).

[99] **Figure 12** (bottom four panels) complements these results by comparing the values of $JKGE_{ss}$ and its components under the same conditions. Again, the x-axis shows results obtained without accounting for non-stationarity of the mean, while the y-axis show results obtained by accounting for non-stationarity of the mean, and we plot $M^{*\prime} = 1 - M^*$ and $V^{*\prime} = 1 - V^*$ so that the optimal values in all four subplots ($JKGE_{ss}$, $M^{*\prime}$, $V^{*\prime}$ and $\rho^*$) are 1.0.

[100] As might be expected, the results indicate that:

- Values obtained for $JKGE_{ss}$ are quite sensitive to the choice of training metric.
- When trained with $JKGE_{ss}$, the values obtained for $JKGE_{ss}$ at different catchments range from $\sim 0.60$ to $0.87$, but tend to be lower and occupy a much wider range ($-0.40$ to $0.75$) when trained with $KGE_{ss}$, with one of the recent rainfall dominated catchments (red dot; $JKGE_{ss} = -0.4 < 0$) performing worse than simply throwing away the model and using the observed time-varying section-wise mean as the predictor.
- As before, components $V^{*\prime}$ (related to anomaly variability) and $\rho^*$ (related to anomaly timing and shape) are relatively insensitive to choice of training metric.
- In contrast, the component $M^{*\prime}$ (related to *anomaly* water balance) is clearly sensitive to the choice of training metric, and therefore accounts for much of the variability seen in values obtained for $JKGE_{ss}$. Again, the worst performance (red dot not shown; $M^{*'} = -3.13$) is obtained for the aforementioned *recent rainfall dominated* catchment, but poor performance is also seen for one of the *arid* catchments (orange dot, $M^{*'} \sim 0.06$).

- Overall the declines in $JKGE_{ss}$ resulting from use of $KGE_{ss}$ for training (i.e., not accounting for temporal non-stationarity) appear to be mainly associated with poor ability of the $KGE_{ss}$-trained model to properly reproduce the water balance associated with <u>*streamflow anomalies*</u> obtained after subtracting the time-varying mean.

[101] We further conducted cross-metric evaluation for each catchment individually to quantify the changes in components of $JKGE_{ss}$ when models were trained using $JKGE_{ss}$ versus $KGE_{ss}$, and vice versa. Specifically, we examined the following eight quantities, for all of which a value of zero represents a perfect score: $KGEss', M, V, C, JKGEss', M^*, V^*,$ and $C^*$, where $KGEss' = 1 - KGEss$ and $JKGEss' = 1 - JKGEss$.

[102] In **Figure 13**, to enhance visual distinction of smaller (better) values, we actually plotted the *square root* of the value. Red circles indicate training with $KGE_{ss}$ and blue squares indicate training with $JKGE_{ss}$.

[103] The results indicate that:

- For almost all catchments, $JKGEss'$ and $M^*$ (anomaly water balance) deviate substantially more from the zero reference when not accounting for temporal non-stationarity (training with $KGE_{ss}$).
- In contrast, $V^*$ (anomaly variability) and $\rho^*$ (anomaly timing and shape) show very similar values under both training configurations, indicating limited sensitivity to the choice of training metric.
- In regards to $KGEss$ and its components, the results show that for <u>*certain*</u> catchments (e.g., 11523200, 4185000, and 3173000), performance in terms of $M$ (water balance) can be somewhat worse when training with the new $JKGEss$ metric, indicating some degree of "*trade-off*" in the two kinds of bias-related performance ($M$ and $M^*$).

## 5. Additional Experimental Results

[104] We further ran several additional experiments to obtain further insight into the behavioral properties of $JKGE_{ss}$. These include testing of *augmented* and *ablated* versions of the new metric, and evaluating the robustness of training performance under different levels of model structural inadequacy (model capacity) and also for purely data-based LSTM machine-learning models. Finally, we examine whether there is any benefit to log-transforming the streamflow before applying the metric.

### 5.1. Testing an Augmented Version of $JKGE_{ss}$

[105] To address the slight decline in $KGEss$ observed when training with $JKGEss$, and particularly the deterioration in long-term water balance component $M$ (albeit relatively small compared to the overall gains achieved by use of $JKGEss$), we examined what would happen if we augmented the $JKGEss$ formulation to explicitly also include the long-term-water balance component $M$; i.e., we defined $JKGE_{ss}^{Aug} = JKGE_{ss} - \sqrt{\frac{M}{2}}$ so that:

$$JKGE_{ss}^{Aug} = 1 - \sqrt{\frac{M+M^*+V^*+C^*}{2}} \qquad \text{(Eq. 11)}$$

which now also penalizes deviations of $M$ from zero. We then retrained the model for three study catchments (11523200, 4185000, and 3173000) where this degradation was relatively most pronounced.

[106] Interestingly, results shown in **Figure S8a** indicate that:

- Penalizing training for deviations of $M$ from zero leaves the original components of $JKGEss$ unaffected.

- We see significant improvements to both $KGEss$ and $M$ (its water balance component).

### 5.2. Testing Ablated Versions of $JKGE_{ss}$

[107] For completeness, we also tested two ablated versions of $JKGEss$ based on the premise that perhaps, once the temporal non-stationarity in the mean is properly accounted for, the variability and cross correlation terms $V^*$ and $R^*$ may lose statistical significance and could potentially be removed.

- First, we omitted only the correlation term $C^*$ and defined $JKGE_{ss}^{Abl:1} = 1 - \sqrt{M + M^* + V^*}$ wherein keeping the remaining terms. We then retrained the model for the same three study catchments (11523200, 4185000, and 3173000).
- Then we removed both $V^*$ and $C^*$ and defined $JKGE_{ss}^{Abl:2} = 1 - \sqrt{M + M^*}$ and trained the model on the same three catchments.

[108] Results shown in **Figure S8b** show clearly that:

- It is not advisable to ignore the $V^*$ and $C^*$ terms; i.e., accounting for temporal non-stationarity in the mean is not, by itself, sufficient to ensure good model performance – useful information is still provided by the $V^*$ and $C^*$ terms.
- The water balance components $M$ and $M^*$ remain nearly identical to those obtained when the model is trained using $JKGE_{ss}$ or $JKGE_{ss}^{Aug}$.
- However, the cross correlation and variability components $C$, $C^*$, $V$, and $V^*$ components deviate substantially from those obtained when the model is trained using $JKGE_{ss}$or $JKGE_{ss}^{Aug}$.

[109] Overall, this suggests that all four of the components included in $JKGE_{ss}^{Aug}$ (i.e., $M$, $M^*$, $V^*$, and $C^*$) contain valuable information, and therefore play significant roles when training the model.

### 5.3. Evaluating the Impacts of Model Architectural Simplification/Inadequacy on Training Performance

[110] To check whether the benefits of using $JKGE_{ss}$ instead of $KGE_{ss}$ persist when the model architecture is simplified to the point of being inadequate, we repeated our training experiments for the $MA_3$ (three-state-variable) model in which the two canopy storages have been excluded, and for the $MA_2$ (two-state-variable) model in which the baseflow storage has been further excluded.

[111] **Figure S9** in the supplementary materials shows how model performance varies across flow groups and from training metric to the other. The results indicate that:

- Across all catchments, flow groups, and model architectures, training with $JKGE_{ss}$ results in consistently better performance that training with $KGE_{ss}$.
- The improvement remains particularly pronounced for low-flow conditions, where the benefits are substantial.
- The performance difference between $JKGE_{ss}$ and $KGE_{ss}$ gradually becomes less as we transition from lower- to higher-flow groups.
- For the very high flows, use of $JKGE_{ss}$ results in a slight degradation of performance compared to $KGE_{ss}$ for some of the catchments (e.g., 2472000, and 11473900). However, the magnitudes of these degradations were relatively small compared to the overall gains achieved by use of $JKGE_{ss}$.

### 5.4. Evaluating the Impacts of Using $JKGE_{ss}$ for Training LSTM Network-Based Models

[112] To check whether the benefits of using $JKGE_{ss}$ instead of $KGE_{ss}$ also persist when training purely data-based models, we trained LSTM-based models with different numbers of hidden nodes (5, 10, and 15) for each study catchment – future work will explore the potential benefits of using $JKGE_{ss}$ rather than

$KGE_{ss}$, or $NSE$, to train continental scale LSTM-based models such as those reported by *Kratzert et al. (2018); Kratzert et al. (2019); Frame et al. (2022)*.

[113] **Figures S10** in the supplementary materials shows how model performance varies across flow groups and from training metric to the other. The results indicate that:

- Across all study catchments, regardless of the numbers of hidden nodes used in the LSTM models, performance achieved using $JKGE_{ss}$ is consistently better than that obtained using $KGE_{ss}$.
- As expected, LSTM model performance improves with increasing numbers of hidden nodes.
- Notable consequences of training with $JKGE_{ss}$ are the generally close-to-zero median values, reduced spread, and fewer extreme negative anomalies.
- In several of the flow groups, training with $JKGE_{ss}$ exhibits more stable performance, with narrower interquartile ranges and less variability across nodes, suggesting better robustness and generalization.
- As with the physical-conceptual models, gains in performance are markedly higher for the low-to-medium-flow groups, consistent with our previous inference that training with $JKGE_{ss}$ is particularly effective in improving the representation of low-flow dynamics, which are typically more difficult to model and more sensitive to bias and variance in the training objective.
- In contrast, the performance gains for high-flow groups are comparatively less, indicating that both $JKGE_{ss}$ and $KGE_{ss}$ perform similarly under high-flow conditions, where the signal-to-noise ratio is generally better.

[114] Of particular interest is that the performance gain obtained by accounting for temporal non-stationarity (training with $JKGE_{ss}$) is particularly significant in certain hydroclimatic regimes. Notably, these include the *recent rainfall-dominated catchments* in the western region, *snowmelt-dominated catchments*, and *arid* catchments, all of which are characterized by strong nonlinearity, seasonal shifts in dominant runoff processes, and pronounced low-flow or intermittent-flow conditions, which can be challenging for data-driven models to capture.

[115] The superior performance achieved in these settings indicates that use of $JKGE_{ss}$ helps to better constrain the training of data-based-models across a diversity of hydroclimatic regimes, which may lead to improved generalization under contrasting flow-generating processes.

### 5.5. Robustness of $JKGE_{ss}$ across different Models

[116] To evaluate the robustness of $JKGE_{ss}$ across different models, **Figure 14** compares FDCs of the physical-conceptual and LSTM-based models for five representative catchments from different hydro-climatic regimes. Columns 1–3 present FDCs obtained for the physical-conceptual models ($MA_2$, $MA_3$, and $MA_5$), whereas Columns 4–6 show results for the LSTM-based models (with nodes 5, 10, and 15). In each panel, the blue curve represents the FDC of simulations obtained using $JKGE_{ss}$, while the red curve represents simulations obtained using $KGE_{ss}$ and the dashed-black curve represents the data.

[117] The results indicate that:

- When trained using $KGE_{ss}$, the physical-conceptual models tend to <u>*underestimate*</u> low flows whereas the LSTM-based models generally <u>*overestimate*</u> low flows; this behavior is particularly evident in the *recent-rainfall dominated* catchments in the western and eastern United States, as well as in the *arid* catchments.
- In contrast, all models trained using $JKGE_{ss}$ reproduce the FDCs across all flow regimes more accurately.

[118] **Tables S1-S4** in the supplementary materials (also see **Section S3**) further compare statistical performance of the models across eight descriptors or metrics: $KGE_{ss}$, $M$, $V$, $C$, $JKGE_{ss}$, $M^*$, $V^*$, and $C^*$, computed from the model output streamflow for runs trained using either $KGE_{ss}$or $JKGE_{ss}$as the objective function.

- For both physical-conceptual and LSTM models, training with $JKGE_{ss}$ consistently leads to improved performance when evaluated in terms of $JKGE_{ss}$ and its component descriptors.
- In contrast, models trained with $KGE_{ss}$ show substantially lower $JKGE_{ss}$ values and noticeably larger component errors, particularly in the anomaly-bias-related term $M^*$, suggesting reduced consistency with respect to the generalized performance formulation.
- Further, the bootstrap ranges also indicate that the superiority of the $JKGE_{ss}$-trained models is consistent across most catchments, supporting the use of $JKGE_{ss}$ as a more robust metric for model training.

[119] Since improvement obtained by training with $JKGE_{ss}$ was found to be more pronounced in dry than wet years, we illustrate its impact on simulated hydrographs. **Figure 15** presents results for the three physical–conceptual models ($MA_2$, $MA_3$, and $MA_5$) and an LSTM model (LSTM-15) for two representative catchments: a wet catchment (11473900, **Figure 15a**) and an arid catchment (13161500, **Figure 15b**). In each figure, the left column shows streamflow on a linear scale, while the right column displays the same hydrographs with a log-transformed y-axis, allowing better visualization of low-flow conditions.

[120] For the *wet catchment* (11473900), both metrics generally capture the seasonal flow dynamics and major peak events. However, models trained with $JKGE_{ss}$ (blue line) tend to follow the observed hydrograph more closely than those trained with $\mathrm{KGE}_{ss}$ (red dashed line), particularly during the rising and recession limbs of peak flows. This improvement is more clearly visible in the log-transformed plots, where $JKGE_{ss}$ better reproduces intermediate- and low-flows while reducing systematic deviations observed in the $\mathrm{KGE}_{ss}$ simulations.

[121] For the *arid catchment* (13161500), the advantages of $JKGE_{ss}$ are even more evident. Simulations obtained with $\mathrm{KGE}_{ss}$ frequently overestimate peak flows and show exaggerated variability, whereas the $JKGE_{ss}$-trained models produce hydrographs that more closely track observed magnitude and timing of events. This difference is particularly noticeable during low-flow periods and recession phases, where $JKGE_{ss}$ maintains closer alignment with the observations in both the linear- and log-scale. Across all models ($MA_2$, $MA_3$, $MA_5$, and LSTM-15), the $JKGE_{ss}$ metric consistently yields simulations that better reproduce overall flow dynamics, peak magnitudes, and low-flow behavior compared to $KGE_{ss}$. Improvements are especially pronounced in the dry catchment, supporting earlier findings that $JKGE_{ss}$ provides greater benefits under low-flow conditions.

[122] Additional diagnostic plots were examined to evaluate the robustness and consistency of $JKGE_{ss}$ relative to $KGE_{ss}$ for the $MA_5$ (**Figure 16**) and LSTM-15 (**Figure 17**) models. These include monthly percent bias computed as $(\text{monthly sim} - \text{monthly obs})/\text{monthly obs} \times 100$ for the first eight water years, as well as quantile–quantile (QQ) plots (column 2), scatter plots of simulated versus observed streamflow (columns 3: $KGE_{ss}$, column 4: $JKGE_{ss}$), and plots of monthly bias (%) against monthly mean observed flow (column 5) computed using 20 years of daily streamflow data

[123] Across both models, $JKGE_{ss}$ consistently demonstrates improved performance over $KGE_{ss}$:

- The monthly bias time series shows that $JKGE_{ss}$ remains tightly centered around zero with relatively small variability, whereas $KGE_{ss}$ exhibits larger fluctuations and frequent extreme positive biases, indicating instability.

- The QQ plots further confirm that $JKGE_{ss}$ more closely follows the 1:1 line across the full range of flows, while $KGE_{ss}$ deviates notably, particularly at low flows (underestimates in $MA_3$ and overestimates in LSTM-15).
- Similarly, the scatter plots reveal that $JKGE_{ss}$ simulations are more tightly clustered along the 1:1 line, reflecting better agreement with observations and reduced dispersion compared to $KGE_{ss}$.
- The relationship between monthly bias and mean observed flow indicates that $JKGE_{ss}$ maintains low and consistent bias across different flow magnitudes, whereas $KGE_{ss}$ shows greater spread and a tendency toward under- or over-estimation, especially under low-flow conditions.
- These patterns are consistent for both $MA_3$ and LSTM-15, demonstrating that the superiority of $JKGE_{ss}$ is robust across different model structures and flow regimes.

[124] Overall, these results suggest that the benefits of using $JKGE_{ss}$ for training extend to both physical-conceptual and data-based models, resulting in enhanced model skill, particularly in regards to capturing flow variability and reducing systematic bias.

### 5.6. The Impacts of Log-Transforming Streamflow Before Applying the Training Metrics

[125] Finally, we repeated training of the $MA_5$ model architecture, but first applied a natural-log-transformation to the observed and simulated streamflow series before applying the methods tested in this paper; i.e., we computed $JKGE_{ss}$ and $KGE_{ss}$ using the log-transformed streamflows.

[126] Motivation for this experiment comes from our earlier analysis showing that training with $KGE_{ss}$ does not result in adequately representation of low-flow behaviors – i.e., the metric tends to predominantly emphasize reproduction of the high-magnitude flow events at the expense of low-flow values. The common argument in favor of log-transformation is that it enables the metric to better target the lower range of the hydrograph by stretching low-flow values while compressing high-flow values, thereby refocusing the emphasis toward low-flow dynamics (*Thirel et al., 2024*). However, some studies (*Santos et al., 2018*) have suggested avoiding log-transformation when computing $KGE$, as it may lead to numerical instability and biased performance evaluation.

[127] **Figure 18** compares FDCs for the five aforementioned catchments representative of each of the five hydroclimatic regions (see **Figure S11** in supplementary materials for results from all 15 study catchments). **Figure 18a** compares full FDCs for these catchments. In **Figure 18b** the exceedance probability (EP) axis has been restricted to the 0–5% range to highlight differences in peak-flow reproduction. The results indicate that:

- Across all five catchments, training with $KGE_{ss}^{Log}$ (i.e., $KGE_{ss}$ applied to log-transformed data) results in systematic underestimation of the high flows.
- In contrast, training with $JKGE_{ss}^{SA(30)}$ results in relatively good reproduction of the peak flows, closely following the observed FDC.

[128] Further, we did encounter issues when using the log-transformation with $JKGE_{ss}$. For some catchments doing so resulted in instability of the $JKGE_{ss}^{Log\ SA(30)}$ metric, due to requiring the calculation of ratios of segmental (section or window) means and standard deviations. When the flow data are log-transformed, the 30-day means and variabilities of <u>*observed*</u> discharge can frequently approach values close to zero during low-flow periods, causing the denominators in the ratios $b_t^s/b_t^o$ and $\psi_s/\psi_o$ to become very small so that even minor deviations in the simulated values become amplified. As a result, the metric can become highly sensitive to small fluctuations in low-flow conditions, leading to large gradients and unstable optimization behavior during training.

[129] More fundamentally, these results indicate that explicitly accounting for temporal non-stationarity in the data-generating process (DGP) during training is necessary so that the model is better able to reproduce the entire range of flow magnitudes. Doing so seems to alleviate the need to implement transformations to the data before computing the training metric.

# 6. Discussion, Conclusions and Future Work

## 6.1. Discussion and Conclusions

### *6.1.1. On $JKGE_{ss}$ and Benchmark Selection*

[130] This study has introduced a new version of the $KGE_{ss}$ metric, referred to as $JKGE_{ss}$, that enables the user to specify a "*benchmark*" time series different from the long-term mean (used in the original $KGE$, and in other metrics such as $NSE$), thereby altering the manner in which it mediates the extraction of information from target data. By varying the choice of benchmark, the user can examine the effects of such choice on model development, and on resulting model performance.

[131] The choice of what benchmark to use will, of course, depend on user goals – for example *Garrick et al. (1978)* and *Schaefli and Gupta (2007)* discuss the use of benchmarks that depend on climate-related seasonality (using the interannual mean value for every calendar day). Benchmarks can also, in principle, be based on seeking to improve on signature properties of the target data (*Gupta et al., 2008*), or on improving the reproduction of different flow groups (*Yilmaz et al., 2008*), among many other possibilities.

[132] This study takes a different approach to benchmark selection, by noting that previous suggestions have all implicitly assumed that the underlying conditional (on driving meteorological variables) DGP is time-stationary. However, our simple analysis (Section 1.4) of hydrometeorological and streamflow data for catchments across the CONUS suggests that seasonal, interannual and longer term temporal variability is sufficiently pronounced that the assumption of time-stationarity is not justified. Further, the long-term mean of observed streamflow serves as a rather weak benchmark against which to evaluate model performance, and there is little to suggest that any of the statistical properties of the conditional catchment-scale streamflow DGP (evaluated over whatever time scale) are actually time-stationary.

[133] It seems sensible, therefore, to adopt an approach to model development and evaluation that *does not* depend strongly on the assumption (implicit or explicit) of stationarity in the underlying DGP, and to instead let the data "*speak for themselves*". Accordingly, the $JKGE_{ss}$ metric is explicitly designed to account for potential dynamical non-stationarity in the statistical properties of the DGP. And since the dynamics of temporal variation of those statistical properties (mean, standard deviation, auto-correlation structure, etc.) are not well understood, use of the metric involves specification of a *hyper-parameter* (the moving average window size $N_w$, or section length $N_s$ as appropriate) that can be *tuned* by the user while observing the impacts on information extraction and resulting model performance during model development/training.

[134] Our experimental results clearly demonstrate considerable benefit to explicitly accounting for data non-stationarity during model development. Tested over a range of model complexities, from the relatively simple two-state physical-conceptual $MA_2$ model, to the intermediate complexity five-state physical-conceptual $MA_5$ model, to the purely data based 15-cell-state LSTM(15) model, use of $JKGE_{ss}$ for model training clearly improved reproduction of all aspects of the long-term flow duration curve (a stationary benchmark signature property of the catchment), while enabling the model to properly track temporally-varying conditional dynamics of high-, medium- and (in particular) low-flow and recession portions of the hydrograph. Further, this performance improvement was consistently obtained on all 15 study catchments representing 5 different hydroclimatic regions of the US, ranging from humid recent-precipitation-dominated to snow-dominated to strongly arid.

#### *6.1.2. On the Causes of Non-Stationarity Considered*

[135] The modeling strategy pursued in this paper explicitly assumes that the "*catchment properties*" *do not* vary with time, and that the conditioning causes of temporal non-stationarity in catchment response (streamflow) are temporal-non-stationarities in the forcing variables (precipitation, temperature, etc.). This is reflected by the catchment being modeled, in the conventional manner, using model architectures and parameter values that are fixed (do not vary with time), so that temporal changes to catchment structure and material and geometric properties are not considered.

[136] To be clear, this is *not* a limitation in the design/application of the proposed $JKGE_{ss}$ metric, which can (in principle) also be applied in the context of modeling strategies that allow for context-dependent-time-varying model architecture and/or parameter values (e.g., conditionally depending on hydro-climatic conditions or other factors such as human intervention). Since the compounded effects of time-varying catchment properties and hydrometeorological forcings will appear together as temporal non-stationarity in the catchment response, the version of $JKGE_{ss}$ metric explored here (based on an empirically estimated non-stationary benchmark time series that varies contextually with the target over the simulation period) will still continue to be applicable.

#### *6.1.3. On Performance under Model Structural Inadequacy*

[137] For brevity of exposition, most of the experimental results reported in this manuscript used the five-state physical-conceptual $MA_5$ model architecture but, as was demonstrated, the performance gains associated with accounting for temporal non-stationarity via the $JKGE_{ss}$ seem to hold regardless of model structural adequacy, and are even realized for the purely data-based LSTM model.

#### *6.1.4. On the Choice of Window Size*

[138] We found no significant advantage to using the more computationally-expensive continuously-varying *Moving-Average Mean* rather than the stepwise-constant *Section-Wise Mean* for representing the non-stationary benchmark time series.

[139] By progressively reducing the window/section-length hyperparameter from $\infty$ (when $JKGE_{ss}$ effectively becomes $KGE_{ss}$) to $365\ days$ (annual), and so on down to $7\ days$ (weekly), we showed that the marginal distribution $p(Z^{AL})$ of standardized anomalies of log-transformed streamflow become increasingly well-behaved and symmetrical, enhancing the ability of the benchmark to capture the full range of hydrologic variability in the data, and leading to distributions that are better suited for robust statistical characterization. Doing so resulted in progressive improvements to reproduction of streamflow hydrograph dynamics over the full range of flow levels and across wet, medium and dry years, and particularly to hydrograph recessions and intermediate-to-low flows (that are characteristically not well reproduced when using $KGE_{ss}$ or $NSE$, or by any metric that does not account for non-stationarity in the DGP). This occurred *while continuing to maintain* good performance in the reproduction of high-flows and hydrograph peaks. Improvements were particularly pronounced during dry years and in catchments with persistent low-flows, especially in *arid* systems.

[140] Overall, our results indicate that a window/section-length of $\sim 30\ days$ (monthly) provides an excellent balance of performance over the entire range of flows, while accurately tracking temporally-varying dynamics of streamflow at all time-scales from annual to daily. While further investigations into how best to specify this hyperparameter need to be conducted, our results suggest that model performance may deteriorate in event-driven systems (such as *arid* and *rainfall-dominated* catchments) if this hyperparameter is made too small (e.g., $7-1\ days$ for this case of daily time-scale modeling), due to statistical instability associated with estimation of the benchmark trajectories.

#### 6.1.5. *On Cross-Metric Evaluation*

[141] Further, training using $JKGE_{ss}$ with $N_s \sim 30\ days$ also maintains excellent values for the original $KGE_{ss}$ metric itself (and also its diagnostic components representing water balance, variability and temporal cross correlation). Cross-metric evaluation showed that $JKGE_{ss}$ acts to remove systematic biases in the reproduction of *anomaly water balance* ($M^*$) that can occur when training with the original $KGE_{ss}$, while the components related to anomaly timing and variability ($V^*$, $\rho^*$) were largely insensitive to metric choice. These results suggest that it is not sufficient to simply account for the *long-term water balance* (i.e., to reproduce the long-term runoff ratio) – information extraction from the data (and hence model performance) can also benefit significantly from improved representation of the *time-varying water balance* (i.e., the time-varying runoff ratio).

#### 6.1.6. *Recommendation on Version of $JKGE_{ss}$ to be Used*

[142] This hypothesis is supported by experiments incorporating an additional term into the $JKGE_{ss}$ to ensure reproduction of *both* the long-term water balance and the time-varying water-balance. Doing so largely mitigated the marginal deterioration in $KGE_{ss}$ (and its diagnostic components) observed when using $JKGE_{ss}$ for model training. This result held up for all catchments across the full range of hydroclimatic regions tested (*humid* to *arid*). Analysis of the log-flow showed that performance improvements are primarily due to substantial reductions in low-flow bias and variability. Accordingly, the version of $JKGE_{ss}$ that we recommend as the starting point for further investigation is the augmented version $JKGE_{ss}^{Aug(N_s)}$ reported via Equation 11 (see Section 4.5) with $N_s = 30\ days$. The step-by-step pseudo-code for implementing this approach is provided in **Appendix A**.

#### 6.1.7. *On Robustness of the Proposed Formulation of $JKGE_{ss}$*

[143] Since non-stationarity is a property of the entire DGP, and not simply related to the mean, we further tested a modified version of $JKGE_{ss}$ that also accounts for temporal variations in the standard deviations (via $M^*$) of the streamflow anomalies – i.e., deviations of streamflow from the (stationary or non-stationary) mean. These experiments resulted in no noticeable improvement to model performance, while causing numerical instability in the cases of *arid* catchments with near-zero flows. Overall, these results suggest that, while accounting for non-stationarity in the standard deviation (and/or other statistical properties of the data generating process) may be theoretically appealing, doing so may offer little if any practical benefit for spatially-lumped daily-timescale catchment modeling. Further, for robust implementation, regularization strategies would be needed.

[144] Of course, in other modeling contexts, it is highly possible that accounting for other aspects of temporal non-stationarity (such as variations in width, shape and autocorrelation structure of the anomaly distributions) could be beneficial, and we strongly recommend deeper investigation of this issue across a variety of geoscientific (and other) modeling contexts and applications.

[145] We also tested ablated versions of $JKGE_{ss}$ in which the terms $V^*$ and $C^*$ that correspond to anomaly variability and/or cross-correlation strength were removed, leaving only an emphasis on reproducing the time-varying mean. Ablation resulted in clear performance degradation to reproduction of anomaly variability and/or cross-correlation strength, indicating that all the components of $JKGE_{ss}$ provide valuable, complementary information and are important for robust model training.

[146] Finally, we also tested what happens when $KGE_{ss}$ and $JKGE_{ss}$ are applied to log-transformed streamflow. While this commonly used strategy does improve low flow performance, it was found to result in systematic underestimation of high flows. In contrast, training with $JKGE_{ss}$ (without applying the log-transform) consistently and accurately reproduces the full range of low-to-high flows. Again, these results support the hypothesis that it is necessary to explicitly account for temporal non-stationarity in the

streamflow-generating process and that instead simply transforming the data cannot fully resolve biases in the reproduction of streamflow hydrographs.

#### *6.1.8. Summary*

[147] In summary, our results indicate that the benefits of accounting for temporal non-stationarity in the catchment-scale data generating process (by adopting the newly proposed $JKGE_{ss}$ metric for model training) are robust, and persist over different levels of physical-conceptual model adequacy and for a wide range of hydroclimatic conditions. Even purely data-based LSTM models trained with $JKGE_{ss}$ consistently outperformed those trained with $KGE_{ss}$, showing reduced bias, narrower anomaly spreads, and fewer extreme errors, with gains being most pronounced for low-to-intermediate-flow regimes, while maintaining high-flow performance. Overall, as might be expected, the most pronounced improvements were seen in hydroclimatic regimes that are characterized by strong nonlinearity or seasonal flow shifts.

### 6.2. Future Work

[148] This work has demonstrated the potential power of explicitly accounting for temporal non-stationarity when developing models for geoscientific applications. More generally, it enables the user to assess the value of various benchmarking schemes to improve information extraction from the data.

[149] While we have primarily focused on using a benchmark that recognizes and accounts for temporal non-stationarity, benchmarks could alternatively be defined in terms of hydrological signatures or other descriptors that can be applied to better characterize the system – examples include using flow groups based on quantiles of the streamflow distribution, partitioning the hydrograph into its driven (rising limb) and non-driven (falling limb) components (see *Boyle et al. 2000*), and/or incorporating descriptors based on the rates of change of the hydrographs rather than their absolute magnitudes.

[150] An even more sophisticated approach might involve simultaneously training a separate (machine-learning based) model to generate a benchmark, which could then be used within the generalized theoretical framework of the $JKGE_{ss}$. Such strategies may offer promising avenues for further improving the process of geoscientific model development.

[151] As always, we solicit and encourage constructive comments and debate on these and related aspects of geoscientific model development in the service of advancing scientific knowledge.

## Appendix A: Pseudo-Code for Computation of the $JKGE_{ss}^{Aug(N_s)}$ Metric

**Inputs:**

Observed time series: $O = \{o_t\}_{t=1}^{N}$

Simulated time series: $S = \{s_t\}_{t=1}^{N}$

Section length: $N_s = 30$

**Output:**

$JKGE_{ss}^{Aug(N_s)}$

**Step 1: Construct section-wise benchmark time series ($\mathrm{b_t}$)**

Divide each time series into consecutive non-overlapping sections of length $N_s$. Let $G(t)$ denote the set of indices belonging to the section that contains time step $t$. For each $t = 1, \ldots, N$:

$$b_t^o = \frac{1}{N_s} \sum_{i \in G(t)} o_i$$

$$b_t^s = \frac{1}{N_s} \sum_{i \in G(t)} s_i$$

The resulting benchmark time series $\{b_t^o\}$ and $\{b_t^s\}$ have the same length $N$ as the original time series, with values constant within each section.

**Step 2: Compute benchmark-centered anomalies ($a_t$)**

For each $t = 1, \dots, N$:

$$a_t^o = o_t - b_t^o$$

$$a_t^s = s_t - b_t^s$$

The anomaly series also have length $N$.

**Step 3: Compute anomaly variability ($\psi_t$)**

$$\psi_o = \sqrt{\frac{1}{N} \sum_{t=1}^{N} (a_t^o)^2}$$

$$\psi_s = \sqrt{\frac{1}{N} \sum_{t=1}^{N} (a_t^s)^2}$$

**Step 4: Compute anomaly cross-correlation**

$$\rho_{so}^* = \frac{1}{N} \sum_{t=1}^{N} \left(\frac{a_t^s}{\psi_s}\right)\left(\frac{a_t^o}{\psi_o}\right)$$

**Step 5: Compute anomaly variability ratio**

$$\alpha_{so}^* = \frac{\psi_s}{\psi_o}$$

**Step 6: Compute benchmark matching component**

$$M^* = \frac{1}{N} \sum_{t=1}^{N} \left(1 - \frac{b_t^s}{b_t^o}\right)^2$$

**Step 7: Compute anomaly variability and correlation components**

$$V^* = (1 - \alpha_{so}^*)^2$$

$$C^* = (1 - \rho_{so}^*)^2$$

**Step 8: Compute long-term water balance component**

$$\mu_o = \frac{1}{N} \sum_{t=1}^{N} o_t$$

$$\mu_s = \frac{1}{N} \sum_{t=1}^{N} s_t$$

$$\beta_{so} = \frac{\boldsymbol{\mu_s}}{\boldsymbol{\mu_o}}$$

$$M = (1 - \beta_{so})^2$$

**Step 9: Compute $JKGE_{ss}^{Aug(N_s)}$**

$$JKGE_{ss}^{Aug(N_s)} = 1 - \sqrt{\frac{M + M^* + V^* + C^*}{2}}$$

**Notes:**
The benchmark is piecewise constant with length $N$. Identical sections must be used for both the observed and simulated series.

## Open Research

The data used in this study are freely available online:

## Acknowledgments

Funding for this project was provided by the National Oceanic and Atmospheric Administration (NOAA), awarded to the Cooperative Institute for Research on Hydrology (CIROH) through the NOAA Cooperative Agreement with The University of Alabama, NA22NWS4320003/A25-0353-S006. We gratefully acknowledge *Peter A. Troch* for generously providing access to his HPC account, enabling the computational analyses presented in this work. We also thank *Jasper Vrugt* for his detailed review and valuable insights. The second author (HVG) acknowledges support for a 4-month research visit to the Karlsruhe Institute of Technology, Germany, provided by the KIT International Excellence Fellowship Award program, and is grateful for the inspiration and encouragement provided by members of the Information Theory in the Geosciences group (geoinfotheory.org), and by *Uwe Ehret*, *Ralf Loritz*, *Erwin Zehe*, *Anneli Guthke*, *Manuel Alvarez-Chavez* and several others during his visits to Germany over the past several years.

## Tables

**Table 1:** Physical properties of the selected CAMELS catchments, including catchment ID, site location, geographic coordinates, and hydroclimatic category.

| ID | Name | Latitude | Longitude | Area ($km^2$) | Criteria |
|---|---|---|---|---|---|
| 11523200 | Trinity River above Coffee Creek, CA | 41.11 | -122.71 | 382.94 | |
| 11473900 | Middle Fork Eel River, near Dos Rios, CA | 39.71 | -123.33 | 1925.01 | Recent rainfall-dominated (West) |
| 14222500 | East Fork Lewis River, near Heisson, WA | 45.81 | -122.29 | 329.06 | |
| 9223000 | Hams Fork below Pole Creek, near Frontier, WY | 42.11 | -110.71 | 333.15 | |
| 9035900 | South Fork of Williams Fork, near Leal, CO | 39.80 | -106.03 | 72.84 | Snowmelt-dominated |
| 12358500 | Middle Fork Flathead River near West Glacier, MT | 48.33 | -113.53 | 2917.03 | |
| 2472000 | Leaf River, near Collins, MS | 31.71 | -89.41 | 1927.13 | |
| 4185000 | Tiffin River at Stryker, OH | 41.74 | -84.34 | 1064.23 | Historical rainfall-dominated |
| 3331500 | Tippecanoe River near Ora, IN | 41.16 | -86.56 | 2248.65 | |
| 3173000 | Walker Creek at Bane, VA | 37.27 | -80.71 | 773.32 | |
| 1539000 | Fishing Creek, near Bloomsburg, PA | 41.08 | -76.43 | 701.78 | Recent rainfall-dominated (East) |
| 1667500 | Rapidan River near Culpeper, VA | 38.36 | -78.27 | 1210.77 | |
| 9505800 | West Clear Creek, near Camp Verde AZ | 34.60 | -111.43 | 624.73 | |
| 8324000 | Jemez River near Jemez, NM | 35.89 | -106.67 | 1222.22 | Arid Catchment |
| 13161500 | Bruneau River at Rowland, NV | 41.70 | -115.56 | 990.36 | |

## Figures

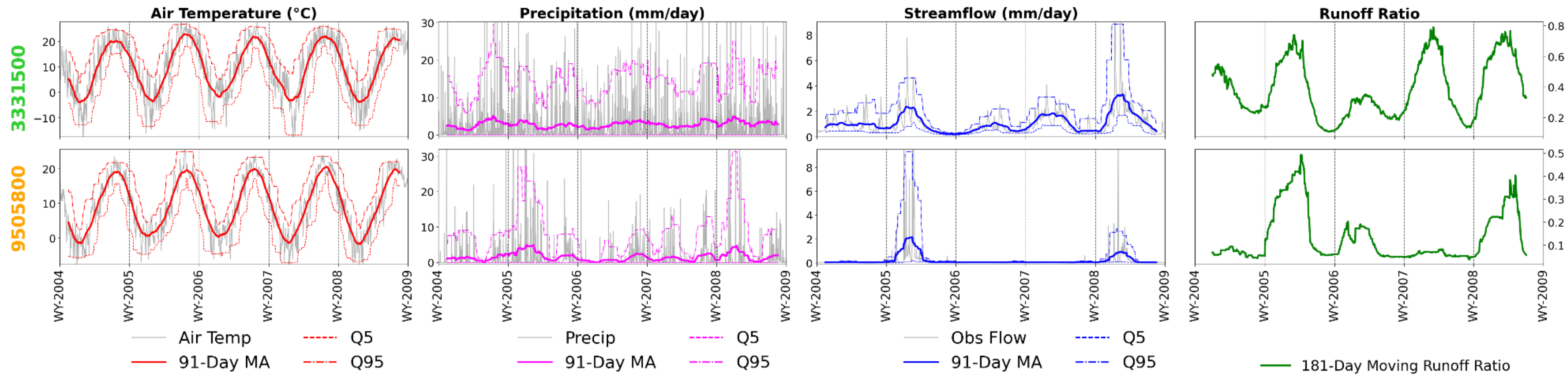

**Figure 1:** Subplots illustrate temporal non-stationarity of the moving-average mean and moving quantile (Q5%, Q95%) bounds for air temperature, precipitation and streamflow response, for two representative catchments; (3331500) is from a *historical-rainfall-dominated* regime and (9505800) is *arid*. Also shown is the resulting time-varying runoff ratio.

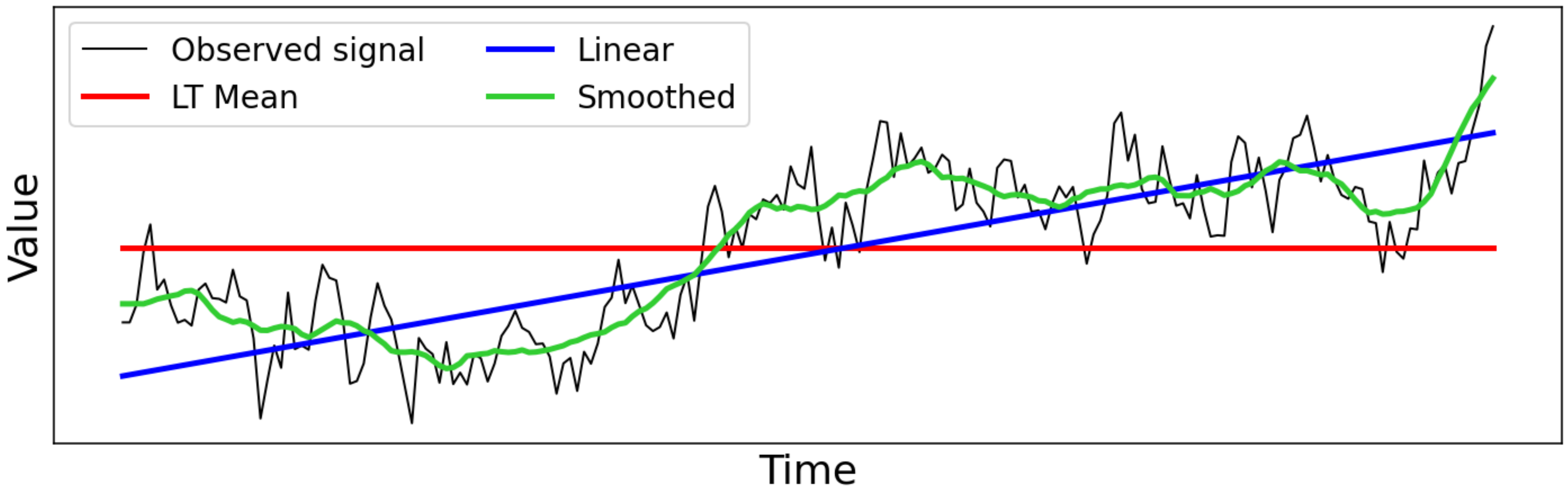

**Figure 2:** A cartoon illustrating some potential benchmarks that could be used in the context of a given signal (black). Shown as the Long-Term Mean (red), a Linear Trend (blue) and a smoothed Trend extracted from the data (green).

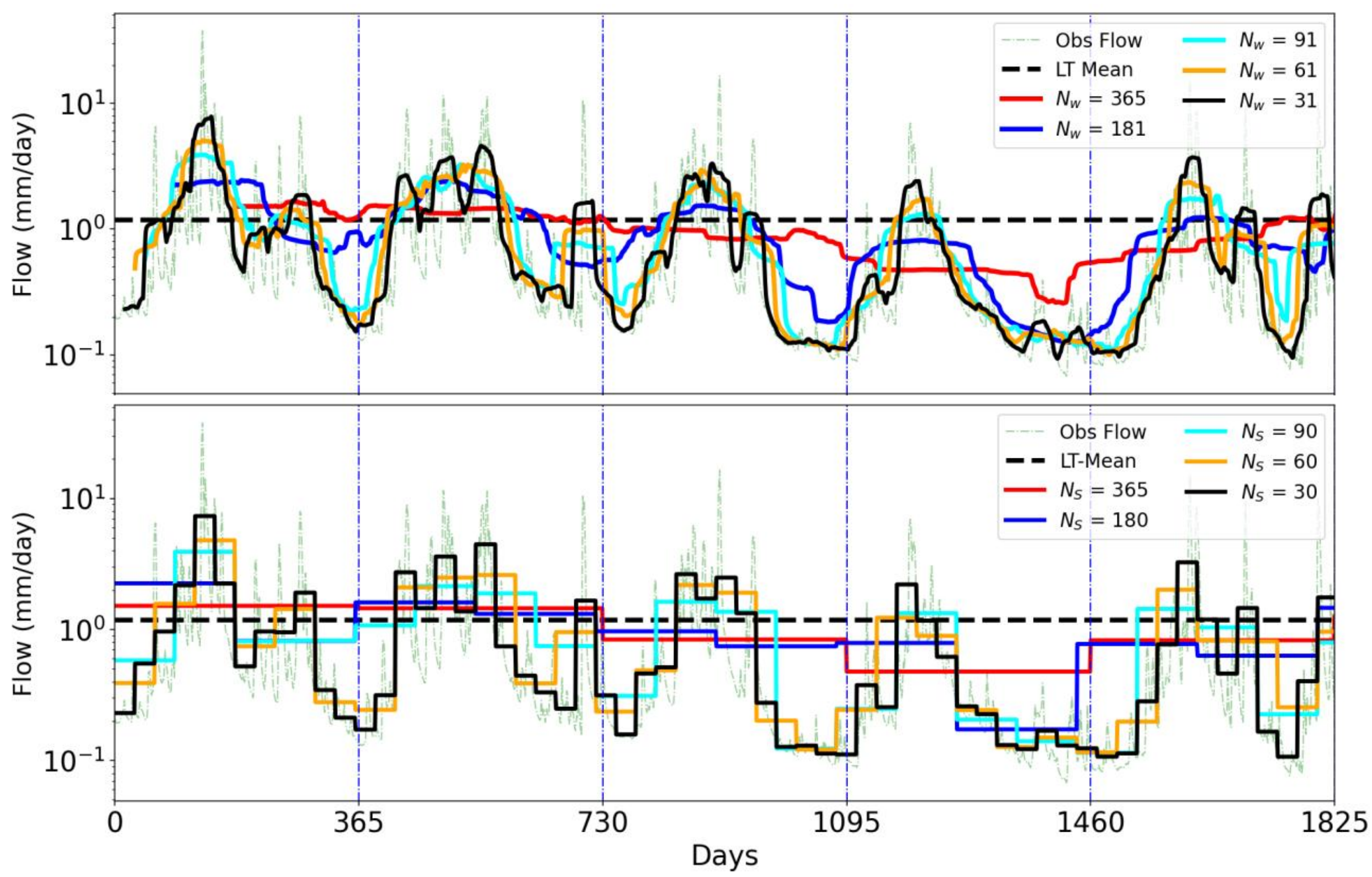

**Figure 3:** Illustration of various potential non-stationary benchmark time-series over five water years extracted from observed USGS streamflow (catchment 247200) at various time-scales. The data starts on October 1, 2003. (a) Shows application of the moving-average mean approach with odd-integer window lengths of 365, 181, 91, 61, and 31 days. (b) Shows application of a fixed sectional-mean approach with section lengths of 365, 180, 90, 61, and 30 days. The black-dashed line represents the stationary long-term-mean.

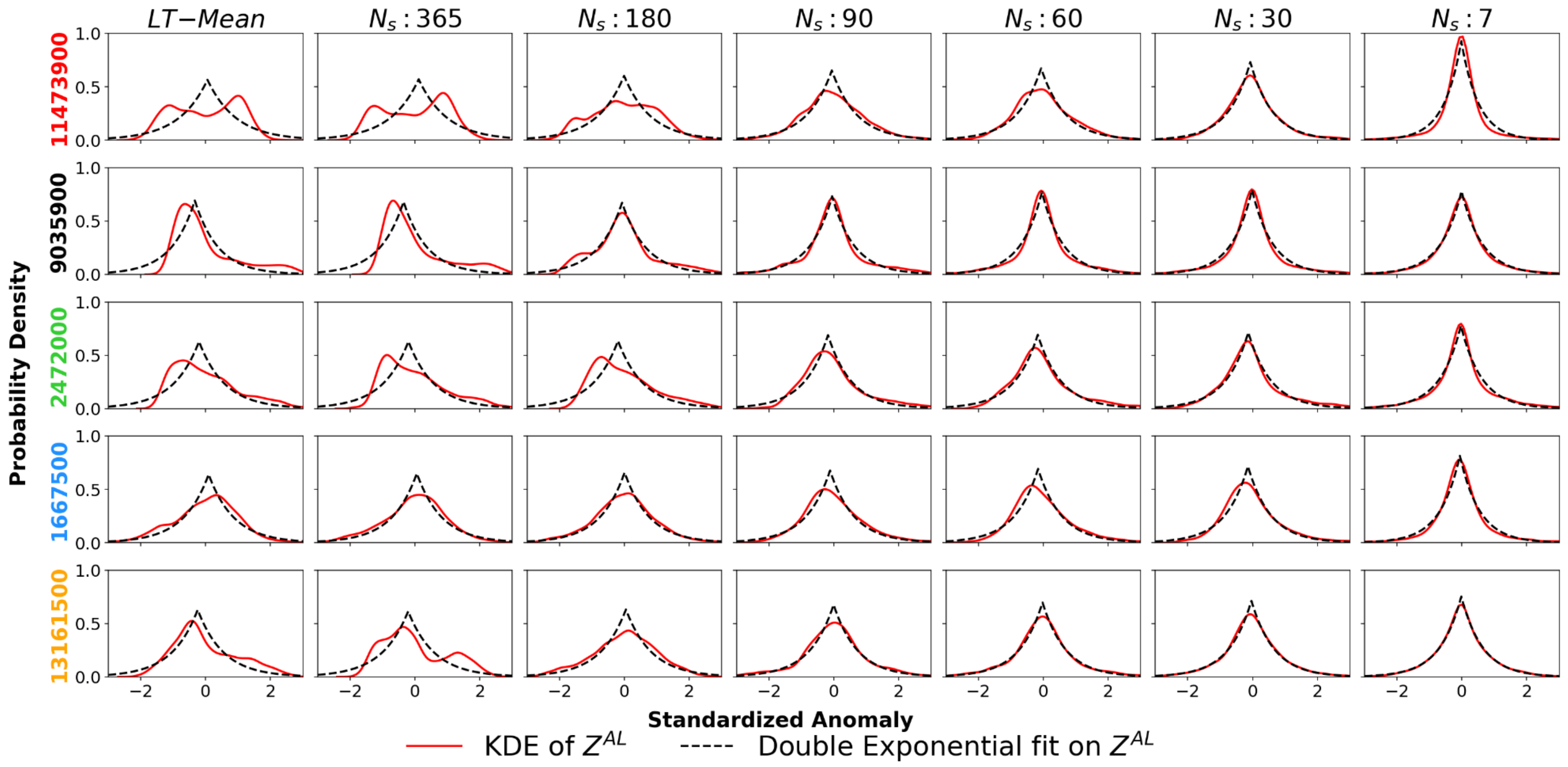

**Figure 4:** Distributions of standardized anomalies ($Z^{AL}$) for five representative catchments, one from each hydroclimatic zone, computed using the SA approach using different choices for the benchmark. Window lengths ($N_s$) progressively decrease from left to right, starting from the long-term mean (LT-Mean) to $N_s = 7$ .The kernel density estimates (KDEs) become increasingly well-behaved as $N_s$ decreases, ultimately showing close agreement with the double-exponential distribution (black dashed line)

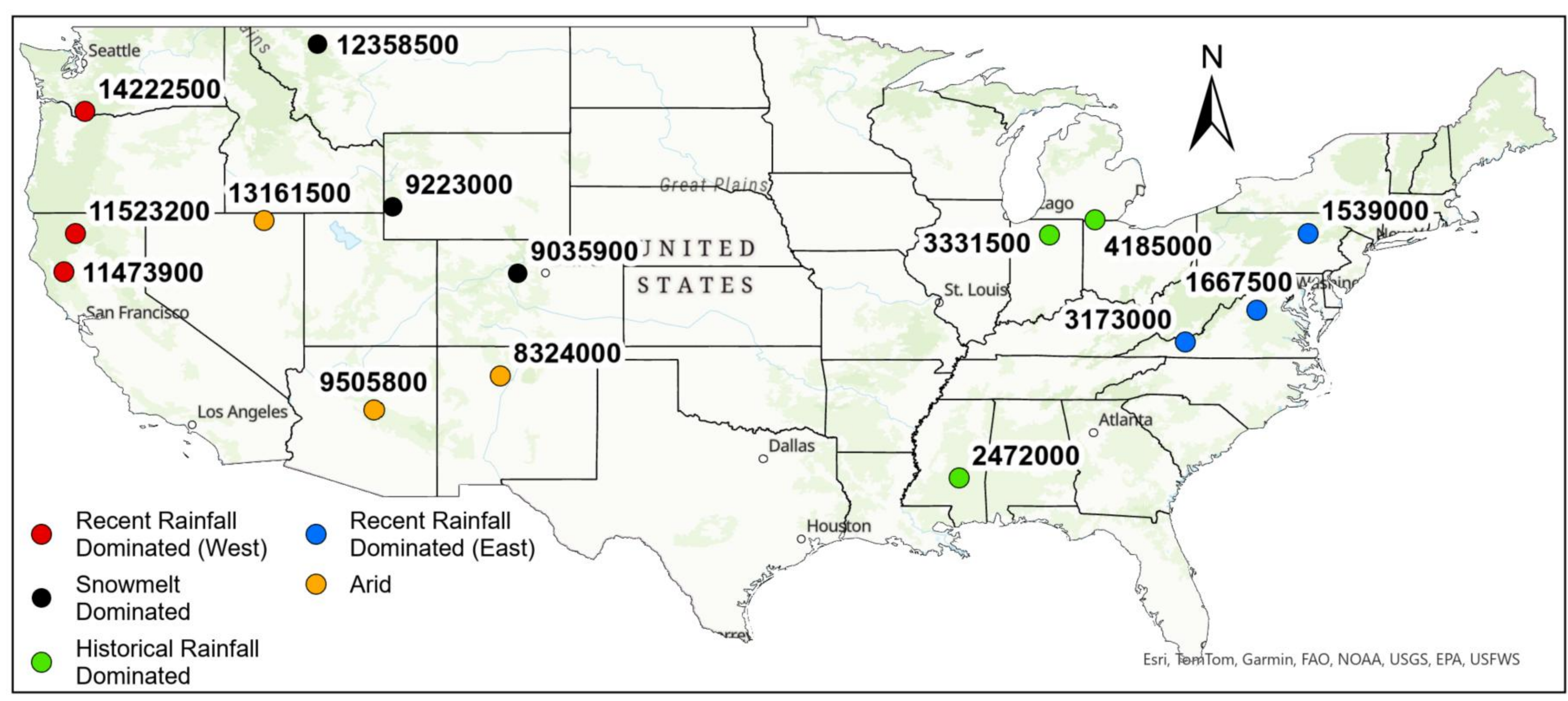

**Figure 5:** Map showing geographical distribution of the 15 selected CAMELS catchments, with colors indicating their hydroclimatic classification.

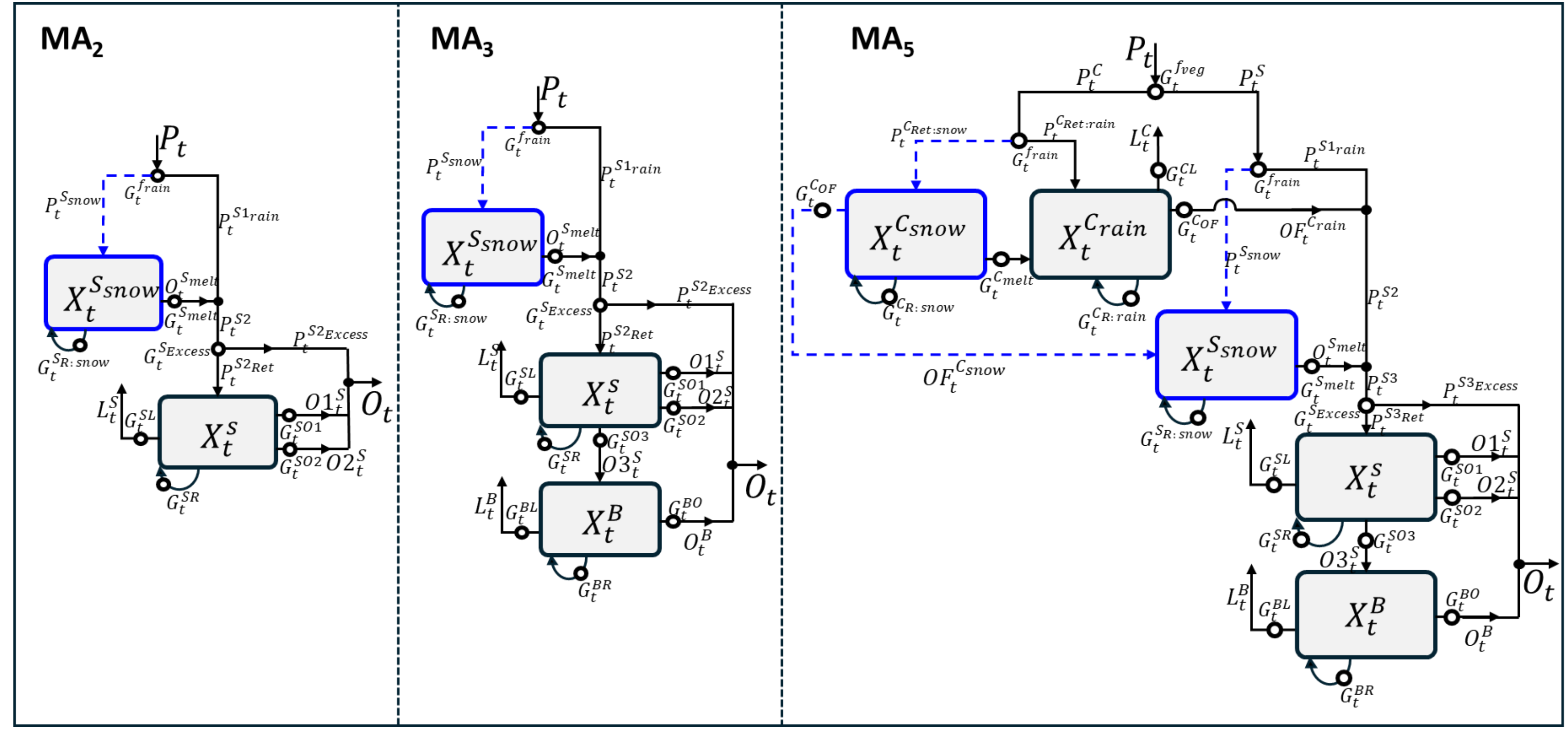

**Figure 6:** The physical-conceptual model architectures MA$_2$ (2), MA$_3$ (3), and MA$_5$ (5) used in this study, exhibiting different levels of complexity, characterized by number of state variables.

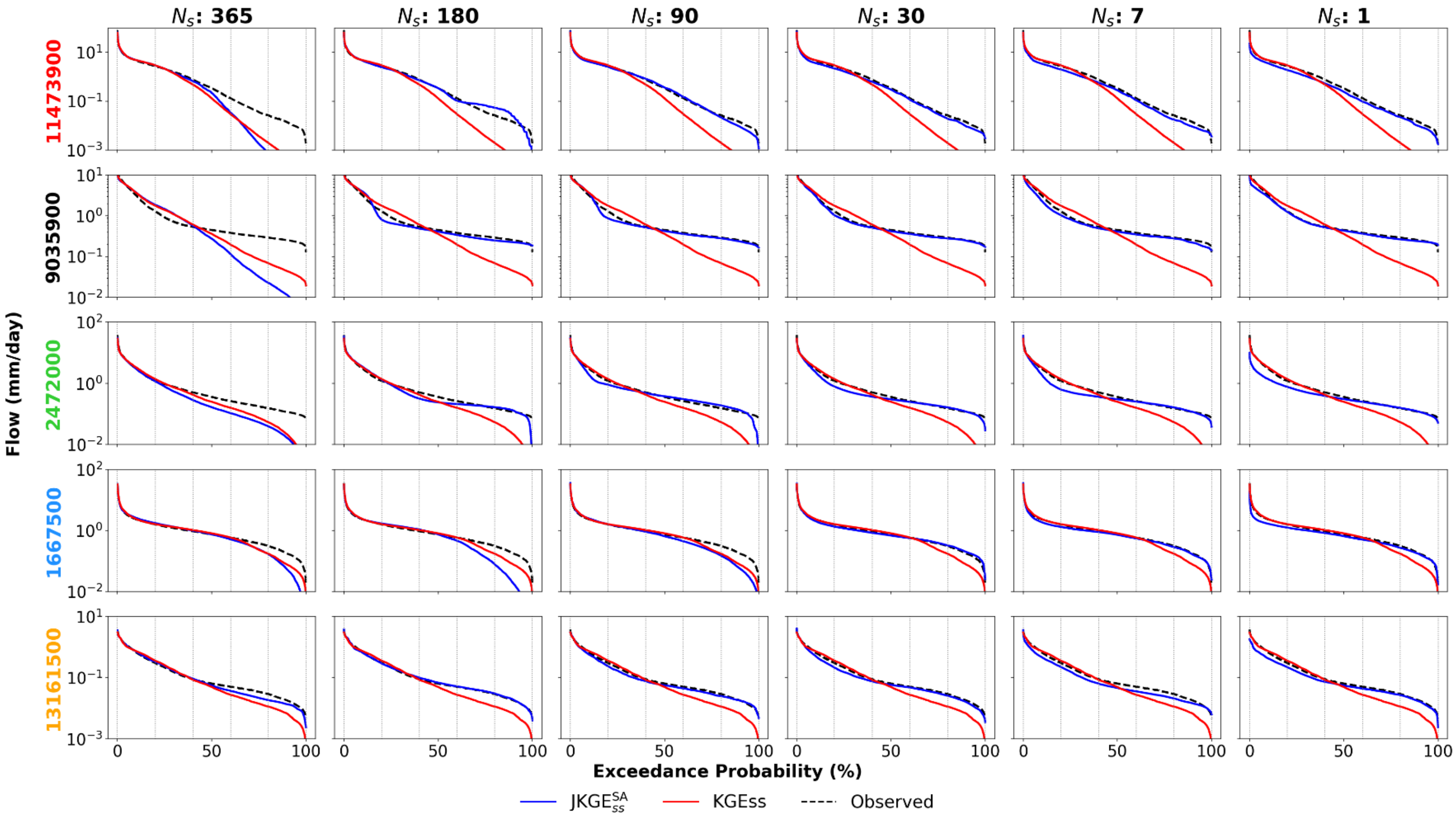

**Figure 7:** Performance of model $MA_5$ in terms of reproducing observed *Flow Duration Curves* (dashed black line) for five representative catchments (one from each hydroclimatic regime). Red shows results of training with $KGE_{ss}$ and blue shows results of training with $JKGE_{ss}$ with progressively section lengths (Ns) from left to right. $JKGE_{ss}$-based performance improves as $N_s$ is decreased, with $N_s = 30$ resulting in best performance across all catchments. Performance degrades for shorter $N_s$ (7 or 1), especially at high and medium flows. Note that the red $KGE_{ss}$ curves shown for reference for each catchment remain the same across all columns.

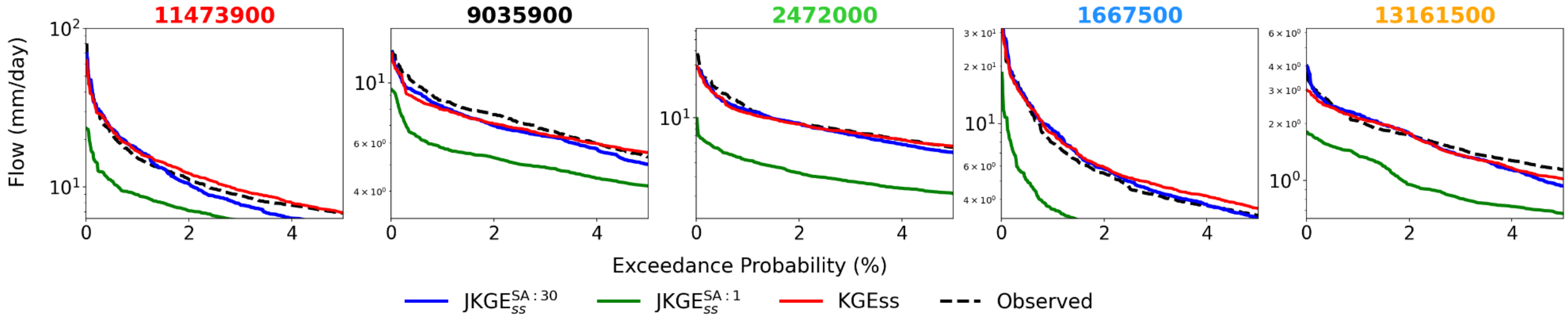

**Figure 8:** Illustration of the poor performance achieved for high flows (0–5% exceedance probability) when training with $JKGE_{ss}$ with $N_s = 1$ (green) compared to $N_s = 30$ (blue). $KGE_{ss}$ results (red) are shown for reference. At high flows, $JKGE_{ss}$ ($N_s$ = 30) and $KGE_{ss}$ perform equally well regardless of hydroclimatic regime.

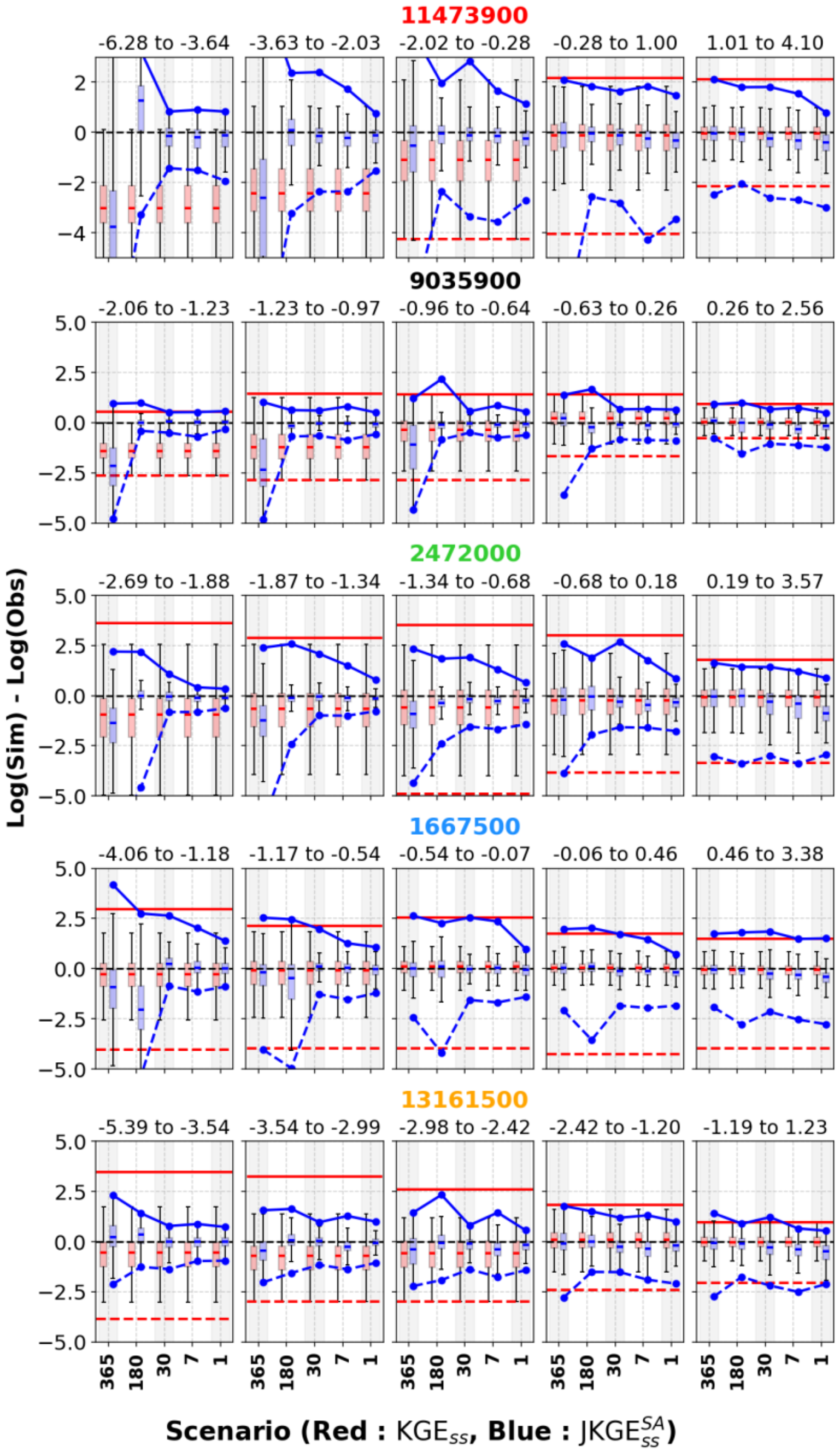

**Figure 9:** Log-flow anomalies across five flow groups (FG1–FG5) for representative catchments. Models trained with $KGE_{ss}$ (red) and $JKGE_{ss}^{SA(N_s)}$ (blue) across different section lengths ($N_s$) are compared. Anomalies, computed as $log\ (sim) - log\ (obs)$ ,are shown as boxplots along with minimum (dashed lines) and maximum (solid lines) values. Training with $JKGE_{ss}^{SA(N_s)}$ improves low- and-intermediate-flow performance as $N_s$ decreases, with $N_s \approx 30$ days yielding the most balanced representation across flow regimes. High-flow anomalies are generally well captured by both metrics. Red horizontal lines show results of training with $KGE_{ss}$ for comparison, while blue lines illustrate changes in anomaly range with decreasing $N_s$.

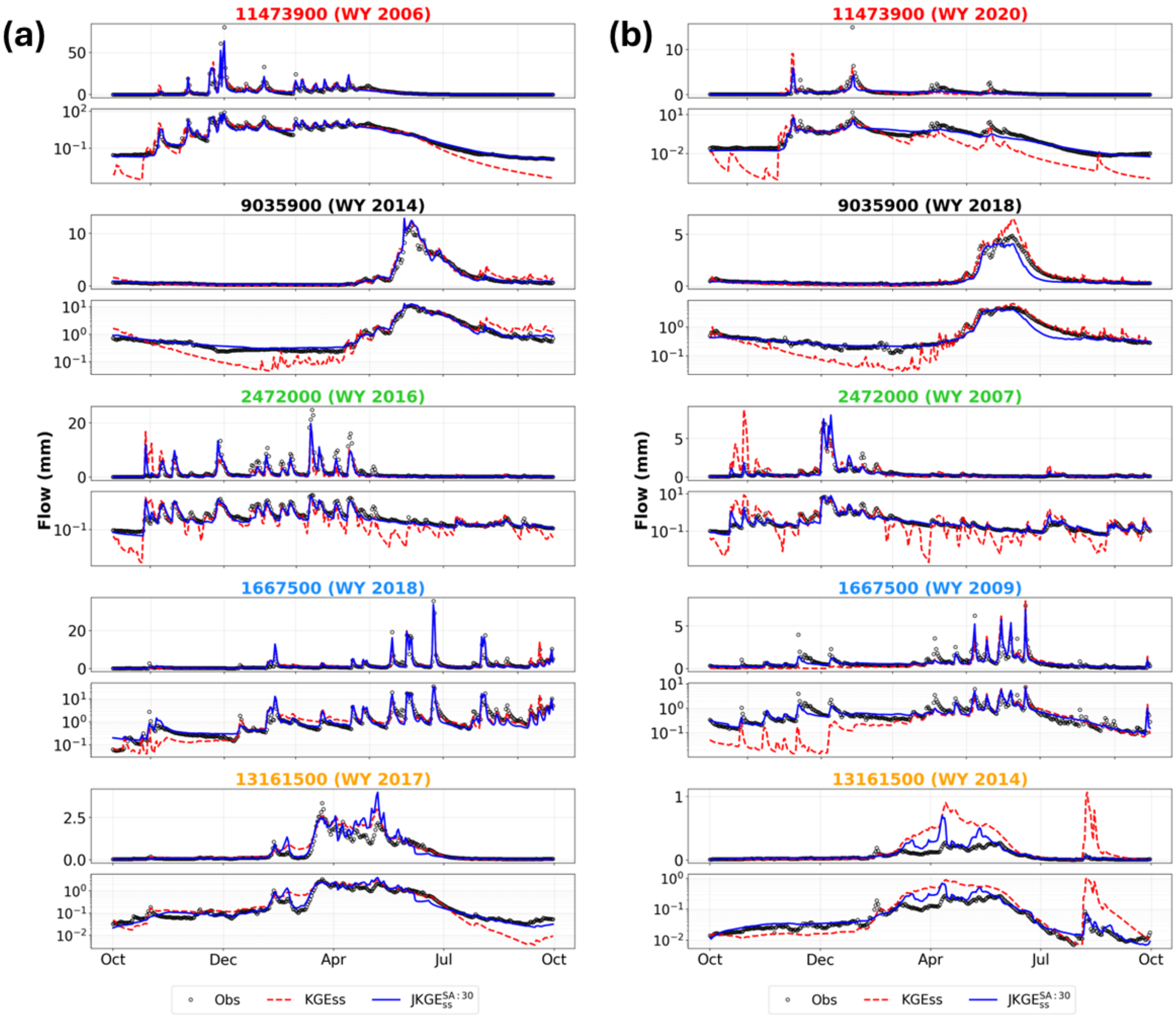

**Figure 10:** Natural- and log-scale observed (black) and simulated streamflow hydrographs (mm/day) for representative wet (a) and dry (b) years across five catchments using. Results are shown in red for $KGE_{ss}$ and blue for $JKGE_{ss}^{SA(30)}$. Training with $JKGE_{ss}^{SA(30)}$ results in better reproduction of both high- and low-flow dynamics, including recession behavior, with improvements being most pronounced during dry years.

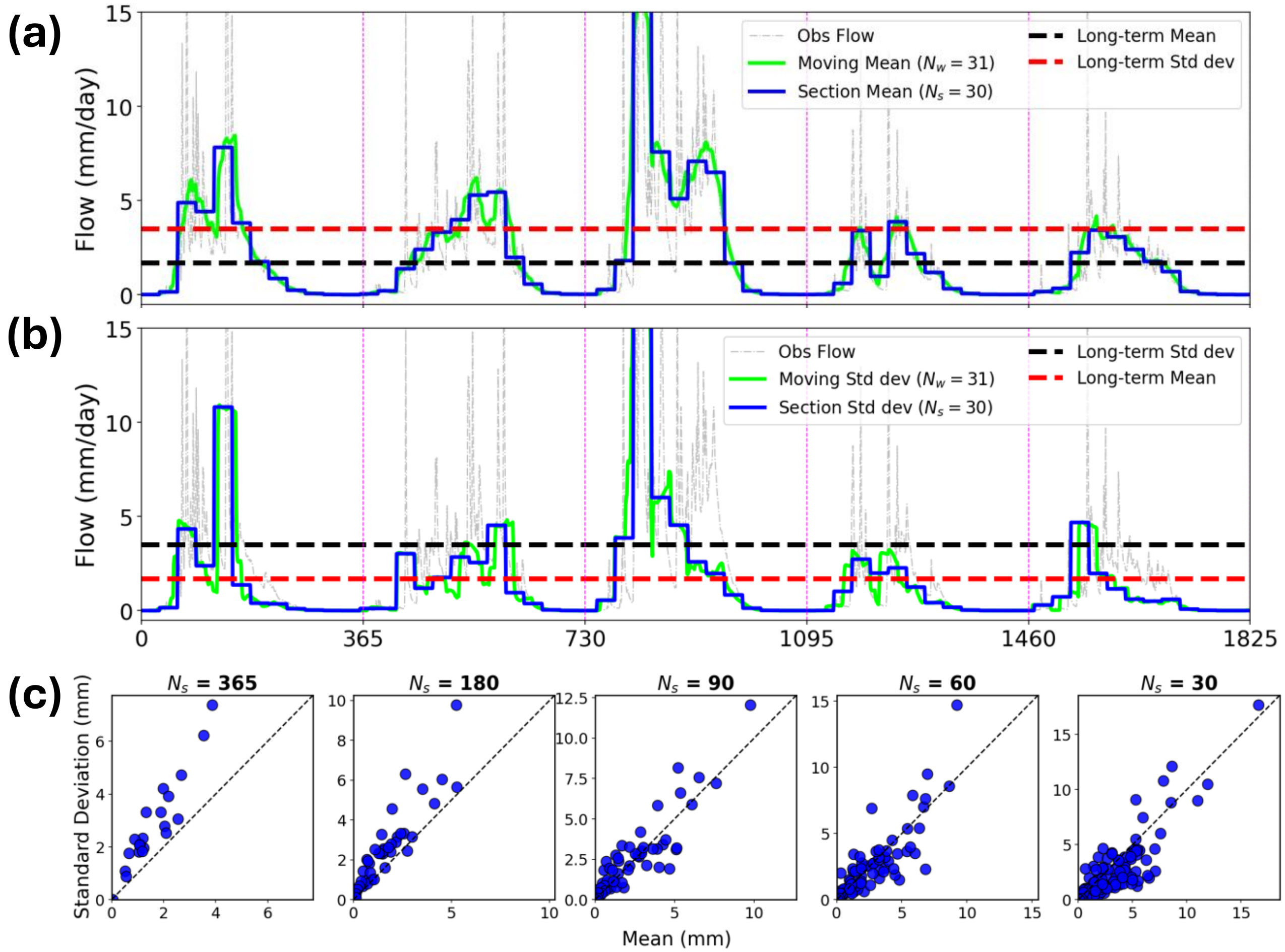

**Figure 11:** Illustration of the non-stationarity of estimated streamflow statistical properties for catchment 11473900 ($N_w = 31$ days, $N_s = 30$ days). (a) Estimated streamflow means, (b) estimated standard deviations of anomalies, and (c) scatterplot showing the relationship between the two. Both *Section-Wise* (blue) and *time-centered Moving-Average* (green) approaches are shown. The relationship pf non-stationary standard deviations to means increasingly aligns with the 1:1 reference line as section length decreases.

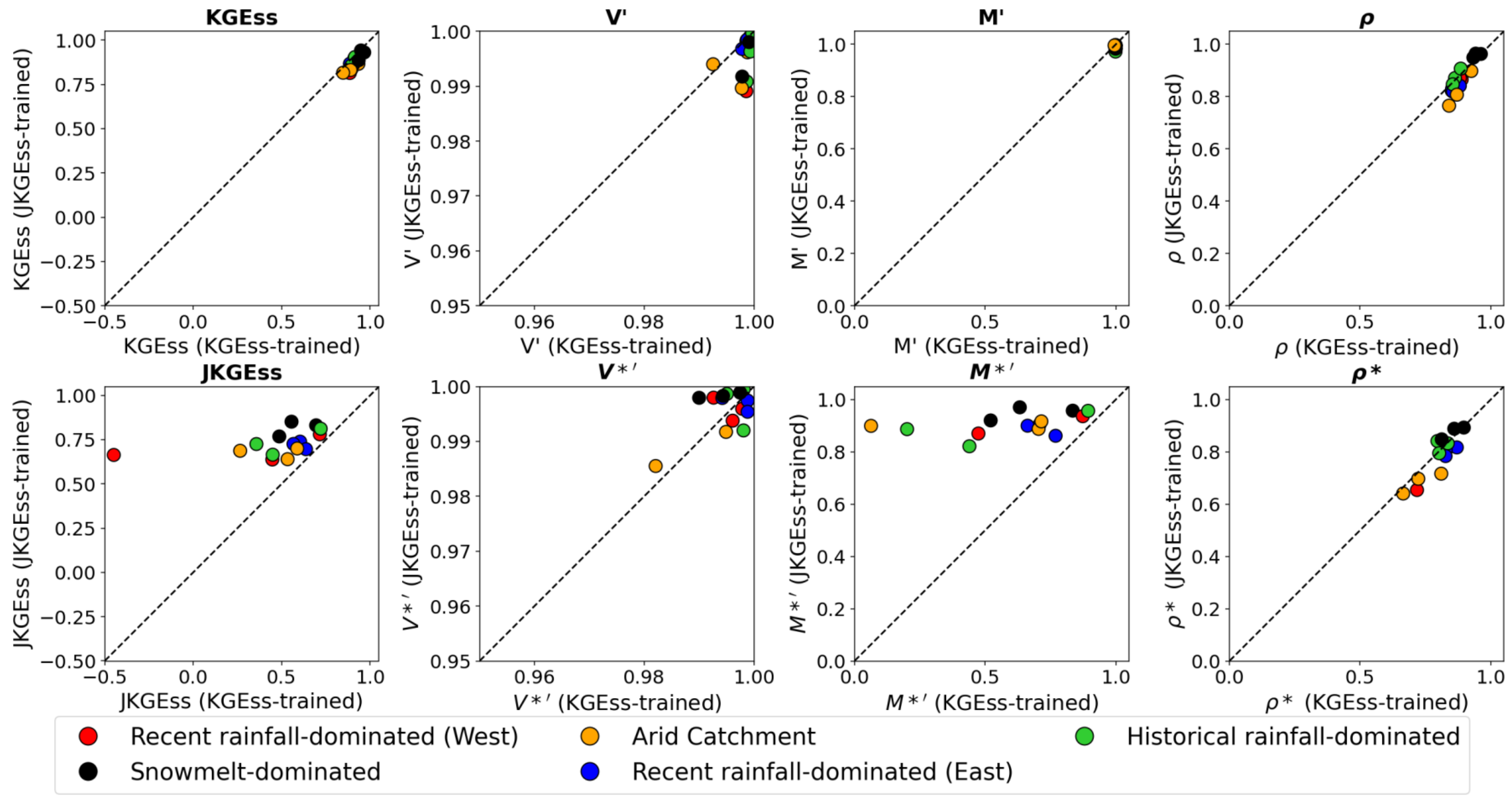

**Figure 12:** Cross-metric comparison of $KGE_{ss}$ and $\mathrm{JKGE}_{ss}$ values and their components. Top panels compare values obtained for $KGE_{ss}$ and its components (shown as $V^{'} = 1 - V$ , $M^{'} = 1 - M$ ,and $\rho$ ) when models are trained with $\mathrm{KGE}_{ss}$ (x-axis) versus trained with $JKGE_{ss}$ (y-axis). Bottom panels compare corresponding values obtained for $JKGE_{ss}$ and its components (shown as $V^{*'} = 1 - V^*$ ,$M^{*'} = 1 - M^*$, and $\rho^*$). The results show that $KGE_{ss}$ values are largely unaffected by training with $JKGE_{ss}$, while $JKGE_{ss}$ values and the anomaly water-balance component $M^{*'}$ are strongly affected by training with $KGE_{ss}$.

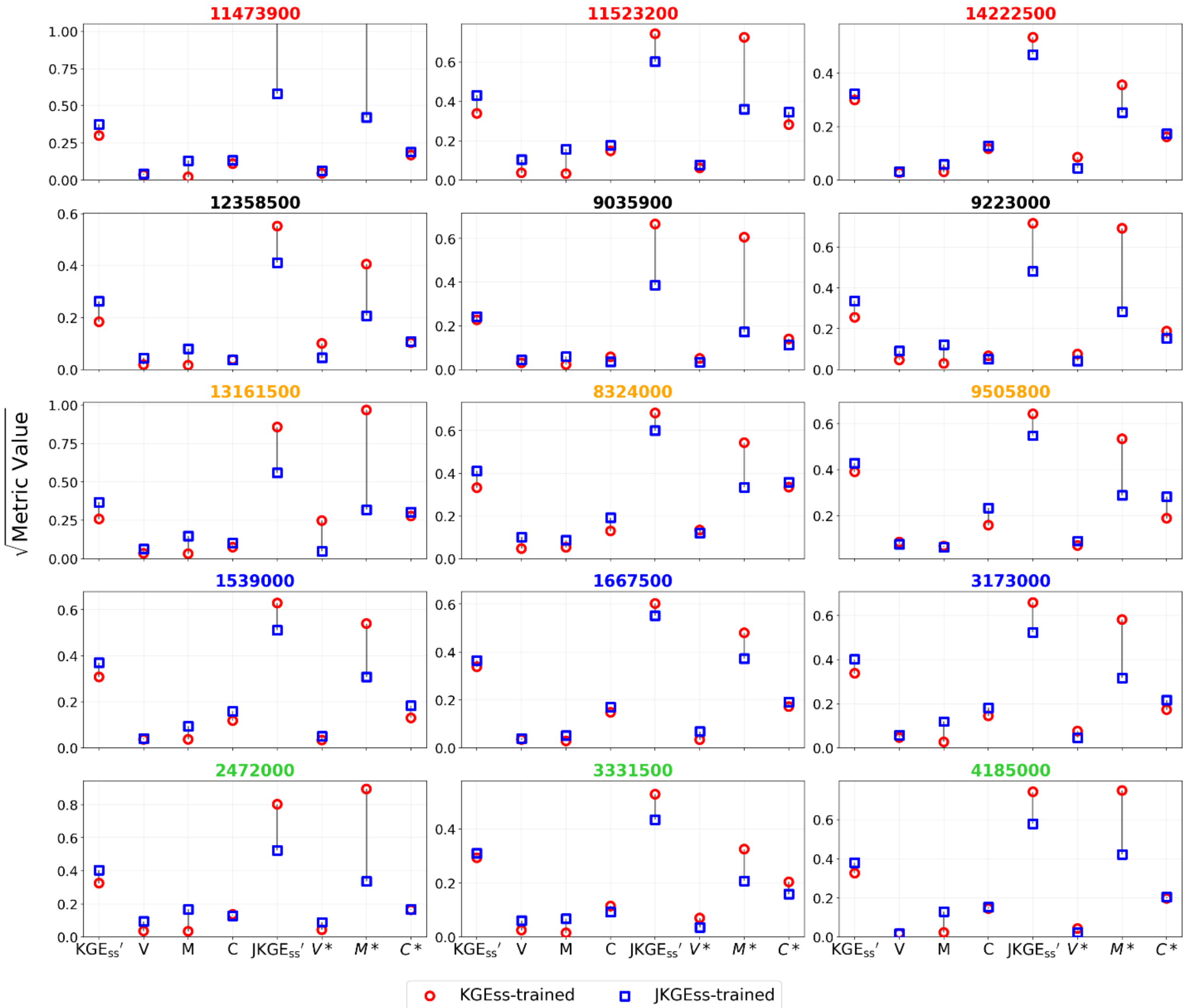

**Figure 13:** Cross-metric evaluation of different metrics when models are trained with $KGE_{ss}$ (red circles) versus $JKGE_{ss}$ (blue squares). Plotted quantities are $KGE_{ss}^{'} = 1 - KGE_{ss}$ ,$V$ ,$M$ ,$C$ , $JKGE_{ss}^{'} = 1 - JKGE_{ss}$ ,$V^{*}$ ,$M^{*}$ ,and $\rho^{*}$ ,with the square root applied for better visualization of smaller values. $JKGE_{ss}^{'}$and $M^{*}$show substantial deviation from zero when training ignores temporal non-stationarity (using $KGE_{ss}$) while $V^{*}$and $\rho^{*}$are largely insensitive to training choice. Some catchments show a trade-off in $M$ performance between the two metrics.

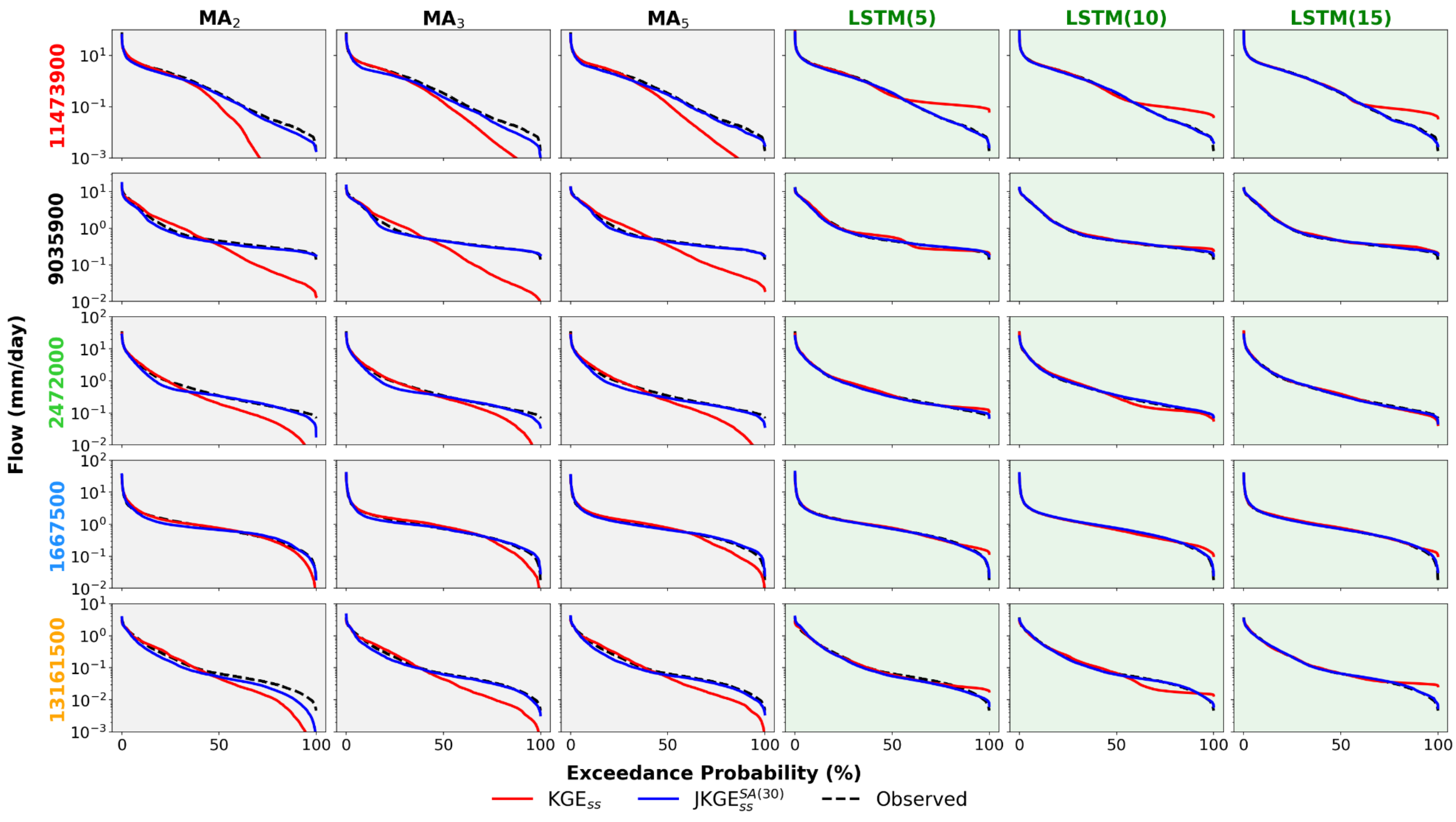

**Figure 14:** Flow Duration Curves (FDCs) for five representative catchments, obtained using physical-conceptual models (MA2, MA3, MA5; columns 1–3) and LSTM-based models (nodes 5, 10, 15; columns 4–6). Observed FDCs are shown with dashed black lines, results of training with $KGE_{ss}$ are shown in red, and results of training with $JKGE_{ss}$ are shown in blue. Models trained with $\mathrm{JKGE}_{\mathrm{ss}}$ reproduce FDCs more accurately across all flow regimes, while $KGE_{ss}$-trained models underestimate low flows for physical-conceptual models and overestimate them for LSTMs. Performance improvements under $\mathrm{JKGE}_{\mathrm{ss}}$ are consistent across all catchments (hydroclimatic regimes).

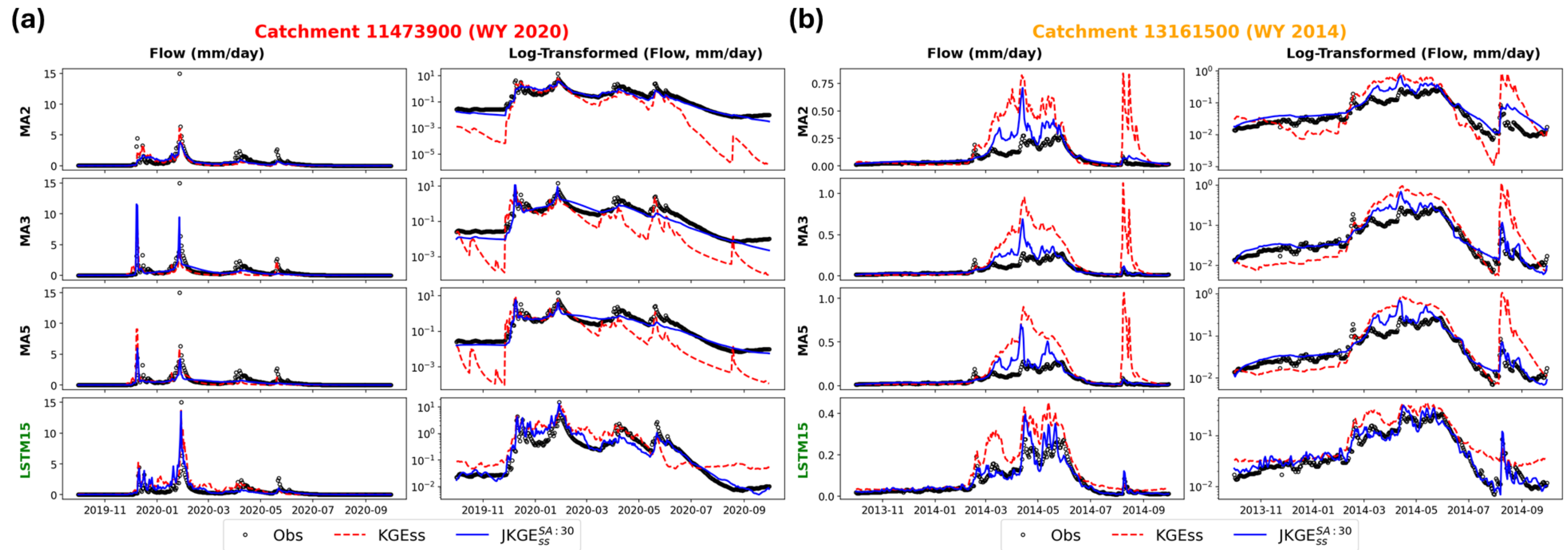

**Figure 15:** Simulated hydrographs for representative wet (11473900, a) and arid (13161500, b) catchments during dry years using physical-conceptual models (MA2, MA3, MA5) and LSTM-15. Left panels show linear-scale flows; right panels show log-transformed flows. Observations are shown in black, results of $\mathrm{KGE}_{ss}$-based training are shown in dashed-red, and results of $\mathrm{JKGE}_{ss}$-based training are shown in blue. Training with $JKGE_{ss}$ results in better reproduction of peak-, intermediate-, and low-flows, with improvements being most pronounced in the arid catchment, particularly during low-flow periods and recession periods.

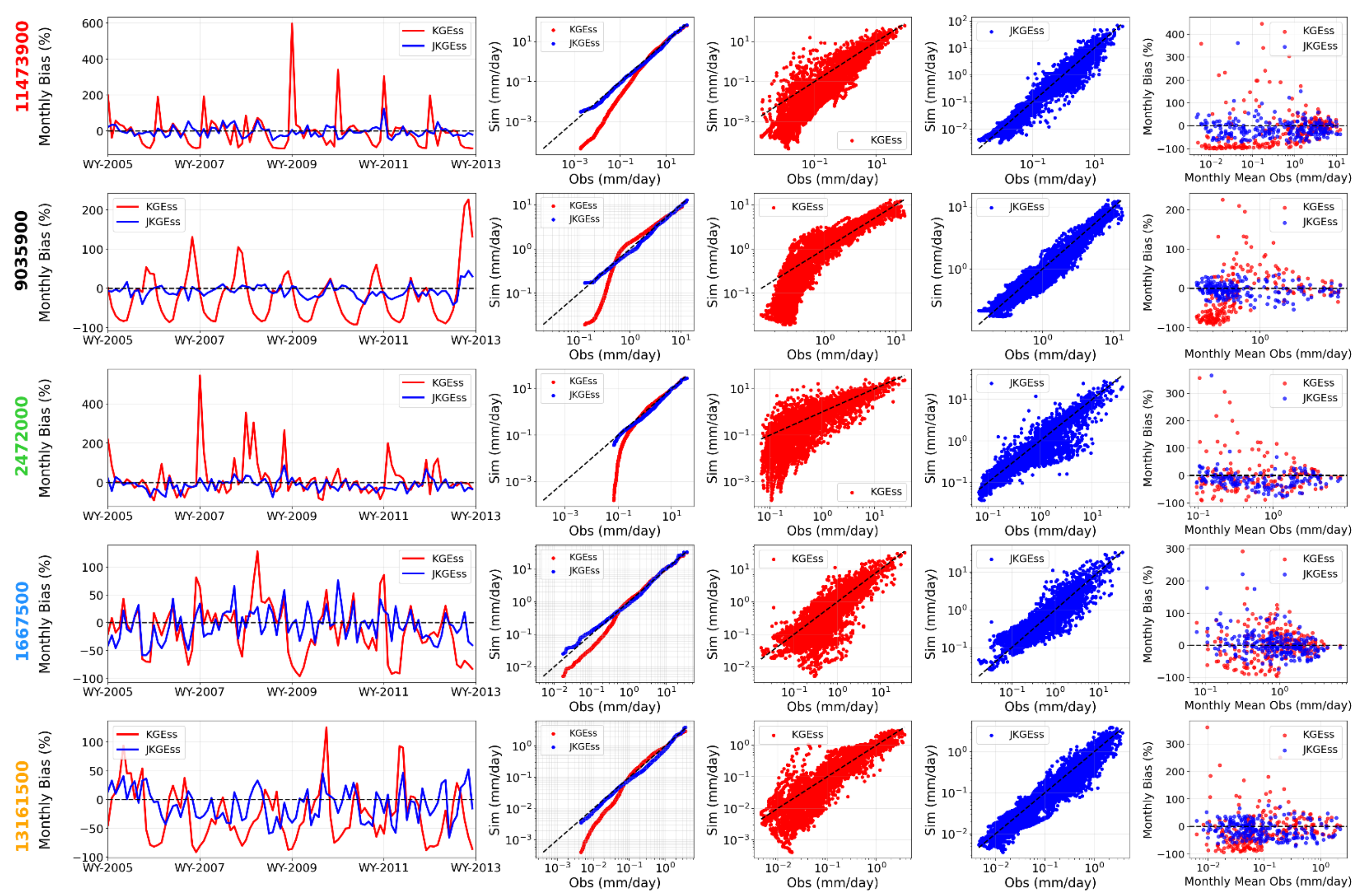

**Figure 16:** Diagnostic evaluation of the MA5 model over 20 years of daily streamflow. Columns show: (1) monthly percent bias for the first eight water years, (2) quantile–quantile (QQ) plots, (3) simulated versus observed flows when trained with $KGE_{ss}$(4) , simulated versus observed flows when trained with $JKGE_{ss}$ ,and (5) monthly bias (%) versus monthly mean observed flow. $\mathrm{KGE}_{ss}$-based results are shown in red, and $JKGE_{ss}$-based results are shown in blue. Training with $JKGE_{ss}$ results in lower and more consistent bias, better alignment with the 1:1 line, and reduced dispersion when compared to training with $KGE_{ss}$.

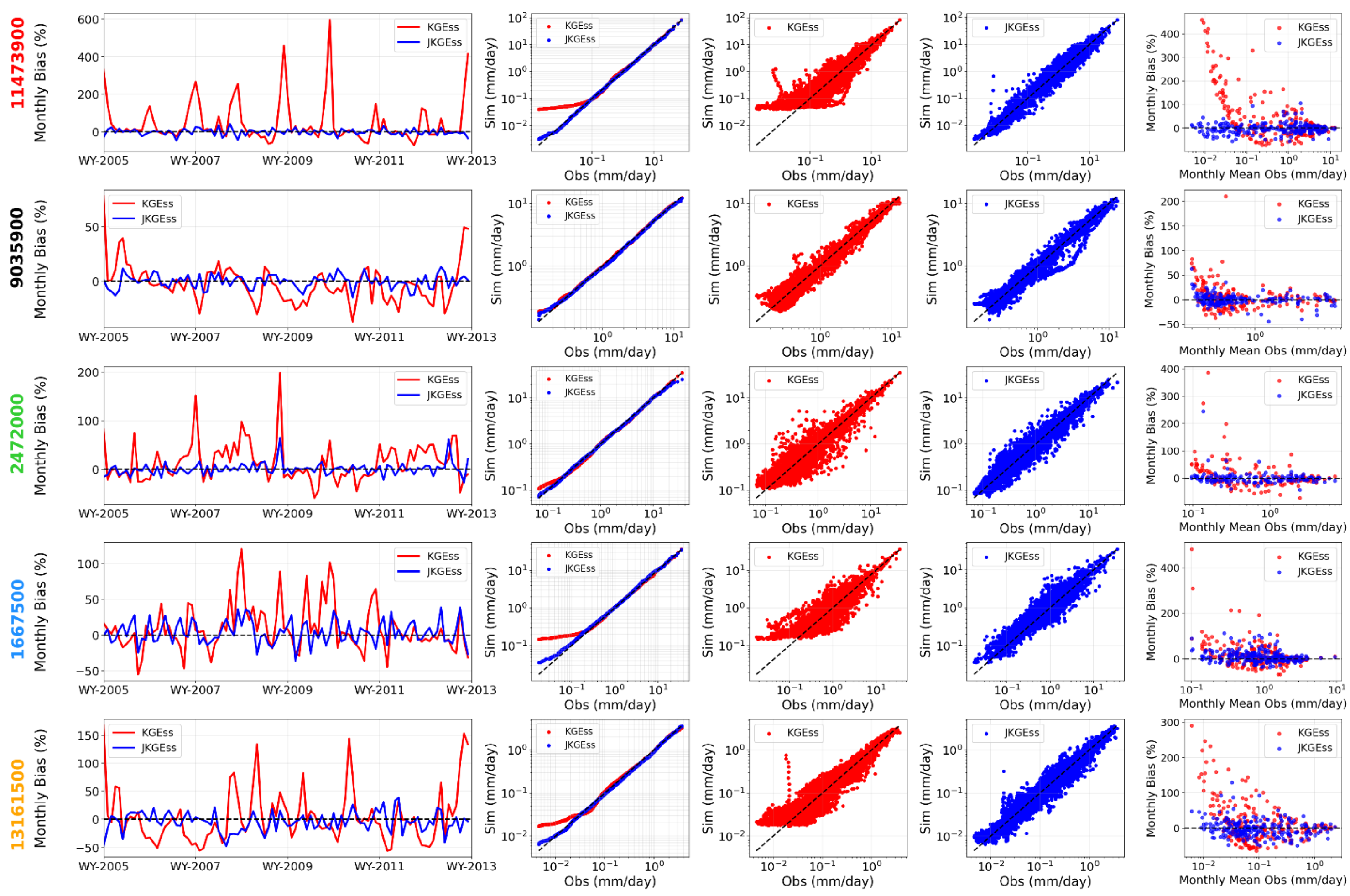

**Figure 17:** Diagnostic evaluation of the LSTM-15 model, with panels organized as in Figure 16. $JKGE_{ss}$ results (blue) exhibit tighter clustering along the 1:1 line, lower variability, and reduced bias across flow magnitudes compared to $KGE_{ss}$ (red), demonstrating the robustness of $JKGE_{ss}$ when used for data-driven modeling.

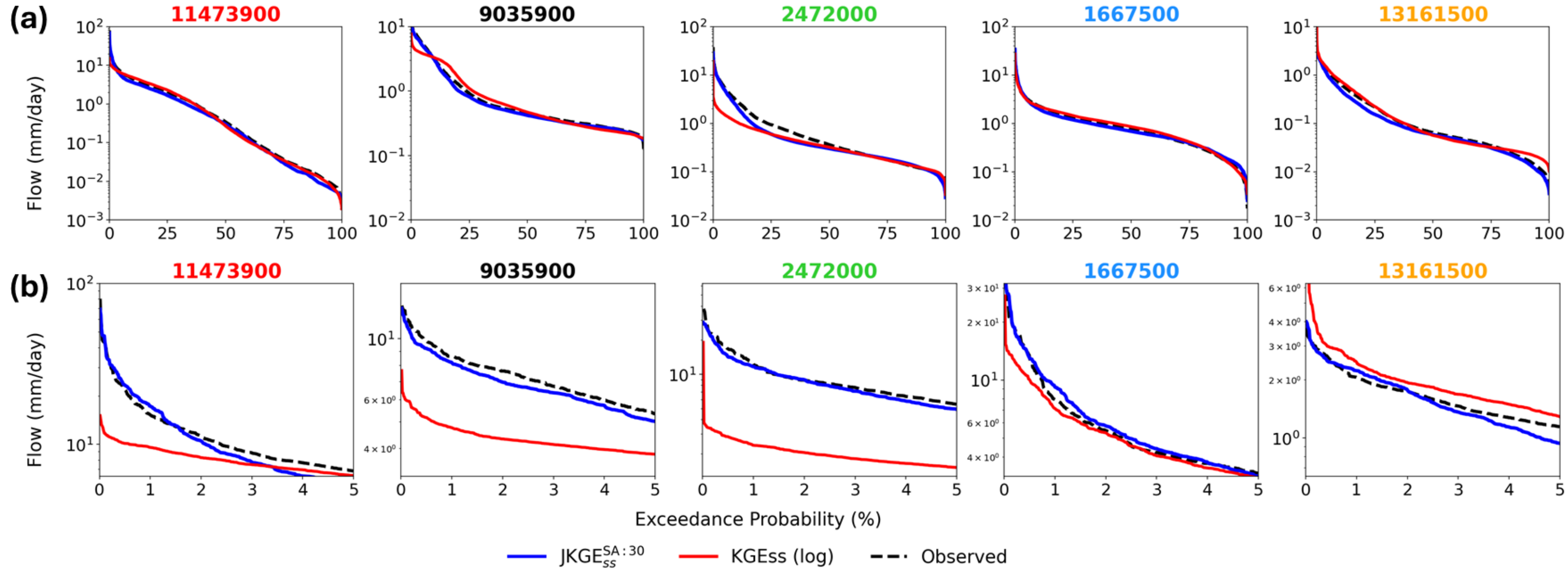

**Figure 18:** Flow Duration Curves (FDCs) for five representative catchments comparing results of training with $KGE_{ss}^{log}$ and $JKGE_{ss}^{SA(30)}$. Panel (a) shows full FDCs, while panel (b) focuses on high flows (0–5% exceedance probability). Training of $KGE_{ss}^{log}$ results in systematic underestimation of peak flows, whereas training with $JKGE_{ss}^{SA(30)}$ typically results in better reproduction.

## Supplementary Materials

M Jawad[1], HV Gupta[1], YH Wang[1,2], MA Farmani[1], A Behrangi[1], and GY Niu[1]

[1]Department of Hydrology and Atmospheric Sciences, The University of Arizona, Tucson, AZ, USA, 85721.

[2] Earth and Environmental Science Area, Lawrence Berkeley National Lab, Berkeley, CA

## Text Sections

### S1. Impacts of Using a Time-Nonstationary *Moving-Average* Mean

[152] **Figure S7** shows, for one representative catchment from each hydrologic regime, the FDCs obtained (in green) by using the *moving-average* mean $JKGE_{ss}^{MA(N_w)}$ for multiple window sizes ($N_w =$ 365, 181, 91, 31, and 7 days; note the requirement that $N_w$ be an odd number). For comparison, the FDCs obtained using the *section-wise* mean $JKGE_{ss}^{SA(N_s)}$ for comparable window sizes ($N_s =$ 365, 180, 90, 30, and 7 days) are shown in blue.

- As with the *section-wise* approach, decreasing the window size leads to systematic performance improvements, best performance is achieved with a window size of around 31 days, and performance tends to degrade for shorter windows (7 days), particularly at high and low flow extremes.
- For window sizes of approximately one month ($N_w = 31$ days and $N_s = 30$ days), the FDC's obtained using the two methods are virtually indistinguishable.

[153] In summary, training using the *moving-average* mean $JKGE_{ss}^{MA(N_w)}$ did not yield additional benefits over use of the *section-wise* mean $JKGE_{ss}^{SA(N_s)}$ – overall performance was comparable.

- Best performance was achieved with a section length of $N_w = 31$ days, which is consistent with the value of $N_s = 30$ days obtained in the previous analysis.

[154] So, while the *moving-average* approach might theoretically seem to be somewhat more elegant (providing a smoothly varying benchmark), it provides *essentially identical results with much higher computational cost* since the average must be computed at each and every time step (in the *section-wise* approach the section is computed only once for all time steps within a section, making it more computationally more efficient).

[155] Overall, these findings reinforce those of the *section-wise* approach, that computing the time varying benchmark using a monthly time scale for aggregation provides an optimal balance – it effectively smooths out shorter-term hydrologic variability while preserving relevant and meaningful information about overall flow dynamics.

### S2. Impact of Non-Stationary Standard Deviation:

[156] The current formulation of $JKGE_{ss}$ incorporates a non-stationary mean component (*M**), implemented either through a section-based mean ($JKGE_{ss}^{SA(N_s)}$) or a moving-average mean

($JKGE_{ss}^{MA(N_w)}$), as described in Sections 2.4.1.1 and 2.4.1.2 and illustrated in Sections 4.1 and 4.2, respectively. The standard deviation component, represented by the $A^*$ term, is assumed to be stationary. In principle, all three terms *V*, M**, and *C** of $JKGE_{ss}$ can be treated as non-stationary. For example, the standard deviation can be evaluated along with the non-stationary mean by computing it separately within each aggregation group.

- In case of the section-approach where the whole timeseries is divided into equal length groups indexed by $k = 1, \dots, K$, the simulated and observed standard deviations can be defined for group $k$ as $\psi_s^k \approx \sqrt{\frac{1}{N_k}\sum_{t\in T_k}(s_t - b_k^s)^2}$ and $\psi_o^k \approx \sqrt{\frac{1}{N_k}\sum_{t\in T_k}(o_t - b_k^o)^2}$ where $T_k$ denotes the set of time indices belonging to group $k$ and $N_k$ is the number of time steps within that group. These group-specific standard deviations are then broadcast to the full time series to compute a time-varying variability ratio $\alpha_t = \frac{\psi^s_{k(t)}}{\psi^o_{k(t)}}$ leading to the variability penalty term $V^* = \frac{1}{N}\sum_{t=1}^{N}(1-\alpha_t)^2$. This formulation (we refer as $JKGE_{ss}^{\mu\&\sigma}$) allows the variability component to evolve across time in a manner consistent with the non-stationary mean computed via either section-based or moving-average aggregation.
- In the *moving-average* (MA) formulation, both the mean and standard deviation are allowed to vary continuously over time. For a given MA window length $N_w$, the time-varying standard deviations can be computed as: $\psi_s^t \approx \sqrt{\frac{1}{N_w}\sum_{t-h}^{t+h}(s_t - b_t^s)^2}$ , $\psi_o^t \approx \sqrt{\frac{1}{N_w}\sum_{t-h}^{t+h}(o_t - b_t^s)^2}$ whereas the time-varying variability ratio is then defined as $\alpha_t = \frac{\psi_t^s}{\psi_t^o}$ and $V *= \frac{1}{N}\sum_{t=1}^{N}(1-\alpha_t)^2$.

**S3. Statistical performance Comparison of Different Models**

[157] **Tables S1-S4** compare statistical performance of the models across eight descriptors or metrics: $KGE_{ss}$, $M$, $V$, $C$, $JKGE_{ss}$, $M^*$, $V^*$, and $C^*$, computed from the model output streamflow for runs trained using either $KGE_{ss}$ or $JKGE_{ss}$ as the metric.

[158] From these bootstrapped distributions (as mentioned in Section 4.4), the median as well as the 5th (Q5) and 95th (Q95) percentiles were extracted for each catchment and scenario (models trained with $KGE_{ss}$ or $JKGE_{ss}$) for all model configurations, including MA$_5$, MA$_3$, MA$_2$, LSTM5, LSTM10, and LSTM15. **Tables S1-S3** present results for the MA$_2$, MA$_3$, and MA$_5$ models, while **Tables S4** shows results for LSTM15. For each catchment, the results are presented in two rows: the upper row reports the median value of the metric, while the lower row shows the corresponding quantile range. For $KGE_{ss}$ and $JKGE_{ss}$, the quantile ranges are reported as $(Q5, Q95)$. For the remaining descriptors ($M$, $V$, $C$, $M^*$, $V^*$, and $C^*$), the ranges are reported as $(Q95, Q5)$ so that the order corresponds to (worst, best), which facilitates interpretation because smaller values approaching zero indicate better performance for these components, in contrast to $KGE_{ss}$ and $JKGE_{ss}$, where larger values indicate better performance.

[159] For the MA$_5$ model (**Table S3**), training with $JKGE_{ss}$ consistently leads to improved performance when evaluated using the generalized metric and its component descriptors.

- Across the catchments, median $JKGE_{ss}$ for the $JKGE_{ss}$-trained models range from 0.66 to 0.86, while corresponding component errors ($M^*$, $V^*$, and $C^*$) remain close to zero, indicating balanced improvements to variability, bias, and correlation.
- In contrast, models trained with $KGE_{ss}$ show substantially lower $JKGE_{ss}$ and noticeably larger component errors, particularly in the bias-related term $M^*$, suggesting reduced consistency with respect to the generalized performance formulation.
- Although $KGE_{ss}$-trained models tend to yield slightly higher values when evaluated with $KGE_{ss}$, this improvement does not translate to the generalized framework.
- The bootstrap ranges also indicate that $JKGE_{ss}$-trained models are consistently superior across most catchments, supporting $JKGE_{ss}$ as a more robust metric for model training.

[160] For the LSTM-15 model (**Table S4**), training with $JKGE_{ss}$ again demonstrates clear advantages when performance is evaluated using $JKGE_{ss}$ and its component descriptors.

- Across the catchments, median $JKGE_{ss}$ for the $JKGE_{ss}$-trained models are consistently high, ranging from 0.79 to 0.91, while associated component errors ($M^*$, $V^*$, and $C^*$) remain close to zero, indicating well-balanced reproduction of variability, bias, and correlation.
- In contrast, models trained with $KGE_{ss}$ show substantially lower $JKGE_{ss}$ and markedly larger $JKGE_{ss}$' component errors, particularly in the bias-related term $M^*$, with some catchments even exhibiting strongly degraded $JKGE_{ss}$ (e.g., negative values).
- Although $KGE_{ss}$-trained models often achieve slightly higher scores when evaluated with $KGE_{ss}$, these gains do not translate to the generalized formulation.
- The bootstrap ranges further indicate that the improved $JKGE_{ss}$ performance of $JKGE_{ss}$-trained models is consistent across most catchments, reinforcing the robustness of $JKGE_{ss}$ as a metric for training.

## Tables

**Table S1**: Statistical performance comparison of Model $MA_2$ across metrics $KGE_{SS}$, $M$, $V$, $C$, $JKGE_{SS}$, $M^*$, $V^*$, and $C^*$, evaluated for models trained with $KGE_{SS}$ and $JKGE_{SS}$. Values represent medians and bootstrap ranges (5th–95th percentiles) computed using yearly-block resampling. For $KGE_{SS}$ and $JKGE_{SS}$, higher values indicate better performance, whereas for component metrics lower values (approaching zero) indicate better performance. Results show that training with $JKGE_{SS}$ yields consistently improved generalized performance and reduced component errors, particularly in the bias-related term $M^*$, across catchments.

| Catchment | $JKGE_{ss}$ Trained | | | | | | | | $KGE_{ss}$ Trained | | | | | | | |
|---|---|---|---|---|---|---|---|---|---|---|---|---|---|---|---|---|
| | $JKGE_{ss}$ | V* | M* | C* | $KGE_{ss}$ | V | M | C | $JKGE_{ss}$ | V* | M* | C* | $KGE_{ss}$ | V | M | C |
| 11473900 | **0.672** | **0.003** | **0.145** | **0.066** | **0.836** | **0.006** | **0.019** | **0.027** | **0.479** | **0.002** | **0.510** | **0.028** | **0.915** | **0.001** | **0.001** | **0.011** |
| | (0.647,0.695) | (0.019,0.000) | (0.168,0.121) | (0.084,0.050) | (0.794,0.868) | (0.024,0.000) | (0.036,0.008) | (0.033,0.022) | (0.447,0.507) | (0.014,0.000) | (0.582,0.457) | (0.037,0.021) | (0.886,0.926) | (0.009,0.000) | (0.007,0.000) | (0.015,0.009) |
| 11523200 | **0.607** | **0.005** | **0.185** | **0.110** | **0.664** | **0.054** | **0.119** | **0.053** | **0.387** | **0.005** | **0.664** | **0.080** | **0.881** | **0.002** | **0.002** | **0.022** |
| | (0.575,0.638) | (0.045,0.000) | (0.215,0.159) | (0.135,0.086) | (0.600,0.724) | (0.112,0.013) | (0.164,0.075) | (0.072,0.038) | (0.235,0.492) | (0.037,0.000) | (1.079,0.429) | (0.098,0.061) | (0.844,0.904) | (0.014,0.000) | (0.013,0.000) | (0.033,0.014) |
| 14222500 | **0.751** | **0.001** | **0.084** | **0.039** | **0.887** | **0.001** | **0.002** | **0.021** | **0.599** | **0.010** | **0.281** | **0.030** | **0.905** | **0.001** | **0.001** | **0.016** |
| | (0.716,0.774) | (0.010,0.000) | (0.108,0.068) | (0.052,0.030) | (0.844,0.909) | (0.009,0.000) | (0.014,0.000) | (0.033,0.014) | (0.570,0.624) | (0.027,0.001) | (0.313,0.251) | (0.039,0.023) | (0.872,0.920) | (0.006,0.000) | (0.006,0.000) | (0.024,0.010) |
| 12358500 | **0.777** | **0.003** | **0.072** | **0.022** | **0.899** | **0.004** | **0.014** | **0.003** | **0.656** | **0.008** | **0.207** | **0.020** | **0.958** | **0.000** | **0.000** | **0.002** |
| | (0.748,0.801) | (0.015,0.000) | (0.087,0.059) | (0.034,0.014) | (0.863,0.930) | (0.012,0.000) | (0.024,0.007) | (0.005,0.002) | (0.610,0.694) | (0.018,0.002) | (0.270,0.161) | (0.030,0.013) | (0.938,0.969) | (0.003,0.000) | (0.003,0.000) | (0.004,0.002) |
| 9035900 | **0.792** | **0.008** | **0.052** | **0.023** | **0.828** | **0.030** | **0.025** | **0.004** | **0.520** | **0.002** | **0.431** | **0.024** | **0.944** | **0.001** | **0.001** | **0.004** |
| | (0.743,0.829) | (0.038,0.000) | (0.068,0.040) | (0.043,0.011) | (0.770,0.890) | (0.062,0.008) | (0.041,0.013) | (0.006,0.002) | (0.456,0.576) | (0.016,0.000) | (0.559,0.338) | (0.043,0.012) | (0.911,0.958) | (0.008,0.000) | (0.004,0.000) | (0.007,0.002) |
| 9223000 | **0.681** | **0.003** | **0.138** | **0.056** | **0.855** | **0.007** | **0.025** | **0.009** | **0.523** | **0.009** | **0.399** | **0.044** | **0.930** | **0.002** | **0.001** | **0.006** |
| | (0.643,0.715) | (0.034,0.000) | (0.163,0.117) | (0.093,0.031) | (0.780,0.913) | (0.032,0.000) | (0.052,0.006) | (0.015,0.005) | (0.436,0.593) | (0.040,0.000) | (0.590,0.261) | (0.068,0.027) | (0.881,0.952) | (0.013,0.000) | (0.010,0.000) | (0.010,0.003) |
| 13161500 | **0.597** | **0.007** | **0.220** | **0.094** | **0.834** | **0.010** | **0.034** | **0.010** | **0.214** | **0.046** | **1.079** | **0.110** | **0.918** | **0.002** | **0.002** | **0.009** |
| | (0.545,0.637) | (0.035,0.000) | (0.270,0.182) | (0.159,0.056) | (0.745,0.899) | (0.049,0.000) | (0.071,0.011) | (0.019,0.006) | (-0.156,0.482) | (0.074,0.017) | (2.512,0.398) | (0.184,0.062) | (0.871,0.941) | (0.014,0.000) | (0.011,0.000) | (0.014,0.005) |
| 8324000 | **0.634** | **0.025** | **0.104** | **0.138** | **0.722** | **0.055** | **0.052** | **0.044** | **0.500** | **0.042** | **0.326** | **0.125** | **0.879** | **0.002** | **0.002** | **0.022** |
| | (0.517,0.703) | (0.167,0.000) | (0.131,0.081) | (0.209,0.070) | (0.593,0.843) | (0.185,0.003) | (0.098,0.015) | (0.074,0.021) | (0.410,0.579) | (0.103,0.005) | (0.481,0.222) | (0.198,0.072) | (0.824,0.921) | (0.017,0.000) | (0.015,0.000) | (0.043,0.010) |
| 9505800 | **0.688** | **0.023** | **0.074** | **0.086** | **0.792** | **0.019** | **0.005** | **0.058** | **0.237** | **0.012** | **1.065** | **0.065** | **0.789** | **0.024** | **0.006** | **0.042** |
| | (0.578,0.764) | (0.108,0.000) | (0.096,0.056) | (0.210,0.029) | (0.668,0.861) | (0.068,0.000) | (0.027,0.000) | (0.153,0.021) | (0.162,0.308) | (0.082,0.000) | (1.322,0.866) | (0.142,0.017) | (0.697,0.882) | (0.107,0.000) | (0.049,0.000) | (0.089,0.016) |
| 2472000 | **0.711** | **0.005** | **0.120** | **0.040** | **0.845** | **0.004** | **0.020** | **0.023** | **0.162** | **0.003** | **1.359** | **0.042** | **0.848** | **0.002** | **0.005** | **0.037** |
| | (0.682,0.733) | (0.021,0.000) | (0.152,0.094) | (0.051,0.031) | (0.808,0.870) | (0.017,0.000) | (0.034,0.010) | (0.029,0.018) | (-0.146,0.465) | (0.013,0.000) | (2.584,0.531) | (0.063,0.027) | (0.805,0.881) | (0.011,0.000) | (0.015,0.000) | (0.062,0.021) |
| 3331500 | **0.752** | **0.004** | **0.066** | **0.051** | **0.883** | **0.003** | **0.007** | **0.016** | **0.682** | **0.004** | **0.143** | **0.051** | **0.904** | **0.001** | **0.000** | **0.017** |
| | (0.695,0.788) | (0.033,0.000) | (0.083,0.051) | (0.090,0.028) | (0.843,0.909) | (0.017,0.000) | (0.013,0.003) | (0.027,0.009) | (0.635,0.722) | (0.023,0.000) | (0.204,0.101) | (0.072,0.034) | (0.881,0.920) | (0.004,0.000) | (0.002,0.000) | (0.026,0.012) |
| 4185000 | **0.650** | **0.001** | **0.192** | **0.048** | **0.837** | **0.000** | **0.026** | **0.026** | **0.240** | **0.003** | **1.114** | **0.040** | **0.886** | **0.000** | **0.001** | **0.024** |
| | (0.617,0.681) | (0.004,0.000) | (0.225,0.162) | (0.075,0.033) | (0.809,0.863) | (0.003,0.000) | (0.036,0.017) | (0.041,0.017) | (0.071,0.414) | (0.012,0.000) | (1.696,0.631) | (0.066,0.026) | (0.854,0.908) | (0.003,0.000) | (0.005,0.000) | (0.038,0.016) |
| 1539000 | **0.725** | **0.004** | **0.106** | **0.036** | **0.855** | **0.002** | **0.010** | **0.027** | **0.631** | **0.003** | **0.239** | **0.026** | **0.887** | **0.002** | **0.000** | **0.021** |
| | (0.676,0.757) | (0.018,0.000) | (0.145,0.085) | (0.077,0.019) | (0.816,0.883) | (0.013,0.000) | (0.019,0.005) | (0.053,0.015) | (0.570,0.696) | (0.018,0.000) | (0.334,0.157) | (0.062,0.014) | (0.850,0.914) | (0.013,0.000) | (0.003,0.000) | (0.042,0.011) |
| 1667500 | **0.709** | **0.004** | **0.105** | **0.057** | **0.834** | **0.002** | **0.010** | **0.041** | **0.586** | **0.002** | **0.279** | **0.058** | **0.843** | **0.002** | **0.001** | **0.043** |
| | (0.668,0.740) | (0.025,0.000) | (0.128,0.084) | (0.101,0.031) | (0.789,0.868) | (0.017,0.000) | (0.021,0.002) | (0.070,0.024) | (0.528,0.633) | (0.021,0.000) | (0.369,0.205) | (0.091,0.036) | (0.807,0.873) | (0.016,0.000) | (0.007,0.000) | (0.067,0.028) |
| 3173000 | **0.687** | **0.007** | **0.133** | **0.051** | **0.840** | **0.005** | **0.008** | **0.035** | **0.597** | **0.003** | **0.277** | **0.038** | **0.868** | **0.004** | **0.001** | **0.027** |
| | (0.655,0.718) | (0.029,0.000) | (0.161,0.108) | (0.077,0.034) | (0.811,0.868) | (0.021,0.000) | (0.022,0.001) | (0.054,0.024) | (0.551,0.638) | (0.017,0.000) | (0.358,0.221) | (0.065,0.024) | (0.834,0.891) | (0.018,0.000) | (0.009,0.000) | (0.043,0.017) |

**Table S2**: Statistical performance comparison of Model $MA_3$ across metrics $KGE_{SS}$, $M$, $V$, $C$, $JKGE_{SS}$, $M^*$, $V^*$, and $C^*$, evaluated for models trained with $KGE_{SS}$ and $JKGE_{SS}$. Values represent medians and bootstrap ranges (5th–95th percentiles) computed using yearly-block resampling. For $KGE_{SS}$ and $JKGE_{SS}$, higher values indicate better performance, whereas for component metrics lower values (approaching zero) indicate better performance. Results show that training with $JKGE_{SS}$ yields consistently improved generalized performance and reduced component errors, particularly in the bias-related term $M^*$, across catchments.

| Catchment | $JKGE_{ss}$ Trained | | | | | | | | $KGE_{ss}$ Trained | | | | | | | |
|---|---|---|---|---|---|---|---|---|---|---|---|---|---|---|---|---|
| | $JKGE_{ss}$ | V* | M* | C* | $KGE_{ss}$ | V | M | C | $JKGE_{ss}$ | V* | M* | C* | $KGE_{ss}$ | V | M | C |
| 11473900 | **0.559** | **0.004** | **0.344** | **0.036** | **0.837** | **0.002** | **0.030** | **0.019** | **-0.738** | **0.003** | **6.000** | **0.033** | **0.902** | **0.002** | **0.001** | **0.015** |
| | (0.295,0.707) | (0.038,0.000) | (0.926,0.138) | (0.061,0.023) | (0.786,0.872) | (0.017,0.000) | (0.051,0.016) | (0.032,0.013) | (-1.601,0.041) | (0.025,0.000) | (13.494,1.792) | (0.059,0.018) | (0.858,0.929) | (0.014,0.000) | (0.009,0.000) | (0.026,0.009) |
| 11523200 | **0.634** | **0.007** | **0.130** | **0.123** | **0.708** | **0.043** | **0.075** | **0.051** | **0.464** | **0.006** | **0.469** | **0.098** | **0.873** | **0.002** | **0.001** | **0.027** |
| | (0.599,0.668) | (0.052,0.000) | (0.164,0.104) | (0.151,0.098) | (0.658,0.761) | (0.091,0.006) | (0.103,0.046) | (0.071,0.033) | (0.348,0.536) | (0.044,0.000) | (0.738,0.332) | (0.123,0.074) | (0.836,0.896) | (0.016,0.000) | (0.008,0.000) | (0.040,0.018) |
| 14222500 | **0.782** | **0.002** | **0.060** | **0.032** | **0.893** | **0.001** | **0.004** | **0.017** | **0.652** | **0.008** | **0.209** | **0.025** | **0.915** | **0.001** | **0.001** | **0.013** |
| | (0.744,0.810) | (0.014,0.000) | (0.084,0.044) | (0.043,0.024) | (0.836,0.920) | (0.010,0.000) | (0.018,0.000) | (0.028,0.011) | (0.622,0.680) | (0.023,0.000) | (0.242,0.179) | (0.034,0.018) | (0.885,0.929) | (0.005,0.000) | (0.005,0.000) | (0.019,0.008) |
| 12358500 | **0.824** | **0.002** | **0.048** | **0.011** | **0.926** | **0.002** | **0.007** | **0.001** | **0.683** | **0.012** | **0.176** | **0.011** | **0.962** | **0.001** | **0.000** | **0.002** |
| | (0.796,0.843) | (0.008,0.000) | (0.067,0.037) | (0.017,0.008) | (0.898,0.950) | (0.008,0.000) | (0.012,0.003) | (0.002,0.001) | (0.646,0.719) | (0.023,0.004) | (0.226,0.136) | (0.017,0.008) | (0.937,0.973) | (0.004,0.000) | (0.003,0.000) | (0.003,0.001) |
| 9035900 | **0.833** | **0.002** | **0.035** | **0.018** | **0.935** | **0.002** | **0.003** | **0.002** | **0.502** | **0.010** | **0.450** | **0.030** | **0.941** | **0.001** | **0.000** | **0.005** |
| | (0.791,0.852) | (0.015,0.000) | (0.041,0.030) | (0.036,0.009) | (0.884,0.963) | (0.014,0.000) | (0.010,0.000) | (0.004,0.001) | (0.437,0.557) | (0.032,0.000) | (0.579,0.360) | (0.059,0.015) | (0.908,0.956) | (0.008,0.000) | (0.005,0.000) | (0.008,0.003) |
| 9223000 | **0.738** | **0.003** | **0.095** | **0.036** | **0.862** | **0.013** | **0.020** | **0.006** | **0.482** | **0.005** | **0.487** | **0.038** | **0.930** | **0.002** | **0.001** | **0.005** |
| | (0.706,0.763) | (0.023,0.000) | (0.113,0.081) | (0.061,0.020) | (0.794,0.927) | (0.036,0.000) | (0.042,0.004) | (0.010,0.003) | (0.390,0.569) | (0.030,0.000) | (0.711,0.308) | (0.065,0.024) | (0.881,0.953) | (0.016,0.000) | (0.008,0.000) | (0.009,0.003) |
| 13161500 | **0.672** | **0.003** | **0.109** | **0.100** | **0.870** | **0.004** | **0.020** | **0.010** | **0.300** | **0.077** | **0.811** | **0.095** | **0.927** | **0.001** | **0.001** | **0.007** |
| | (0.624,0.709) | (0.025,0.000) | (0.127,0.090) | (0.163,0.060) | (0.790,0.911) | (0.029,0.000) | (0.047,0.004) | (0.019,0.006) | (0.013,0.521) | (0.113,0.043) | (1.778,0.294) | (0.162,0.056) | (0.882,0.947) | (0.012,0.000) | (0.009,0.000) | (0.012,0.004) |
| 8324000 | **0.635** | **0.024** | **0.105** | **0.134** | **0.725** | **0.052** | **0.053** | **0.043** | **0.502** | **0.031** | **0.312** | **0.146** | **0.883** | **0.002** | **0.002** | **0.022** |
| | (0.519,0.711) | (0.174,0.000) | (0.134,0.078) | (0.203,0.068) | (0.574,0.854) | (0.200,0.001) | (0.107,0.015) | (0.075,0.020) | (0.407,0.596) | (0.116,0.002) | (0.422,0.208) | (0.257,0.075) | (0.822,0.919) | (0.013,0.000) | (0.015,0.000) | (0.049,0.010) |
| 9505800 | **0.720** | **0.013** | **0.054** | **0.079** | **0.782** | **0.006** | **0.022** | **0.060** | **0.245** | **0.014** | **1.026** | **0.065** | **0.804** | **0.017** | **0.004** | **0.043** |
| | (0.614,0.787) | (0.075,0.000) | (0.068,0.041) | (0.214,0.030) | (0.664,0.859) | (0.048,0.000) | (0.052,0.006) | (0.152,0.022) | (0.169,0.310) | (0.095,0.000) | (1.273,0.881) | (0.149,0.015) | (0.704,0.896) | (0.082,0.000) | (0.038,0.000) | (0.098,0.013) |
| 2472000 | **0.710** | **0.005** | **0.121** | **0.041** | **0.838** | **0.004** | **0.024** | **0.024** | **0.334** | **0.002** | **0.843** | **0.040** | **0.871** | **0.002** | **0.002** | **0.029** |
| | (0.677,0.736) | (0.019,0.000) | (0.157,0.093) | (0.053,0.031) | (0.801,0.865) | (0.018,0.000) | (0.038,0.012) | (0.031,0.018) | (0.145,0.541) | (0.011,0.000) | (1.411,0.389) | (0.058,0.028) | (0.829,0.901) | (0.010,0.000) | (0.009,0.000) | (0.046,0.018) |
| 3331500 | **0.794** | **0.001** | **0.049** | **0.032** | **0.888** | **0.005** | **0.008** | **0.011** | **0.702** | **0.010** | **0.142** | **0.022** | **0.928** | **0.000** | **0.000** | **0.009** |
| | (0.765,0.819) | (0.010,0.000) | (0.063,0.040) | (0.053,0.019) | (0.864,0.912) | (0.015,0.000) | (0.015,0.003) | (0.018,0.007) | (0.667,0.732) | (0.027,0.001) | (0.182,0.113) | (0.037,0.013) | (0.904,0.943) | (0.004,0.000) | (0.002,0.000) | (0.016,0.005) |
| 4185000 | **0.662** | **0.001** | **0.173** | **0.052** | **0.851** | **0.000** | **0.016** | **0.028** | **0.443** | **0.002** | **0.575** | **0.043** | **0.887** | **0.000** | **0.001** | **0.023** |
| | (0.621,0.694) | (0.007,0.000) | (0.220,0.140) | (0.082,0.035) | (0.817,0.873) | (0.004,0.000) | (0.025,0.010) | (0.044,0.019) | (0.263,0.581) | (0.011,0.000) | (1.032,0.307) | (0.065,0.029) | (0.859,0.910) | (0.003,0.000) | (0.005,0.000) | (0.036,0.015) |
| 1539000 | **0.725** | **0.004** | **0.107** | **0.037** | **0.848** | **0.003** | **0.014** | **0.028** | **0.633** | **0.003** | **0.236** | **0.026** | **0.886** | **0.002** | **0.001** | **0.020** |
| | (0.681,0.756) | (0.017,0.000) | (0.144,0.085) | (0.086,0.019) | (0.804,0.879) | (0.014,0.000) | (0.024,0.007) | (0.057,0.015) | (0.578,0.679) | (0.016,0.000) | (0.310,0.179) | (0.056,0.013) | (0.850,0.914) | (0.012,0.000) | (0.006,0.000) | (0.039,0.011) |
| 1667500 | **0.702** | **0.004** | **0.123** | **0.046** | **0.849** | **0.002** | **0.007** | **0.036** | **0.634** | **0.002** | **0.218** | **0.042** | **0.862** | **0.002** | **0.003** | **0.030** |
| | (0.660,0.733) | (0.020,0.000) | (0.164,0.093) | (0.082,0.024) | (0.794,0.883) | (0.012,0.000) | (0.020,0.000) | (0.063,0.020) | (0.586,0.671) | (0.016,0.000) | (0.296,0.165) | (0.073,0.023) | (0.828,0.891) | (0.012,0.000) | (0.014,0.000) | (0.050,0.017) |
| 3173000 | **0.708** | **0.008** | **0.119** | **0.040** | **0.858** | **0.003** | **0.006** | **0.028** | **0.654** | **0.002** | **0.204** | **0.031** | **0.877** | **0.004** | **0.002** | **0.021** |
| | (0.659,0.747) | (0.029,0.000) | (0.167,0.085) | (0.064,0.025) | (0.824,0.881) | (0.017,0.000) | (0.016,0.001) | (0.045,0.018) | (0.618,0.687) | (0.013,0.000) | (0.249,0.162) | (0.056,0.019) | (0.847,0.900) | (0.016,0.000) | (0.010,0.000) | (0.038,0.013) |

**Table S3**: Statistical performance comparison of Model $MA_5$ across metrics $KGE_{SS}$, $M$, $V$, $C$, $JKGE_{SS}$, $M^*$, $V^*$, and $C^*$, evaluated for models trained with $KGE_{SS}$ and $JKGE_{SS}$. Values represent medians and bootstrap ranges (5th–95th percentiles) computed using yearly-block resampling. For $KGE_{SS}$ and $JKGE_{SS}$, higher values indicate better performance, whereas for component metrics lower values (approaching zero) indicate better performance. Results show that training with $JKGE_{SS}$ yields consistently improved generalized performance and reduced component errors, particularly in the bias-related term $M^*$, across catchments.

| Catchment | $JKGE_{ss}$ Trained | | | | | | | | $KGE_{ss}$ Trained | | | | | | | |
|---|---|---|---|---|---|---|---|---|---|---|---|---|---|---|---|---|
| | $JKGE_{ss}$ | V* | M* | C* | $KGE_{ss}$ | V | M | C | $JKGE_{ss}$ | V* | M* | C* | $KGE_{ss}$ | V | M | C |
| 11473900 | 0.662 | 0.004 | 0.189 | 0.029 | 0.890 | 0.002 | 0.004 | 0.016 | 0.073 | 0.002 | 1.689 | 0.024 | 0.910 | 0.002 | 0.001 | 0.012 |
| | (0.476,0.772) | (0.033,0.000) | (0.491,0.078) | (0.048,0.018) | (0.848,0.919) | (0.016,0.000) | (0.015,0.000) | (0.026,0.010) | (-0.737,0.471) | (0.021,0.000) | (6.001,0.537) | (0.043,0.015) | (0.866,0.931) | (0.017,0.000) | (0.005,0.000) | (0.022,0.008) |
| 11523200 | 0.682 | 0.009 | 0.109 | 0.081 | 0.712 | 0.046 | 0.083 | 0.038 | 0.517 | 0.003 | 0.399 | 0.061 | 0.883 | 0.002 | 0.001 | 0.023 |
| | (0.643,0.711) | (0.049,0.000) | (0.130,0.089) | (0.100,0.068) | (0.653,0.771) | (0.089,0.007) | (0.119,0.053) | (0.049,0.029) | (0.434,0.570) | (0.024,0.000) | (0.568,0.304) | (0.077,0.047) | (0.857,0.902) | (0.010,0.000) | (0.008,0.000) | (0.031,0.016) |
| 14222500 | 0.805 | 0.002 | 0.049 | 0.024 | 0.900 | 0.002 | 0.003 | 0.015 | 0.758 | 0.004 | 0.089 | 0.022 | 0.911 | 0.001 | 0.001 | 0.014 |
| | (0.765,0.838) | (0.012,0.000) | (0.074,0.032) | (0.033,0.019) | (0.847,0.926) | (0.011,0.000) | (0.015,0.000) | (0.023,0.010) | (0.719,0.789) | (0.018,0.000) | (0.117,0.068) | (0.030,0.017) | (0.872,0.926) | (0.008,0.000) | (0.006,0.000) | (0.022,0.009) |
| 12358500 | 0.846 | 0.002 | 0.036 | 0.007 | 0.946 | 0.001 | 0.003 | 0.001 | 0.750 | 0.004 | 0.114 | 0.006 | 0.966 | 0.000 | 0.000 | 0.001 |
| | (0.824,0.868) | (0.008,0.000) | (0.050,0.026) | (0.012,0.004) | (0.913,0.966) | (0.006,0.000) | (0.008,0.001) | (0.002,0.001) | (0.716,0.782) | (0.011,0.001) | (0.148,0.087) | (0.008,0.004) | (0.943,0.975) | (0.002,0.000) | (0.003,0.000) | (0.002,0.001) |
| 9035900 | 0.863 | 0.002 | 0.023 | 0.012 | 0.940 | 0.002 | 0.002 | 0.003 | 0.590 | 0.002 | 0.320 | 0.010 | 0.947 | 0.001 | 0.001 | 0.003 |
| | (0.833,0.881) | (0.013,0.000) | (0.029,0.019) | (0.021,0.006) | (0.893,0.965) | (0.012,0.000) | (0.007,0.000) | (0.004,0.001) | (0.529,0.642) | (0.016,0.000) | (0.424,0.245) | (0.018,0.005) | (0.908,0.962) | (0.009,0.000) | (0.004,0.000) | (0.006,0.002) |
| 9223000 | 0.797 | 0.002 | 0.058 | 0.018 | 0.909 | 0.003 | 0.008 | 0.005 | 0.603 | 0.002 | 0.292 | 0.017 | 0.936 | 0.002 | 0.001 | 0.004 |
| | (0.770,0.820) | (0.020,0.000) | (0.074,0.046) | (0.032,0.011) | (0.849,0.945) | (0.019,0.000) | (0.023,0.002) | (0.008,0.003) | (0.516,0.667) | (0.017,0.000) | (0.451,0.197) | (0.029,0.010) | (0.892,0.957) | (0.012,0.000) | (0.008,0.000) | (0.007,0.003) |
| 13161500 | 0.738 | 0.003 | 0.084 | 0.046 | 0.897 | 0.011 | 0.001 | 0.007 | 0.383 | 0.026 | 0.705 | 0.036 | 0.935 | 0.001 | 0.001 | 0.005 |
| | (0.699,0.778) | (0.026,0.000) | (0.104,0.065) | (0.076,0.022) | (0.825,0.937) | (0.047,0.000) | (0.010,0.000) | (0.012,0.004) | (0.063,0.632) | (0.060,0.008) | (1.682,0.211) | (0.061,0.019) | (0.893,0.954) | (0.009,0.000) | (0.008,0.000) | (0.010,0.003) |
| 8324000 | 0.674 | 0.025 | 0.089 | 0.089 | 0.680 | 0.079 | 0.072 | 0.050 | 0.626 | 0.003 | 0.211 | 0.058 | 0.891 | 0.002 | 0.002 | 0.017 |
| | (0.598,0.728) | (0.122,0.000) | (0.115,0.066) | (0.130,0.063) | (0.571,0.790) | (0.197,0.013) | (0.122,0.032) | (0.086,0.032) | (0.553,0.693) | (0.037,0.000) | (0.296,0.141) | (0.118,0.026) | (0.823,0.930) | (0.016,0.000) | (0.019,0.000) | (0.041,0.008) |
| 9505800 | 0.746 | 0.015 | 0.043 | 0.066 | 0.795 | 0.008 | 0.016 | 0.053 | 0.639 | 0.006 | 0.214 | 0.035 | 0.846 | 0.007 | 0.004 | 0.026 |
| | (0.623,0.812) | (0.077,0.000) | (0.056,0.030) | (0.170,0.021) | (0.653,0.879) | (0.066,0.000) | (0.047,0.003) | (0.136,0.016) | (0.535,0.732) | (0.042,0.000) | (0.370,0.116) | (0.079,0.008) | (0.770,0.919) | (0.037,0.000) | (0.036,0.000) | (0.060,0.007) |
| 2472000 | 0.794 | 0.003 | 0.060 | 0.021 | 0.872 | 0.005 | 0.012 | 0.015 | 0.525 | 0.001 | 0.424 | 0.023 | 0.896 | 0.001 | 0.001 | 0.018 |
| | (0.758,0.823) | (0.012,0.000) | (0.086,0.042) | (0.026,0.017) | (0.829,0.900) | (0.017,0.000) | (0.023,0.006) | (0.020,0.012) | (0.355,0.670) | (0.008,0.000) | (0.797,0.199) | (0.039,0.015) | (0.845,0.922) | (0.009,0.000) | (0.010,0.000) | (0.034,0.011) |
| 3331500 | 0.842 | 0.001 | 0.028 | 0.020 | 0.920 | 0.002 | 0.002 | 0.008 | 0.771 | 0.001 | 0.069 | 0.034 | 0.913 | 0.001 | 0.000 | 0.013 |
| | (0.809,0.864) | (0.007,0.000) | (0.036,0.021) | (0.035,0.012) | (0.895,0.938) | (0.010,0.000) | (0.005,0.000) | (0.013,0.005) | (0.743,0.796) | (0.008,0.000) | (0.091,0.054) | (0.049,0.022) | (0.891,0.929) | (0.006,0.000) | (0.002,0.000) | (0.020,0.009) |
| 4185000 | 0.718 | 0.001 | 0.116 | 0.039 | 0.875 | 0.000 | 0.007 | 0.023 | 0.604 | 0.001 | 0.273 | 0.036 | 0.892 | 0.000 | 0.001 | 0.022 |
| | (0.624,0.759) | (0.004,0.000) | (0.236,0.083) | (0.064,0.025) | (0.842,0.896) | (0.003,0.000) | (0.015,0.003) | (0.038,0.015) | (0.497,0.695) | (0.006,0.000) | (0.466,0.144) | (0.060,0.023) | (0.862,0.914) | (0.003,0.000) | (0.003,0.000) | (0.036,0.014) |
| 1539000 | 0.778 | 0.002 | 0.068 | 0.024 | 0.884 | 0.001 | 0.003 | 0.020 | 0.707 | 0.001 | 0.150 | 0.017 | 0.906 | 0.001 | 0.001 | 0.014 |
| | (0.731,0.816) | (0.013,0.000) | (0.092,0.048) | (0.060,0.012) | (0.838,0.912) | (0.010,0.000) | (0.009,0.000) | (0.046,0.010) | (0.667,0.744) | (0.007,0.000) | (0.194,0.116) | (0.040,0.009) | (0.866,0.930) | (0.007,0.000) | (0.006,0.000) | (0.030,0.008) |
| 1667500 | 0.732 | 0.003 | 0.083 | 0.054 | 0.836 | 0.002 | 0.009 | 0.041 | 0.680 | 0.001 | 0.171 | 0.030 | 0.884 | 0.001 | 0.001 | 0.023 |
| | (0.687,0.765) | (0.024,0.000) | (0.109,0.065) | (0.093,0.030) | (0.787,0.868) | (0.016,0.000) | (0.020,0.002) | (0.069,0.025) | (0.627,0.718) | (0.011,0.000) | (0.242,0.123) | (0.049,0.017) | (0.849,0.910) | (0.011,0.000) | (0.006,0.000) | (0.036,0.014) |
| 3173000 | 0.748 | 0.001 | 0.082 | 0.043 | 0.851 | 0.001 | 0.009 | 0.033 | 0.628 | 0.005 | 0.241 | 0.028 | 0.886 | 0.002 | 0.001 | 0.021 |
| | (0.723,0.770) | (0.011,0.000) | (0.099,0.066) | (0.060,0.028) | (0.821,0.875) | (0.010,0.000) | (0.020,0.003) | (0.046,0.022) | (0.585,0.664) | (0.021,0.000) | (0.305,0.191) | (0.039,0.021) | (0.864,0.902) | (0.014,0.000) | (0.006,0.000) | (0.028,0.016) |

**Table S4**: Statistical performance comparison of Model LSTM15 across metrics $KGE_{SS}$, $M$, $V$, $C$, $JKGE_{SS}$, $M^*$, $V^*$, and $C^*$, evaluated for models trained with $KGE_{SS}$ and $JKGE_{SS}$. Values represent medians and bootstrap ranges (5th–95th percentiles) computed using yearly-block resampling. For $KGE_{SS}$ and $JKGE_{SS}$, higher values indicate better performance, whereas for component metrics lower values (approaching zero) indicate better performance. Results show that training with $JKGE_{SS}$ yields consistently improved generalized performance and reduced component errors, particularly in the bias-related term $M^*$, across catchments.

| Catchment | $JKGE_{ss}$ Trained | | | | | | | | $KGE_{ss}$ Trained | | | | | | | |
|---|---|---|---|---|---|---|---|---|---|---|---|---|---|---|---|---|
| | $JKGE_{ss}$ | V* | M* | C* | $KGE_{ss}$ | V | M | C | $JKGE_{ss}$ | V* | M* | C* | $KGE_{ss}$ | V | M | C |
| 11473900 | **0.801** | **0.001** | **0.068** | **0.006** | **0.955** | **0.001** | **0.001** | **0.002** | **-0.523** | **0.000** | **4.637** | **0.003** | **0.966** | **0.000** | **0.000** | **0.001** |
| | (0.697,0.855) | (0.008,0.000) | (0.168,0.035) | (0.014,0.003) | (0.919,0.970) | (0.006,0.000) | (0.002,0.000) | (0.005,0.001) | (-1.471,-0.060) | (0.003,0.000) | (12.213,2.245) | (0.007,0.001) | (0.946,0.979) | (0.002,0.000) | (0.002,0.000) | (0.003,0.001) |
| 11523200 | **0.881** | **0.000** | **0.024** | **0.003** | **0.982** | **0.000** | **0.000** | **0.000** | **0.724** | **0.001** | **0.146** | **0.006** | **0.972** | **0.000** | **0.000** | **0.001** |
| | (0.851,0.910) | (0.002,0.000) | (0.036,0.014) | (0.008,0.001) | (0.974,0.988) | (0.000,0.000) | (0.001,0.000) | (0.001,0.000) | (0.656,0.781) | (0.007,0.000) | (0.228,0.090) | (0.010,0.003) | (0.961,0.980) | (0.001,0.000) | (0.001,0.000) | (0.002,0.001) |
| 14222500 | **0.892** | **0.000** | **0.020** | **0.003** | **0.974** | **0.000** | **0.000** | **0.001** | **0.819** | **0.003** | **0.056** | **0.006** | **0.962** | **0.000** | **0.000** | **0.002** |
| | (0.848,0.938) | (0.001,0.000) | (0.042,0.006) | (0.005,0.001) | (0.961,0.983) | (0.000,0.000) | (0.001,0.000) | (0.002,0.001) | (0.799,0.841) | (0.007,0.000) | (0.071,0.043) | (0.009,0.004) | (0.952,0.970) | (0.001,0.000) | (0.001,0.000) | (0.003,0.002) |
| 12358500 | **0.905** | **0.000** | **0.014** | **0.003** | **0.982** | **0.000** | **0.000** | **0.000** | **0.780** | **0.001** | **0.090** | **0.006** | **0.980** | **0.000** | **0.000** | **0.001** |
| | (0.881,0.927) | (0.005,0.000) | (0.022,0.008) | (0.004,0.002) | (0.969,0.990) | (0.001,0.000) | (0.001,0.000) | (0.000,0.000) | (0.746,0.811) | (0.007,0.000) | (0.122,0.062) | (0.009,0.004) | (0.968,0.986) | (0.001,0.000) | (0.001,0.000) | (0.001,0.000) |
| 9035900 | **0.912** | **0.000** | **0.012** | **0.002** | **0.978** | **0.000** | **0.000** | **0.000** | **0.845** | **0.000** | **0.045** | **0.003** | **0.983** | **0.000** | **0.000** | **0.000** |
| | (0.878,0.941) | (0.002,0.000) | (0.023,0.005) | (0.006,0.001) | (0.953,0.990) | (0.003,0.000) | (0.002,0.000) | (0.001,0.000) | (0.812,0.874) | (0.003,0.000) | (0.066,0.029) | (0.005,0.001) | (0.966,0.991) | (0.001,0.000) | (0.001,0.000) | (0.000,0.000) |
| 9223000 | **0.806** | **0.001** | **0.063** | **0.010** | **0.968** | **0.000** | **0.000** | **0.001** | **0.586** | **0.012** | **0.316** | **0.014** | **0.962** | **0.000** | **0.001** | **0.001** |
| | (0.752,0.846) | (0.008,0.000) | (0.096,0.038) | (0.023,0.005) | (0.926,0.982) | (0.005,0.000) | (0.003,0.000) | (0.003,0.000) | (0.523,0.639) | (0.031,0.001) | (0.436,0.229) | (0.025,0.007) | (0.934,0.977) | (0.003,0.000) | (0.006,0.000) | (0.002,0.001) |
| 13161500 | **0.805** | **0.001** | **0.055** | **0.018** | **0.961** | **0.000** | **0.001** | **0.001** | **0.519** | **0.007** | **0.442** | **0.012** | **0.977** | **0.000** | **0.000** | **0.001** |
| | (0.762,0.846) | (0.008,0.000) | (0.088,0.033) | (0.032,0.009) | (0.935,0.975) | (0.003,0.000) | (0.005,0.000) | (0.002,0.001) | (0.374,0.620) | (0.018,0.001) | (0.766,0.262) | (0.020,0.008) | (0.956,0.984) | (0.001,0.000) | (0.003,0.000) | (0.001,0.000) |
| 8324000 | **0.822** | **0.001** | **0.046** | **0.014** | **0.948** | **0.001** | **0.002** | **0.003** | **0.559** | **0.011** | **0.325** | **0.044** | **0.918** | **0.002** | **0.001** | **0.008** |
| | (0.783,0.853) | (0.007,0.000) | (0.059,0.035) | (0.038,0.006) | (0.904,0.968) | (0.005,0.000) | (0.008,0.000) | (0.007,0.001) | (0.469,0.642) | (0.038,0.001) | (0.487,0.202) | (0.106,0.020) | (0.866,0.950) | (0.012,0.000) | (0.009,0.000) | (0.023,0.004) |
| 9505800 | **0.841** | **0.004** | **0.034** | **0.009** | **0.875** | **0.008** | **0.012** | **0.007** | **0.581** | **0.002** | **0.341** | **0.004** | **0.946** | **0.001** | **0.001** | **0.003** |
| | (0.735,0.889) | (0.050,0.000) | (0.055,0.020) | (0.045,0.002) | (0.761,0.939) | (0.059,0.001) | (0.028,0.002) | (0.035,0.002) | (0.449,0.697) | (0.017,0.000) | (0.601,0.170) | (0.017,0.001) | (0.889,0.969) | (0.013,0.000) | (0.005,0.000) | (0.012,0.001) |
| 2472000 | **0.885** | **0.003** | **0.020** | **0.003** | **0.953** | **0.002** | **0.000** | **0.002** | **0.712** | **0.000** | **0.162** | **0.002** | **0.970** | **0.000** | **0.000** | **0.001** |
| | (0.857,0.910) | (0.012,0.000) | (0.030,0.012) | (0.004,0.002) | (0.922,0.976) | (0.008,0.000) | (0.002,0.000) | (0.002,0.001) | (0.590,0.779) | (0.002,0.000) | (0.333,0.094) | (0.004,0.001) | (0.944,0.982) | (0.001,0.000) | (0.003,0.000) | (0.003,0.000) |
| 3331500 | **0.911** | **0.000** | **0.013** | **0.002** | **0.981** | **0.000** | **0.000** | **0.001** | **0.836** | **0.001** | **0.047** | **0.005** | **0.962** | **0.000** | **0.000** | **0.002** |
| | (0.875,0.943) | (0.002,0.000) | (0.026,0.005) | (0.006,0.001) | (0.969,0.989) | (0.000,0.000) | (0.000,0.000) | (0.002,0.000) | (0.805,0.859) | (0.007,0.000) | (0.066,0.034) | (0.010,0.003) | (0.942,0.973) | (0.002,0.000) | (0.002,0.000) | (0.004,0.001) |
| 4185000 | **0.812** | **0.001** | **0.063** | **0.007** | **0.947** | **0.000** | **0.002** | **0.003** | **0.650** | **0.000** | **0.243** | **0.002** | **0.968** | **0.000** | **0.000** | **0.001** |
| | (0.755,0.859) | (0.004,0.000) | (0.107,0.033) | (0.011,0.004) | (0.925,0.964) | (0.002,0.000) | (0.005,0.000) | (0.005,0.002) | (0.548,0.736) | (0.001,0.000) | (0.403,0.135) | (0.005,0.001) | (0.954,0.977) | (0.001,0.000) | (0.001,0.000) | (0.003,0.001) |
| 1539000 | **0.873** | **0.000** | **0.027** | **0.005** | **0.958** | **0.000** | **0.000** | **0.003** | **0.622** | **0.000** | **0.284** | **0.002** | **0.968** | **0.000** | **0.000** | **0.002** |
| | (0.835,0.900) | (0.003,0.000) | (0.045,0.016) | (0.015,0.002) | (0.929,0.972) | (0.002,0.000) | (0.001,0.000) | (0.008,0.001) | (0.506,0.713) | (0.002,0.000) | (0.486,0.162) | (0.006,0.001) | (0.937,0.980) | (0.002,0.000) | (0.001,0.000) | (0.005,0.001) |
| 1667500 | **0.791** | **0.002** | **0.077** | **0.008** | **0.926** | **0.002** | **0.003** | **0.006** | **0.621** | **0.000** | **0.284** | **0.004** | **0.947** | **0.000** | **0.001** | **0.004** |
| | (0.735,0.846) | (0.008,0.000) | (0.124,0.040) | (0.014,0.005) | (0.890,0.950) | (0.006,0.000) | (0.011,0.000) | (0.010,0.004) | (0.432,0.769) | (0.001,0.000) | (0.637,0.104) | (0.009,0.002) | (0.914,0.967) | (0.001,0.000) | (0.008,0.000) | (0.009,0.002) |
| 3173000 | **0.874** | **0.002** | **0.025** | **0.003** | **0.949** | **0.002** | **0.001** | **0.002** | **0.780** | **0.005** | **0.085** | **0.005** | **0.935** | **0.003** | **0.001** | **0.004** |
| | (0.827,0.915) | (0.006,0.000) | (0.050,0.011) | (0.007,0.002) | (0.928,0.969) | (0.005,0.000) | (0.002,0.000) | (0.005,0.001) | (0.728,0.824) | (0.020,0.001) | (0.128,0.057) | (0.014,0.002) | (0.874,0.969) | (0.012,0.000) | (0.008,0.000) | (0.012,0.001) |

## Figures

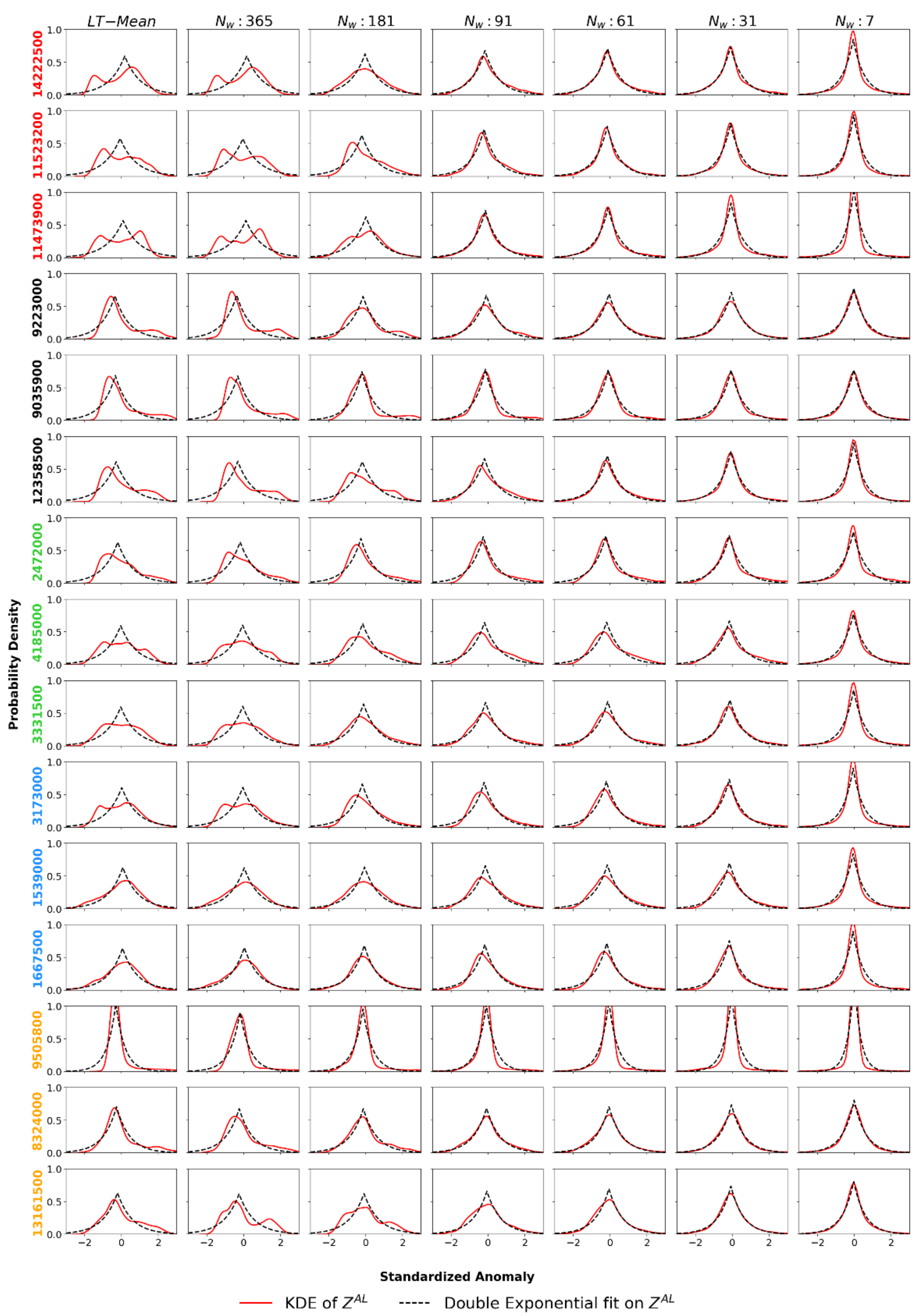

**Figure S1:** Distribution of standardized anomalies ($Z^{AL}$) for all fifteen catchments, belonging to different hydroclimatic zones, computed under the MA approach using different benchmark options. Window lengths ($N_w$) progressively decrease from left to right, starting from the long-term mean (LT-Mean) to $N_w = 7$ .The kernel density estimates (KDEs) become increasingly well-behaved as $N_w$ decreases, ultimately showing close agreement with the double-exponential distribution (black dashed line)

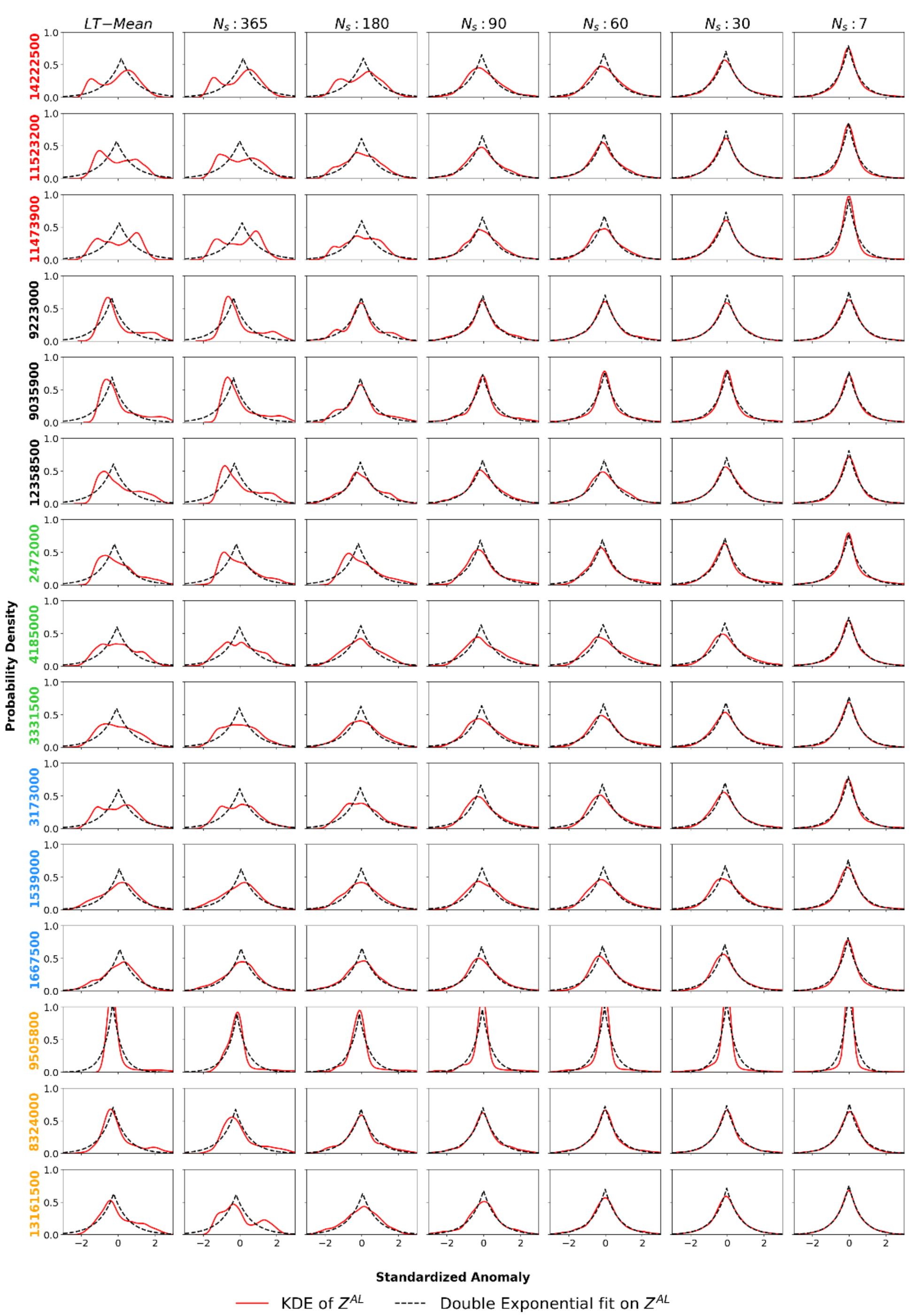

**Figure S2:** Distribution of standardized anomalies ($Z^{AL}$) for all fifteen catchments, belonging to different hydroclimatic zones, computed under the SA approach using different benchmark options. Section lengths ($N_s$) progressively decrease from left to right, starting from the long-term mean (LT-Mean) to $N_s = 7$ .The kernel density estimates (KDEs) become increasingly well-behaved as $N_s$ decreases, ultimately showing close agreement with the double-exponential distribution (black dashed line)

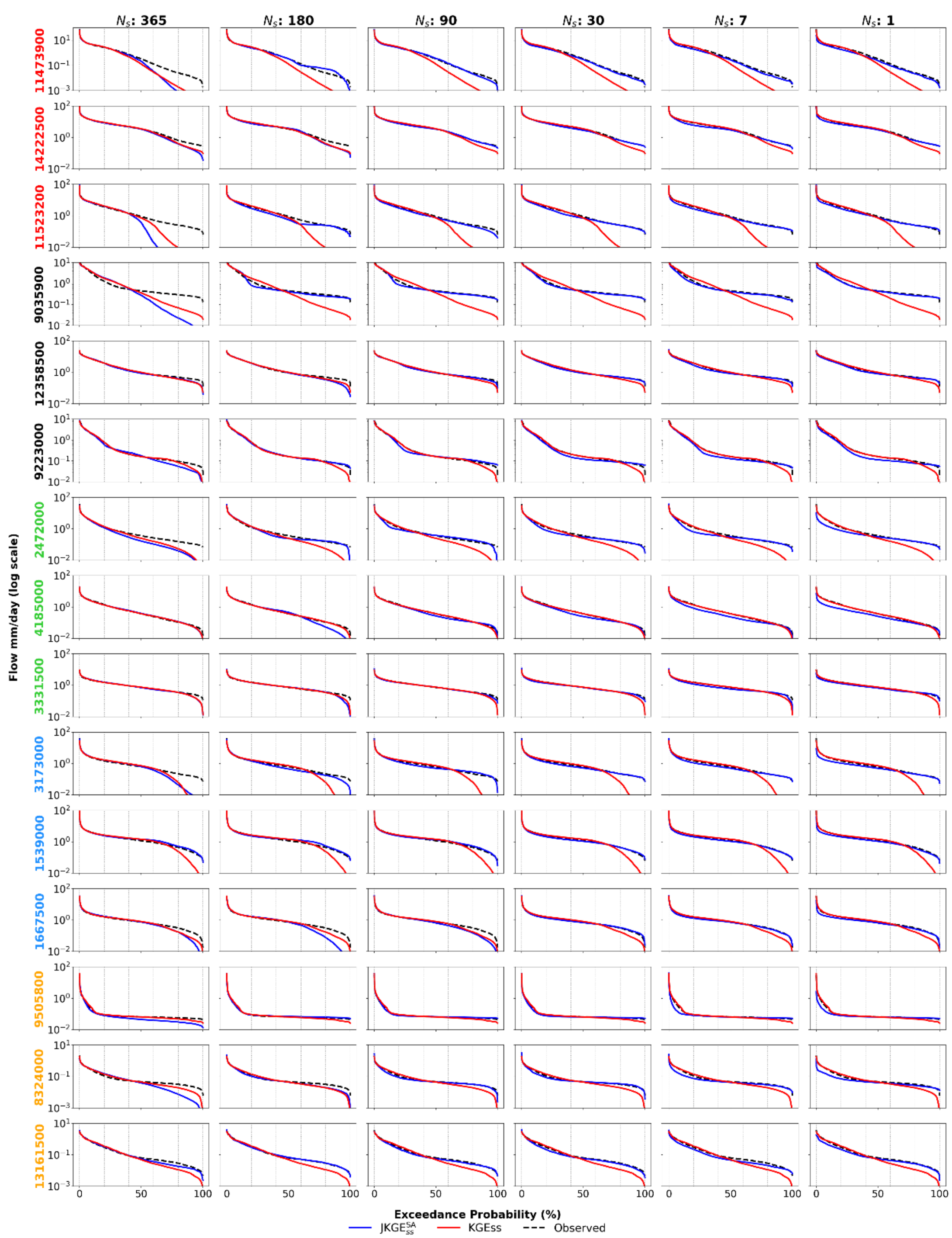

**Figure S3**: Flow Duration Curves (FDCs) of all 15 catchments. The JKGEss model (blue) is trained with varying section lengths (Ns), showing improved performance as Ns decreases from 365 to 30 days, with Ns = 30 performing best across all catchments. Performance degrades for shorter Ns (7 or 1), especially at high and medium flows. KGEss (red) is shown for reference and remains constant across columns for each catchment.

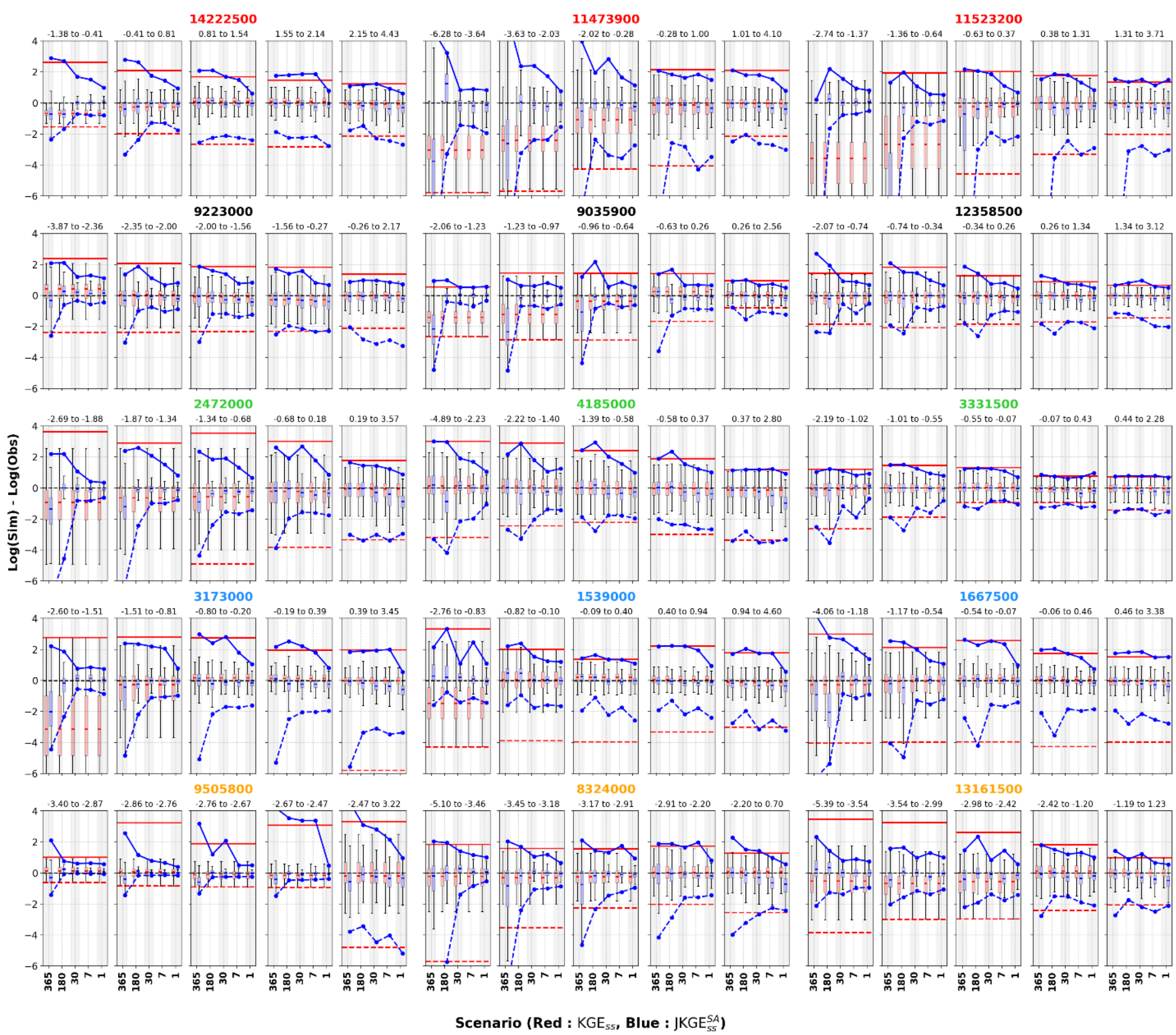

**Figure S4:** Log-flow anomalies across five flow groups (FG1–FG5). Models trained with $\mathrm{KGE}_{ss}$ (red) are compared with $\mathrm{JKGE}_{ss}^{\mathrm{SA}(N_s)}$ (blue) across different section lengths ($N_s$). Anomalies, computed as $\log(sim) - \log(obs)$ ,are shown as boxplots with minimum (dashed lines) and maximum (solid lines) values. $\mathrm{JKGE}_{ss}^{\mathrm{SA}(N_s)}$improves low- and intermediate-flow performance as $N_s$decreases, with $N_s \approx 30$days yielding the most balanced representation across flow regimes. High-flow anomalies are generally well captured by both metrics .Red horizontal lines show $\mathrm{KGE}_{ss}$values constant across $N_s$for comparison, while blue lines illustrate changes in anomaly range with decreasing $N_s$.

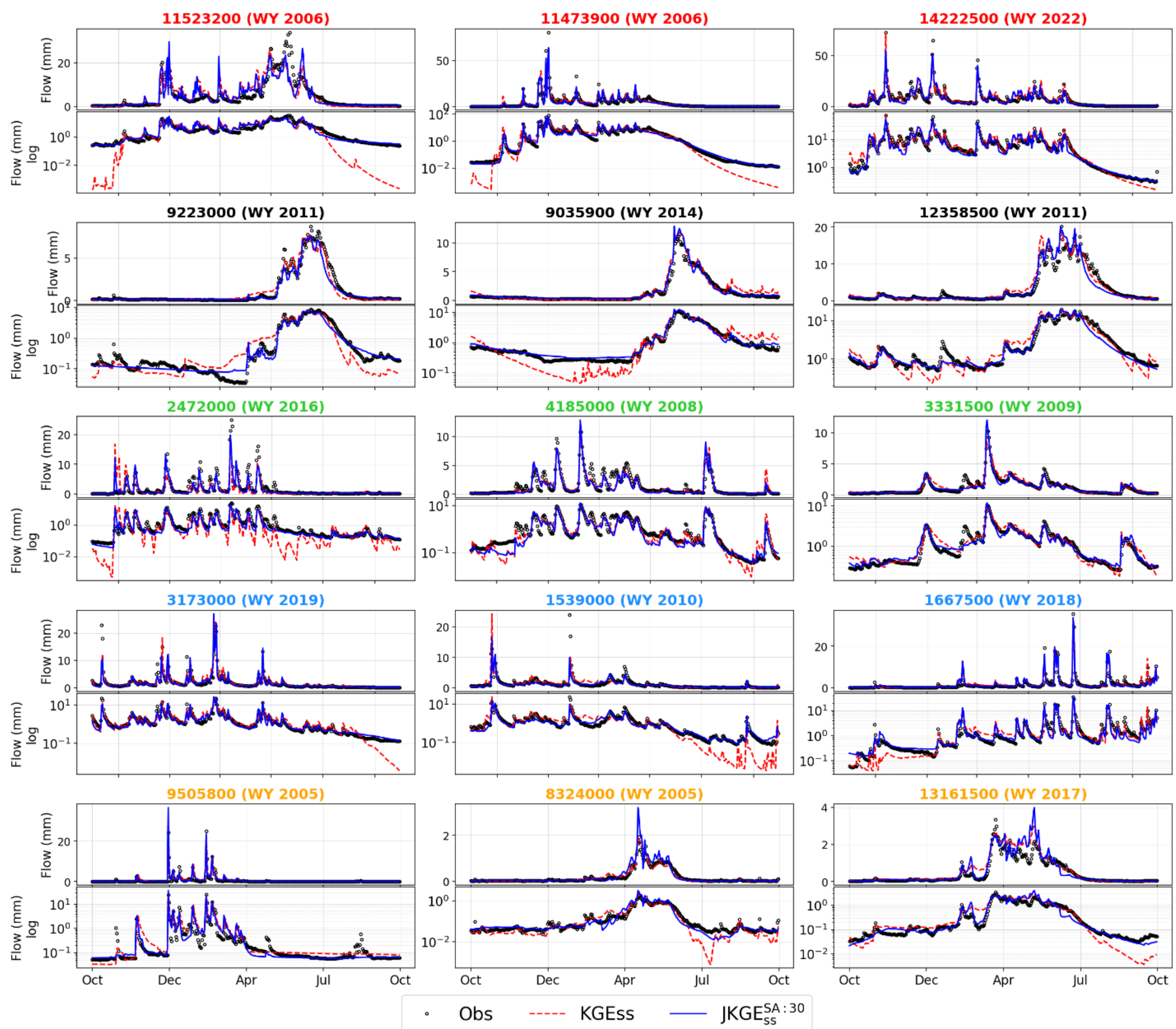

**Figure S5**: Observed (black) and simulated streamflow hydrographs (mm/day) for representative wet years on natural and log-scale. Simulations from $\mathrm{KGE}_{ss}$are shown in red, and $\mathrm{JKGE}_{ss}^{\mathrm{SA(30)}}$in blue. $\mathrm{JKGE}_{ss}^{\mathrm{SA(30)}}$better captures both high- and low-flow dynamics, including recession behavior, compared to $\mathrm{KGE}_{ss}$

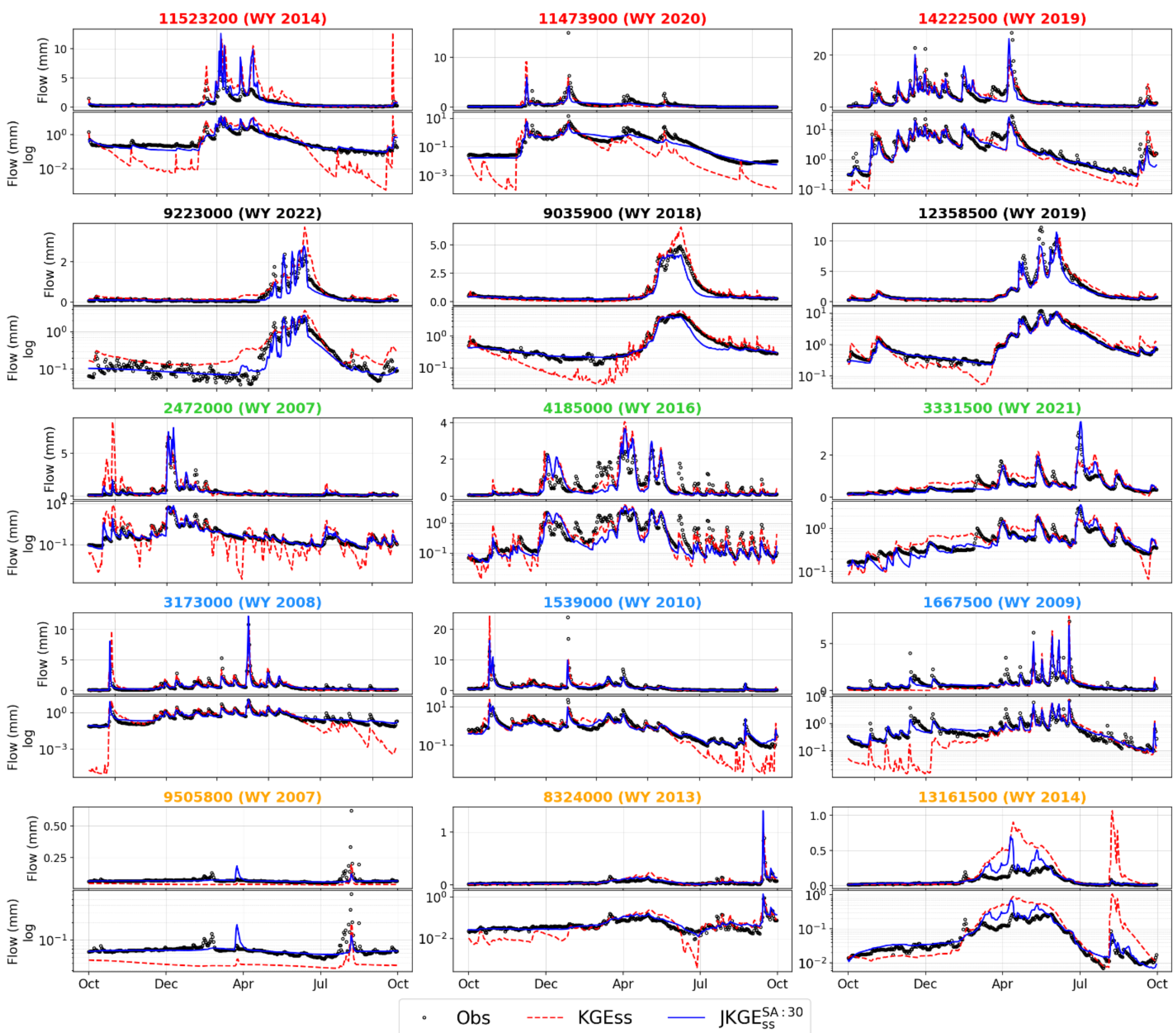

**Figure S6**: Observed (black) and simulated streamflow hydrographs (mm/day) for representative dry years on natural and log-scale. Simulations from $\mathrm{KGE}_{ss}$ are shown in red, and $\mathrm{JKGE}_{ss}^{\mathrm{SA}(30)}$ in blue. $\mathrm{JKGE}_{ss}^{\mathrm{SA}(30)}$ better captures both high- and low-flow dynamics, including recession behavior, compared to $\mathrm{KGE}_{ss}$ ,with improvements most pronounced during dry years.

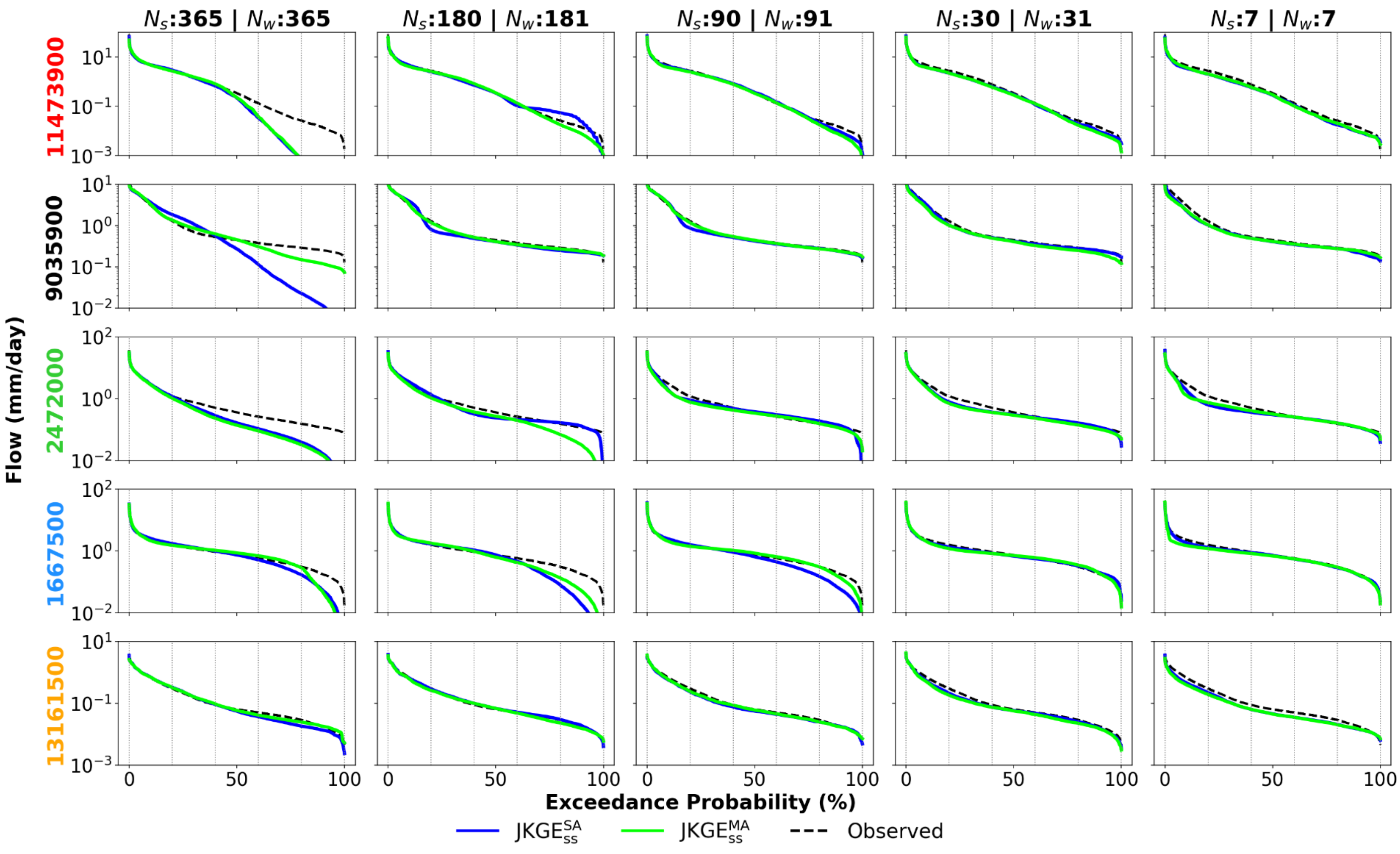

**Figure S7**: Comparison of Flow Duration Curves (FDCs) obtained using $\mathrm{JKGE}_{ss}^{\mathrm{MA}(N_w)}$ (moving-average mean) and $\mathrm{JKGE}_{ss}^{\mathrm{SA}(N_s)}$(section-wise mean) across different segment sizes. Both approaches show similar performance, with improvements as window size decreases and optimal results at $N_w \approx 31$ days (comparable to $N_s = 30$ days).

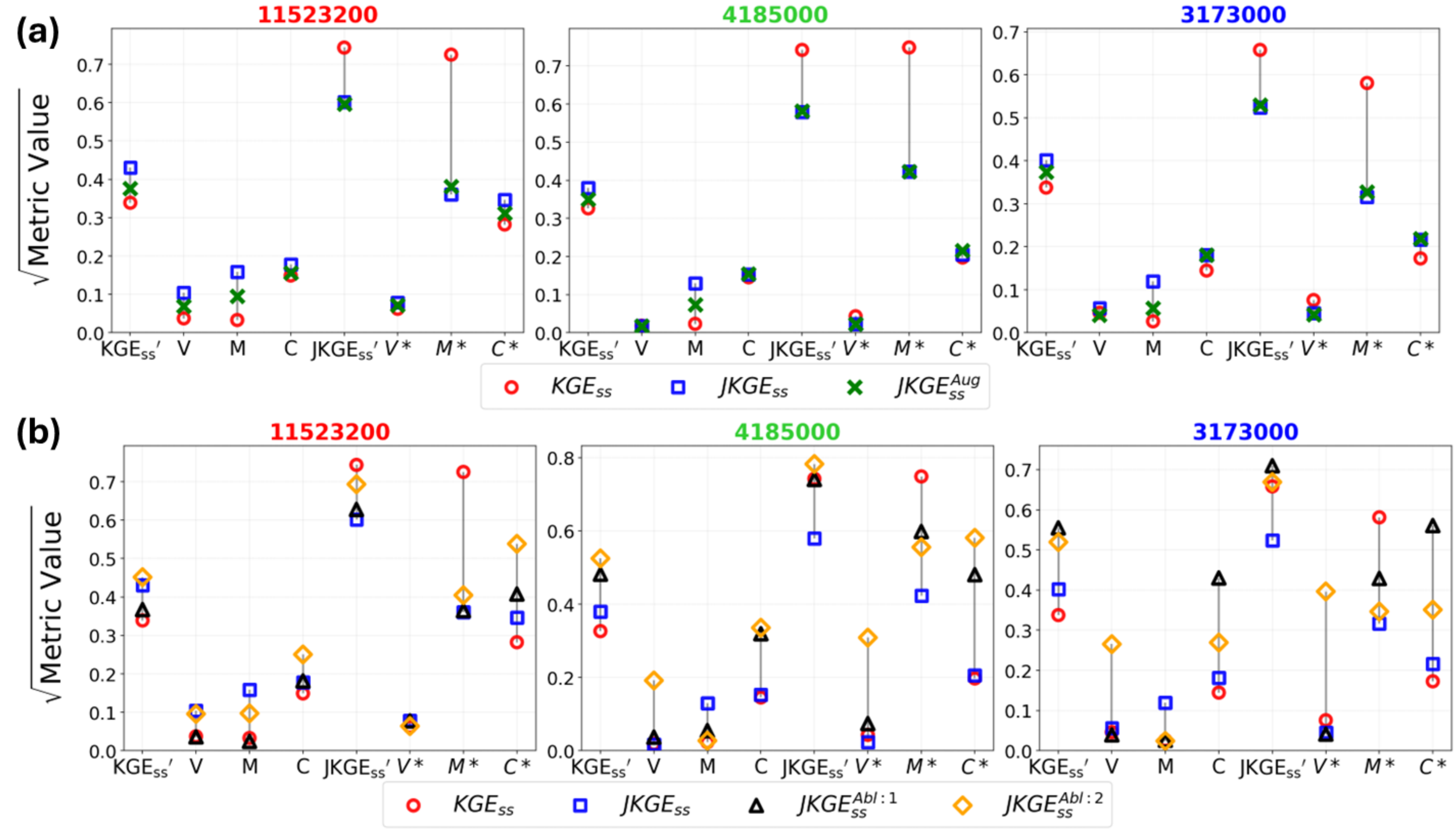

**Figure S8**: Evaluation of augmented and ablated formulations of $\mathrm{JKGE}_{ss}$ (a) Results for the augmented metric $\mathrm{JKGE}_{ss}^{\mathrm{Aug}} = 1 - \sqrt{(M + M^{*} + V^{*} + C^{*})/2}$ ,showing improved $\mathrm{KGE}_{ss}$ and water-balance component $M$ without affecting other components. (b) Results for ablated versions $\mathrm{JKGE}_{ss}^{(\mathrm{Abl:1})} = 1 - \sqrt{M + M^{*} + V^{*}}$ and $\mathrm{JKGE}_{ss}^{(\mathrm{Abl:2})} = 1 - \sqrt{M + M^{*}}$ ,demonstrating degraded performance when omitting $V^{*}$ and $C^{*}$. Results indicate that all components $(M, M^{*}, V^{*}, C^{*})$ contribute meaningfully to model performance.

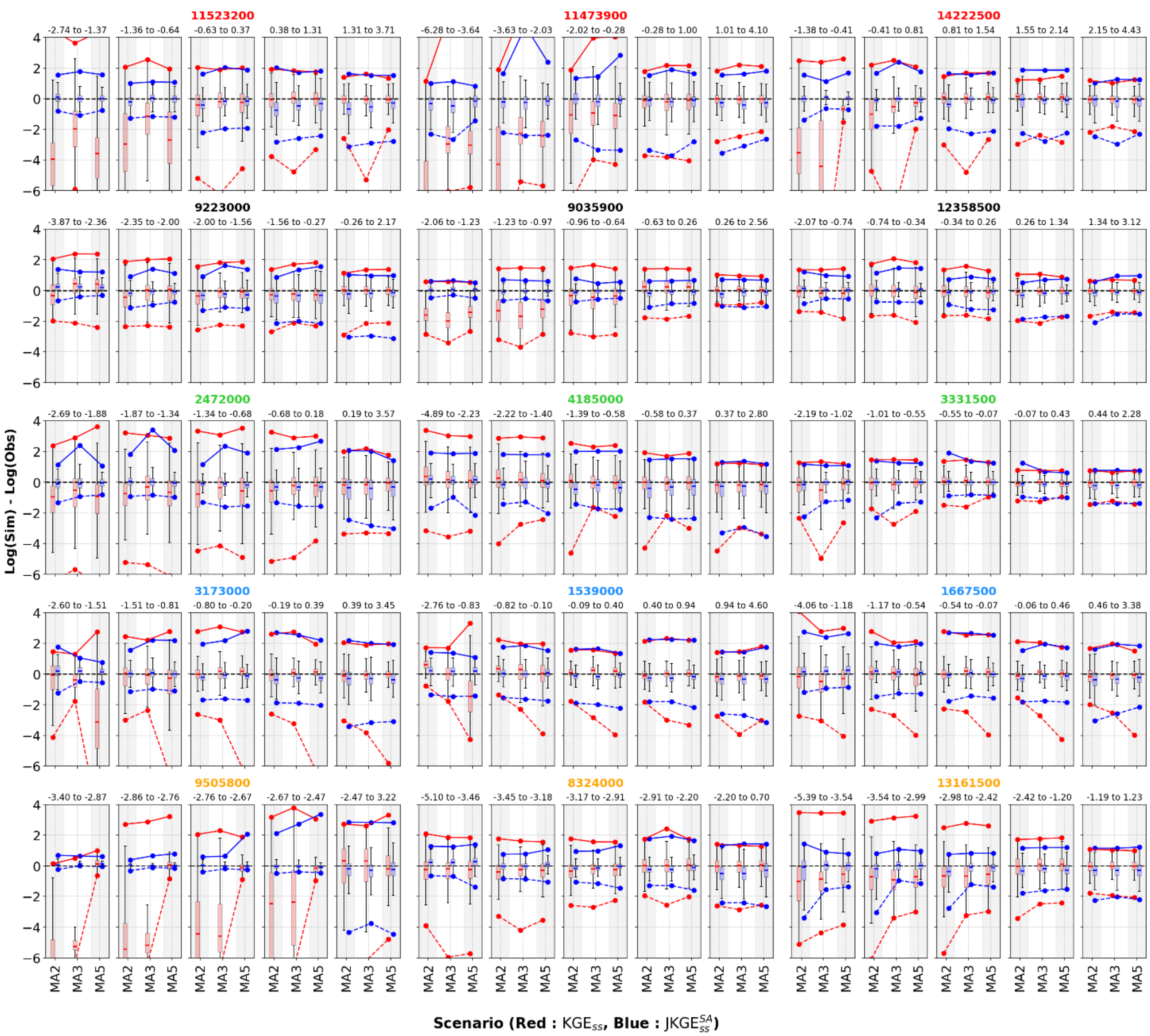

**Figure S9:** Boxplots showing impact of model simplification on performance across flow groups for MA3 and MA2 models. Models trained with $\mathrm{JKGE}_{ss}$ (blue) consistently outperform those trained with $\mathrm{KGE}_{ss}$(red) with improvements most pronounced for low-flow conditions. Performance differences decrease toward higher flows, with only minor degradation in some catchments at very high flows under $\mathrm{JKGE}_{ss}$.

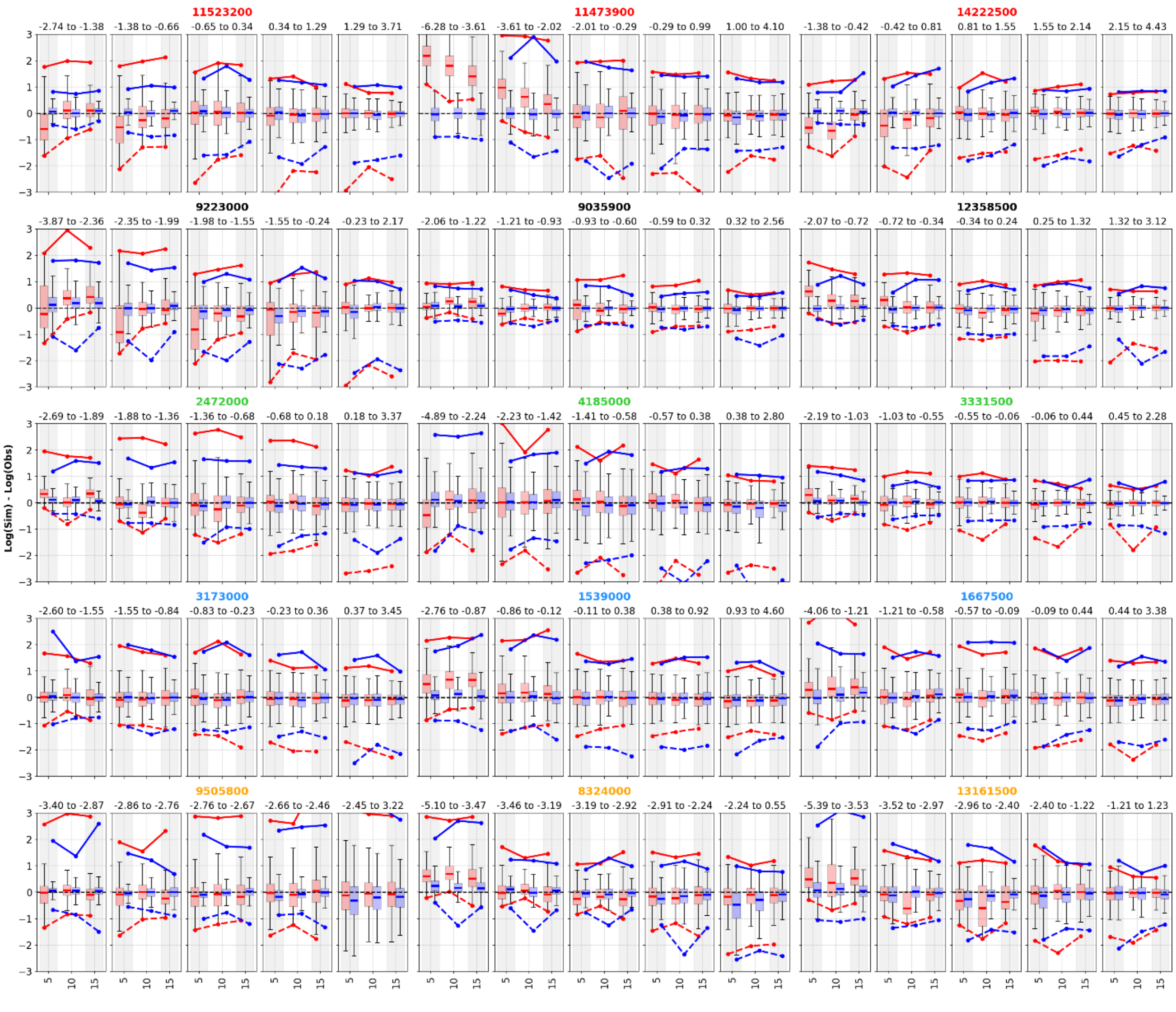

**Figure S10:** Performance of LSTM-based models (5, 10, and 15 nodes) across flow groups when trained with $\mathrm{KGE}_{ss}$versus $\mathrm{JKGE}_{ss}$ .$\mathrm{JKGE}_{ss}$consistently yields improved and more stable performance, with reduced spread and fewer extreme anomalies, particularly for low- to medium-flow conditions. Performance gains increase with model complexity and are most pronounced in hydroclimatic regimes with strong non-stationarity, while differences are smaller for high flows.

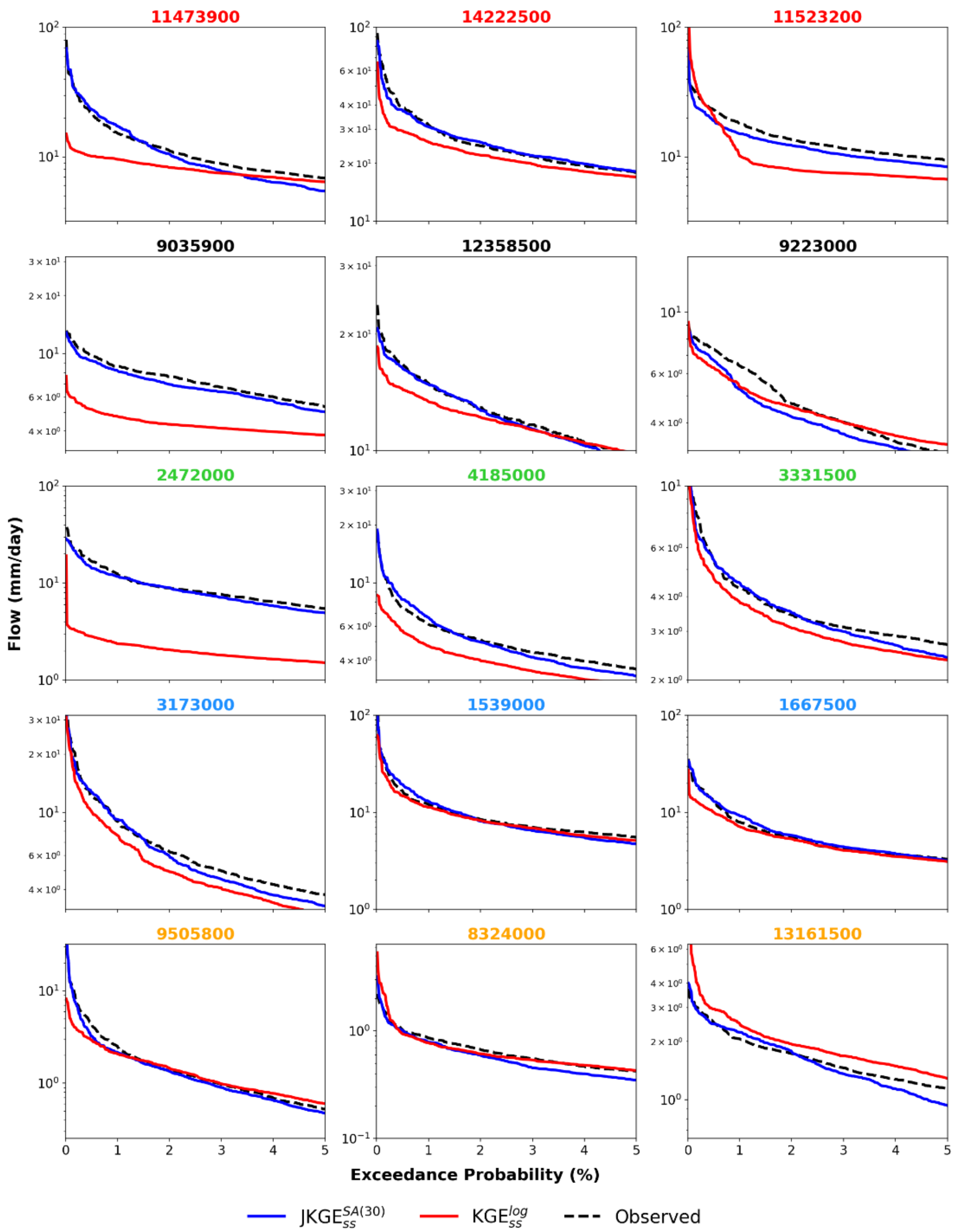

**Figure S11:** Flow Duration Curves (FDCs) comparing $\mathrm{KGE}_{ss}^{\mathrm{log}}$ and $\mathrm{JKGE}_{ss}^{\mathrm{SA(30)}}$ focusing on high flows (0–5% exceedance probability). $\mathrm{KGE}_{ss}^{\mathrm{log}}$ systematically underestimates peak flows, whereas $\mathrm{JKGE}_{ss}^{\mathrm{SA(30)}}$ better reproduces them.